\documentclass{article} 
\usepackage{iclr2023_conference,times}

\usepackage{hyperref}
\usepackage{url}
\usepackage{graphicx}
\usepackage{amsmath}
\usepackage{amssymb}
\usepackage{bm,upgreek}
\usepackage{multirow}
\usepackage{booktabs, multirow} 
\usepackage{soul}
\usepackage{caption}
\usepackage{subcaption}
\usepackage{changepage,threeparttable} 
\usepackage{enumitem}
\usepackage[title]{appendix}

\usepackage{natbib}
\setcitestyle{numbers}
\setcitestyle{square}

\newcommand{\bW}{\mathbf{W}}
\newcommand{\by}{\mathbf{y}}
\newcommand{\bx}{\mathbf{x}}
\newcommand{\bk}{\mathbf{k}}
\newcommand{\bz}{\mathbf{z}}
\newcommand{\bq}{\mathbf{q}}
\newcommand{\bQ}{\mathbf{Q}}
\newcommand{\bs}{\mathbf{s}}
\newcommand{\bh}{\mathbf{h}}
\newcommand{\bJ}{\mathbf{J}}
\newcommand{\bH}{\mathbf{H}}
\newcommand{\bsigma}{\bm{\sigma}}

\newcommand{\cD}{\mathcal{D}}

\newcommand{\ie}{\textit{i}.\textit{e}.}
\newcommand{\eg}{\textit{e}.\textit{g}.}
\newcommand{\etal}{\textit{et al}.}

\title{Training Multilayer Perceptrons by Sampling with Quantum Annealers}

\author{Frances Fengyi Yang, Michele Sasdelli \& Tat-Jun Chin \\
School of Computer Science\\
The University of Adelaide\\
Adelaide, SA 5005, Australia \\
\texttt{\{fengyi.yang,michele.sasdelli,tat-jun.chin\}@adelaide.edu.au} \\
}

\iclrfinalcopy 
\begin{document}

\maketitle

\begin{abstract}
A successful application of quantum annealing to machine learning is training restricted Boltzmann machines (RBM). However, many neural networks for vision applications are feedforward structures, such as multilayer perceptrons (MLP). Backpropagation is currently the most effective technique to train MLPs for supervised learning. This paper aims to be forward-looking by exploring the training of MLPs using quantum annealers. We exploit an equivalence between MLPs and energy-based models (EBM), which are a variation of RBMs with a maximum conditional likelihood objective. This leads to a strategy to train MLPs with quantum annealers as a sampling engine. We prove our setup for MLPs with sigmoid activation functions and one hidden layer, and demonstrated training of binary image classifiers on small subsets of the MNIST and Fashion-MNIST datasets using the D-Wave quantum annealer. Although problem sizes that are feasible on current annealers are limited, we obtained comprehensive results on feasible instances that validate our ideas. Our work establishes the potential of quantum computing for training MLPs.
\end{abstract}

\section{Introduction}\label{intro}
The encouraging progress in building quantum computers in the past decade has sparked interest in quantum computing for computer vision (CV)~\cite{larasati2022trends, birdal2021quantum, benkner2020adiabatic, benkner2021q} and machine learning (ML)~\cite{biamonte2017quantum, schuld2015introduction, schuld2018supervised, wittek2014quantum}. Major foci include using quantum computers to speed up computations and learning more expressive models using quantum representations.

There are two main quantum computing paradigms: gate quantum computing (GQC) and adiabatic quantum computing (AQC). Currently, the practical realisation of AQC in the form of quantum annealers has yielded machines with a larger number of qubits. The primary example is D-Wave Advantage~\cite{dwave2022advantage}, whose quantum processing unit (QPU) contains $>$ 5,000 qubits. While this is still far from a size that will make quantum computing truly revolutionary ($>$~1~M qubits), it can already support exploratory research.

 A notable example of using quantum annealers in ML is training Boltzmann machines (BM)~\cite{dixit2021training, winci2020path, adachi2015application}, which are powerful generative models that can solve complex CV tasks such as classification \cite{nie2015generative}, image labelling \cite{kae2013augmenting} and facial feature tracking \cite{wu2013facial}. BMs are usually trained via contrastive divergence (CD) \cite{carreira2005contrastive}, which involves generating samples from proposal distributions. Computationally this requires matrix multiplications that scale cubically with the number of variables \cite{cormen2022introduction}, which can be costly for large problem sizes. AQC enables a workaround that samples directly from the distribution in constant time with quantum annealers. In practice, quantum sampling is not straightforward since current noisy intermediate-scale quantum (NISQ) devices are prone to noise, and careful tuning of the control parameters such as the qubit biases, coupling strength and the system temperature are required. Nonetheless, promising results from training BMs with quantum sampling have been reported~\cite{adachi2015application, dixit2021training}.

 \begin{figure}[t]\centering
    \includegraphics[width=.85\linewidth]{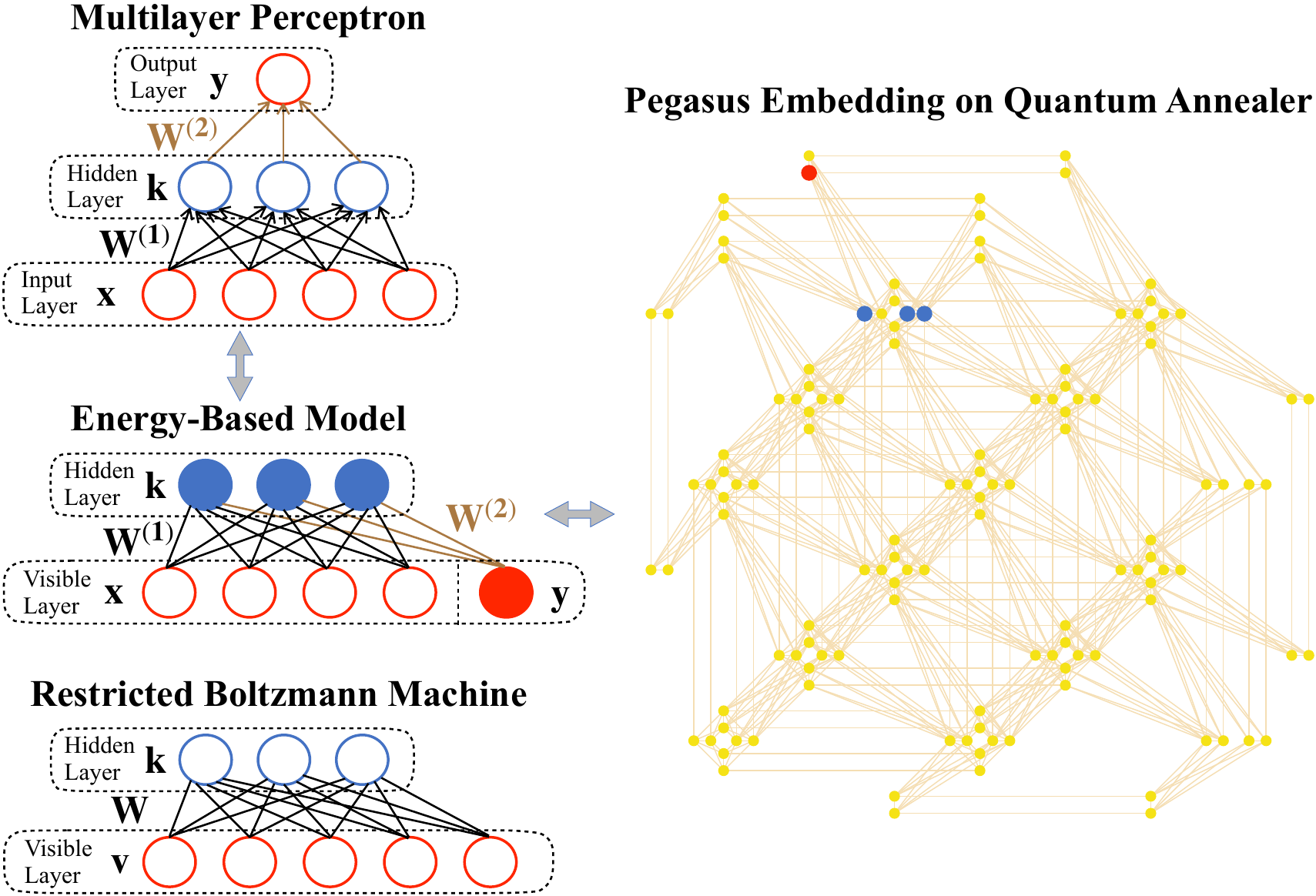}
    \captionsetup{width=12cm}
    \caption{RBM and EBM are generative models, where EBM is a variant of RBM with maximum conditional likelihood objective. MLP is non-generative and is typically employed for supervised learning. We establish the equivalence in training MLP and EBM, which allows to embed the corresponding proposal distribution for MLP in the QPU to achieve quantum sampling for training MLPs.}\vspace{-10pt}
    \label{models}
\end{figure}

Less explored is the training of feedforward neural networks (FNN)~\cite{svozil1997introduction} with quantum annealers. Unlike BMs which are usually used to model data distributions, FNNs are usually employed to learn functions, \eg, for classification or regression, which underpin many important CV applications. Notable FNNs are multilayer perceptrons (MLP)~\cite{taud2018multilayer}, convolutional neural networks (CNN)~\cite{albawi2017understanding} and recurrent neural networks (RNN)~\cite{sherstinsky2020fundamentals}. The most successful method to train FNNs is backpropagation \cite{rumelhart1986learning}, which minimises a loss that is typically related to supervised learning. Since FNNs are not generative, it is unclear how quantum sampling can be employed for training FNNs.

Sasdelli and Chin~\cite{sasdelli21quantum} leveraged the ability of quantum annealers to solve combinatorial problems to train binary neural networks (BNN), which are a type of FNN. This requires formulating learning as quadratic unconstrained binary optimisation (QUBO). However, the resulting QUBO consumes a number of qubits that scales with the input dimensions and the number of weights in the BNN, which restricts the method to low-dimensional problems and/or low capacity models on currently limited quantum annealers. 

\subsection{Contributions}\label{sec:contributions}

In this paper, we explore quantum sampling to train MLPs. We exploit an equivalence between training optimisation in MLPs and sampling proposal distributions in energy-based models (EBM)~\cite{lecun2006tutorial}. The latter are a variation to restricted Boltzmann machines (RBM)~\cite[Chap.~6]{smolensky1986information} with a maximum conditional likelihood objective; see Fig.~\ref{models}.

By exploiting the equivalence above, we developed a novel quantum sampling method to train MLPs for supervised learning. Our approach requires the number of qubits equal to the number of hidden units plus one, and allows for non-binary weights and batch training. Compared to~\cite{sasdelli21quantum}, our method allows larger networks to be trained given the same qubit count. Though the network size is ultimately limited by the capacity of current QPUs (see Sec.~\ref{sec:shortcomings}), comprehensive results on feasible instances validate our ideas.

Our work differs from implementing and training neural networks on GQC, \eg,~\cite{beer2020training}, where the neural networks are inherently quantum, thus requiring quantum computers for training and inference. In contrast, our work employs quantum computers on the training optimisation problem, and inference on the model can be performed classically.

\subsection{Shortcomings and outlook}\label{sec:shortcomings}

Our method cannot yet beat backpropagation.
The capacity of current QPUs limits the size of feasible MLPs (note that the sparse topology of the QPU necessitates a minor embedding step that increases the number of qubits consumed; see Fig.~\ref{models}). We were able to train an MLP with one hidden layer of 548 units and input dimension of 784 for the MNIST and Fashion-MNIST datasets. Overheads in data transfer between CPU and QPU also add to training times that are not competitive against classical solutions.

The value of our work lies in the conceptual innovations that bring quantum computing to bear on a problem of interest in the CV and ML community. Thus, our work enables the community to leverage the potential benefits of quantum technology, which is improving rapidly.

\section{Related work}\label{related}

\subsection{BM and MLP}

A BM is an undirected graphical neural network with stochastic variables in both the visible and hidden layers, with full connections in between. The structure of a BM encodes a joint probability distribution of the variables. 

Unlike BMs, standard MLPs are non-generative and are usually employed to learn classification and regression functions~\cite{rumelhart1985learning}. MLP is a fundamental FNN structure that underpin more advanced models such as CNN and RNN.

\subsection{Quantum annealing in CV and ML}

Quantum computing offers a new paradigm to solve challenging computational problems. The potential for CV and ML was recognised quite early on~\cite{dalal2005histograms, golyanik2016gravitational, viola2001rapid, girshick2014rich, do2016learning}. Recent progress in building quantum computers has reinvigorated interested in the topic. In particular, quantum annealers now contain $>5,000$ qubits~\cite{dwave2022advantage}, which allow small instances of practical problems to be examined. We survey recent works on quantum annealing (QA) for CV and ML. 
\vspace{-12pt}
\paragraph{Combinatorial optimisation}

The ability of quantum annealers to solve combinatorial optimisation has been examined for graph matching \cite{benkner2020adiabatic}, view synchronisation \cite{birdal2021quantum}, robust geometric fitting \cite{doan2022hybrid}, object tracking \cite{zaech2022adiabatic} and training binary neural networks \cite{sasdelli21quantum} on a small scale. However, real quantum annealers are imperfect; fundamentally they do not fully satisfy the assumptions of the adiabatic theorem \cite{farhi2000quantum}. Other constraints such as limited precision and hyperparameter tuning~\cite{doan2022hybrid} also affect solution optimality.

\vspace{-12pt}
\paragraph{Quantum sampling}

By exerting some level of control over the annealing process, quantum annealers can sample from Boltzmann-like distributions~\cite{benedetti2016estimation, amin2018quantum, raymond2016global}. Benedetti \etal~\cite{benedetti2017quantum} built a hardware-embedded generative model on a D-Wave quantum annealer with 1,152 qubits for image reconstruction, generation and classification on simple input images of size 7$\times$6. In the classification task, the generative model was only used for pre-training. Amin \etal~\cite{amin2018quantum} implemented the energy function of BM on the quantum annealer, which led to a new quantum Boltzmann distribution of the problem Hamiltonian. Their quantum Boltzmann Machine (QBM) was trained by introducing a lower-bound to the log-likelihood to allow effective gradient estimation by sampling. Adachi and Henderson~\cite{adachi2015application} trained a deep belief net (DBN) on a 512-qubit D-Wave machine with 8 faulty qubits. The input size is limited to 32 and classification is achieved by adding a downstream classifier. Dorband~\cite{dorband2015boltzmann} implemented a multi-layer semi-restricted BM on a D-Wave quantum annealer, where they alleviate the qubit limitation by implementing a virtual quantum annealer that was configured and run separately from the actual QPU.

As mentioned in Sec.~\ref{sec:contributions}, our work employs quantum annealer as a sampling engine. However, unlike previous works that subscribe to this usage paradigm~\cite{benedetti2016estimation, amin2018quantum, raymond2016global, benedetti2017quantum, dorband2015boltzmann, adachi2015application}, we use quantum sampling to train an MLP, which is a non-generative model usually used for supervised learning.

\section{Equivalence between MLP and EBM}\label{sec:duality}

In this section, we establish an equivalence between the training optimisation problem of the following models:
\begin{itemize}[leftmargin=1em,itemsep=0pt,parsep=0pt,topsep=2pt]
\item MLP with sigmoid activation, trained for a classification task with a binary cross-entropy loss.
\item EBM with binary stochastic units and quadratic energy, trained by maximising the log-likelihood of generating the correct label, conditioned on the input data.
\end{itemize}

The model structures are shown in Fig.~\ref{models}. Later, Sec.~\ref{sampler} will discuss quantum sampling to train EBMs.

\subsection{Main idea}\label{approach}

We show that the models are ``dual'' up to first-order approximations, in the sense that training one implies training the other, and the optimised weights of one can be transferred to the other without changing the decisions. The strategy is to show that by assuming small weights, the gradient of the log-likelihood of conditional probability of the EBM equals the gradient of the binary cross entropy of the MLP. The rest of this section will mathematically prove this equivalence, while Sec.~\ref{experiments} will empirically verify it.

\subsection{EBM}\label{duality}

An EBM with one hidden layer defines the energy
\begin{equation}\label{eq:energy}
E(\bx,\bk,\by) = \bk^T{\bW^{(1)}}\bx +\by^T{\bW^{(2)}}\bk,
\end{equation}
where $\bx$ is a vector of $N$ inputs and $\by$ is a vector of $M$ outputs. Together, $\bx$ and $\by$ make up the visible layer. The hidden layer $\bk$ is a vector of $K$ features. We consider only inputs $\bx$ that are binary (taking values of $0$ or $1$).
The weights $\bW^{(1)}$ and $\bW^{(2)}$ are continuous. $\by$ and $\bk$ are also constrained to be binary. This energy corresponds to the physical energy of the system (for a correct temperature) for a state defined by ($\bx$, $\bk$, $\by$).

To simplify exposition, we omit the biases from~\eqref{eq:energy}, though biases can be included without affecting our results; see Appendix \ref{sec:appendix_equivalence} for details. Also, our derivations hold for $M \ge 1$ output units, though in our experiments (Sec.~\ref{experiments}) we tested only on cases with $M = 1$.

From~\eqref{eq:energy}, there are $2^{N+M+K}$ possible states and the probability of a specific state $(\bx,\bk,\by)$ is defined as:
\begin{equation}
P(\bx,\bk,\by) = \frac{ e^{-E(\bx,\bk,\by)}}{Z}
\label{eq:boltzmann}
\end{equation}
where the partition function $Z$ is 
\begin{equation}
Z= \sum_{\{\bx,\bk,\by\}}^{} e^{-E(\bx,\bk,\by)},
\label{eq:partition}
\end{equation}
which sums over the $2^{N+M+K}$ possible combinations of $(\bx,\bk,\by)$.
The probability of generating outputs $\by$ and inputs $\bx$ is obtained marginalising $\bk$
\begin{equation}\label{eq:p_xy}
P(\by,\bx) = \sum_{\{\bk\}} P(\by,\bk,\bx),   
\end{equation}
and the probability of generating an input $\bx$ is
\begin{equation}
P(\bx) = \sum_{\{\bk,\by\}} P(\by,\bk,\bx)   .
\label{eq:p_x}
\end{equation}

Given an EBM with trained weights $\bW^{(1)}$ and $\bW^{(2)}$, we predict the outputs (labels) $\by$ for a given set of input values $\bx$ by calculating the conditional probability
\begin{equation}\label{eq:p_y|x}
 P(\by|\bx) = \frac{P(\by,\bx)}{P(\bx)}.
\end{equation}

\subsection{Training an EBM}

We aim to learn the weights from a training dataset $\cD = \{ (\bx_\ell,\by_\ell) \}^{L}_{\ell=1}$. This is achieved by maximising the expectation of the log conditional probability
\begin{equation}\label{eq:expect_log_p}
\mathbb{E}_{\hat{\bx},\hat{\by}} \left[  \log P({\by}|{\bx}) \right],
\end{equation}
where  $\mathbb{E}_{\hat{\bx},\hat{\by}}$ indicates that the expectation is taken over the true underlying joint distribution $P_D$ of $\bx$ and $\by$, \ie, 
\begin{align}
\mathbb{E}_{\hat{\bx},\hat{\by}} \left[  f(\bx,\by) \right] = \sum_{\{\bx,\by\}} f(\bx,\by) P_{D}(\bx,\by)
\end{align}
for an arbitrary $f(\bx,\by)$, which sums over all possible combinations of $\bx,\by$.
Equation~\eqref{eq:expect_log_p} can be expanded as
\begin{equation} \label{eq:p_difference}
\begin{split}
\mathbb{E}_{\hat{\bx},\hat{\by}} \left[\log P(\by|\bx)\right]  &=  \mathbb{E}_{\hat{\bx},\hat{\by}} \left[\log \frac{ P(\by,\bx)}{P(\bx)}    \right] \\
&= \mathbb{E}_{\hat{\bx},\hat{\by}} \left[\log { P(\by,\bx)}    \right] 
- \mathbb{E}_{\hat{\bx},\hat{\by}} \left[\log { P(\bx)}    \right] \\
&= \mathbb{E}_{\hat{\bx},\hat{\by}} \left[\log { P(\by,\bx)}    \right] 
- \mathbb{E}_{\hat{\bx}} \left[\log { P(\bx)}    \right],
\end{split}
\end{equation}
where $\mathbb{E}_{\hat{\bx}}$ is the expectation taken over the true underlying distribution of variable $\bx$ only.

Following CD~\cite[Sec.~1]{carreira2005contrastive}, the partial differentiation of the terms in~\eqref{eq:p_difference} against the $(j,i)$-th element of $\bW^{(1)}$ are
\begin{align}\label{eq:grad1}
\partial_{{W^{(1)}}_{j,i}}   \mathbb{E}_{\hat{\bx},\hat{\by}} {\left[\log { P({\by},{\bx})}    \right]} &=  \mathbb{E}_{\hat{\bx},\hat{\by}} {[ { {k_j} {x}_i}     ]} -  \mathbb{E} {\left[ { {k_j}{x_i}}    \right]}, \\
\label{eq:grad2}
\partial_{{W^{(1)}}_{j,i}}   \mathbb{E}_{\hat{\bx}} {\left[\log { P({\bx})}    \right]} &=  \mathbb{E}_{\hat{\bx}} {[ { {k_j}{x}_i}  ]} -  \mathbb{E} {\left[ { {k_j}{x_i}}    \right]},
\end{align}
where $k_j$ and $x_i$ are the $j$-th and $i$-th element of $\bk$ and $\bx$, $\mathbb{E}$ is the expectation over the model distribution, \ie,
\begin{align}\label{eq:modelexp}
    \mathbb{E} \left[  f(\bx,\by) \right] = \sum_{\{\bx,\by\}} f(\bx,\by) P(\bx,\by),
\end{align}
and $\mathbb{E}_{\hat{\bx}}$ is the expectation over the true distribution of $\bx$
\begin{align}
\mathbb{E}_{\hat{\bx}} \left[  f(\bx,\by) \right] = \sum_{\{\bx,\by\}} f(\bx,\by) P(\by|\bx) P_{D}(\bx)
\end{align}
with the other variable $\by$ marginalised over the conditional model distribution. Note that $k_j$ is conditionally dependent on $\bx$ and $\by$, as per the structure of the EBM (see Fig.~\ref{models}).
Subtracting~\eqref{eq:grad2} from~\eqref{eq:grad1} yields the gradient of our training objective
\begin{equation}
\partial_{{W^{(1)}}_{j,i}}   \mathbb{E}_{\hat{\bx},\hat{\by}} {\left[\log { P({\by}|{\bx})}    \right]} =  \mathbb{E}_{\hat{\bx},\hat{\by}} {[ { {k_j}{x}_i}    ]} -  \mathbb{E}_{\hat{\bx}} {[ { {k_j}{x}_i}    ]}.
\label{eq:grad_log_likelihood}
\end{equation}
In a similar way, the gradient for $\bW^{(2)}$ is obtained as
\begin{equation} 
\label{eq:grad_log_likelihood_W2}
\partial_{{W^{(2)}}_{j,i}}   \mathbb{E}_{\hat{\bx},\hat{\by}} {\left[\log { P({\by}|{\bx})}    \right]} =  \mathbb{E}_{\hat{\bx},\hat{\by}} {\left[ { {y}_j{k_i}}    \right]} -  \mathbb{E}_{\hat{\bx}} {\left[ { {y_j}{k_i}}    \right]}.
\end{equation}
To develop explicit formulae for the above, we use the definitions of the expectations to rewrite
\begin{align}
\mathbb{E}_{\hat{\bx},\hat{\by}} {\left[ { {k_j}{x}_i}    \right]} &
=\mathbb{E}_{\hat{\bx},\hat{\by}}  \left[ \mathbb{E} {\left[ { {k_j}}    \right | {\bx},{\by}]} 
x_i \right],\\
\nonumber 
\mathbb{E}_{\hat{\bx}} {\left[ { {k_j}{x}_i}    \right]} 
&= \mathbb{E}_{\hat{\bx}}  \left[ \mathbb{E} {\left[ { {k_j}}    \right | {\bx}]}
 {x}_i \right]\\
&= \mathbb{E}_{\hat{\bx}}
\left[ 
       \sum_{\{\by\}}
       \left(  \mathbb{E} \left[ 
                                       { {k_j}} | {\bx}, \by 
                                    \right]
                                    {P}{\left( \by | {\bx}  \right)}
       \right)      
 {x}_i \right].
\end{align}
Then, by factoring in the EBM structure, we have
\begin{align} 
\label{eq:first_term_expectation}
\mathbb{E}_{\hat{\bx},\hat{\by}} {\left[ { {k_j}{x}_i}    \right]} &= \mathbb{E}_{\hat{\bx},\hat{\by}}  \left[ x_i  \sigma(\bW^{(1)} { \bx} + {\bW^{(2)}}^T {\by})_j  \right], \\
\label{eq:second_term_expectation}
\mathbb{E}_{\hat{\bx}} {\left[ { {k_j}{x}_i}    \right]} &=
\mathbb{E}_{\hat{\bx}}
\left[ 
x_i \sum_{\{\by\}}
       \left(
       \sigma( \bW^{(1)} \bx + {\bW^{(2)}}^T \by )
           \mathop{{P}}{\left( \by | {\bx}  \right)}
       \right)_j      
  \right], 
\end{align}
where $\sigma(z) = \frac{1}{1+e^{-z}}$ is the logistic function. The expectation~\eqref{eq:first_term_expectation} can be estimated from the dataset $\cD$, while samples from $P(\by|\bx)$ are required in addition to $\cD$ to estimate~\eqref{eq:second_term_expectation}; the standard way to sample is Gibbs sampling, which is replaced by quantum sampling in our method; more details in Sec.~\ref{sampler}. With the computed gradients~\eqref{eq:grad_log_likelihood} and~\eqref{eq:grad_log_likelihood_W2}, the weights $\bW^{(1)}$ and $\bW^{(2)}$ can be iteratively updated using any gradient-based update rule.

\subsection{Connecting EBM to MLP}

To connect EBM to MLP, we begin by taking first-order Taylor expansions of the expectations and activation functions in~\eqref{eq:grad_log_likelihood} and~\eqref{eq:grad_log_likelihood_W2}. Define
\begin{align}
    f(\by) = \mathbb{E}[k_j|{\bx},\by] = \sigma( \bW^{(1)} {\bx} + {\bW^{(2)}}^T \by )_j.
\end{align}
We expand $f(\by)$ over the moments of the distribution of $P ( \by| \bx) $ around $\mathbb{E} \left[ 
                                       { {\by}} | {\bx} 
                                    \right]$:
\begin{equation}\label{eq:taylor}
\mathbb{E}[f(\by)] \approx  f(\mathbb{E}[\by]) + \frac{\partial^2 f(\mathbb{E}[\by])}{2} \text{Var}[\by],
\end{equation}
where the second derivative of $f$ is quadratic in the weights, \ie, the second term of~\eqref{eq:taylor} is of order $O({\bW^{(2)}}^2)$.

We assume that the weights are small, \ie, $|W^{(l)}_{i,j}| \ll 1$ for $l = 1,2$ and for all $(i,j)$. This is justifiable since the weights of BMs are usually initialised to be small random values, \eg, from a Gaussian distribution with zero mean and a standard deviation of 0.01 \cite[Section 8]{hinton2012practical}, and they remain typically smaller than one in absolute value during training. This was observed throughout our experiments.

It follows~\eqref{eq:second_term_expectation} can be approximated as
\begin{equation} \label{eq:second_term_2}
\mathbb{E}_{\hat{\bx}} {\left[ { {k_j}{x}_i}    \right]} 
\approx
 \mathbb{E}_{\hat{\bx}} \left[ 
    x_i   \sigma( \bW^{(1)} {\bx} + {\bW^{(2)}}^T \mathbb{E}[\by|{\bx} ])_j
      \right)
   \biggr],
\end{equation} 
where the second-order term has been removed. We now perform Taylor expansions of the sigmoid function around $\bW^{(1)} {\bx}$ in~\eqref{eq:first_term_expectation}~and~\eqref{eq:second_term_2} as
\begin{equation} 
\begin{split}
\label{eq:first_term_2}
 \mathbb{E}_{\hat{\bx}, \hat{\by}} {\left[ { {k_j}{x}_i}    \right]} 
& \approx \mathbb{E}_{\hat{\bx},\hat{\by}} \left[ x_i \left(
  \sigma( \bW^{(1)} {\bx})+
      {\bW^{(2)}}^T \by   \partial\sigma( \bW^{(1)} {\bx} )
       \right)_j      
  \right] \\
& \approx
 \mathbb{E}_{\hat{\bx}} \left[ 
    x_i
       \sigma( \bW^{(1)} {\bx})_j  \right] +
     \mathbb{E}_{\hat{\bx},\hat{\by}} \left[ x_i \left( {\bW^{(2)}}^T \by   \partial\sigma( \bW^{(1)} {\bx} )
       \right)_j      
 \right]
\end{split}
\end{equation} 
and
\begin{equation}
\begin{split}
\label{eq:second_term_3}
\mathbb{E}_{\hat{\bx}} {\left[ { {k_j}{x}_i}    \right]} 
& \approx 
 \mathbb{E}_{\hat{\bx}} \left[ x_i \left(
       \sigma( \bW^{(1)} {\bx})+
      {\bW^{(2)}}^T \mathbb{E}[\by|{\bx}]   \partial\sigma( \bW^{(1)} {\bx} )
       \right)_j      
 \right] \\
& \approx \mathbb{E}_{\hat{\bx}} \left[ x_i 
       \sigma( \bW^{(1)} {\bx})_j  \right] +
     \mathbb{E}_{\hat{\bx}} \left[ x_i \left( {\bW^{(2)}}^T \mathbb{E}[\by|{\bx}]   \partial\sigma( \bW^{(1)} {\bx} )
       \right)_j      
 \right],
\end{split}
\end{equation}
where in~\eqref{eq:first_term_2} the first expectation can be done on $\hat{\bx}$ alone because the quantity does not depend on $\by$.

Taking Taylor expansion of $\mathbb{E}[ \by|\bx ]$ over the moments of the distribution $P(\bk|\bx)$ and ignoring second order terms of $\bW^{(2)}$ again, we can rewrite
\begin{align}
\nonumber \mathbb{E}[\by|\bx] &= \sum_{\{\bk\}} \mathbb{E}[\by|\bk] P(\bk|\bx) \\  
  &= \sum_{\{\bk\}} \sigma( \bW^{(2)} \bk  ) P(\bk|\bx)
  \approx \sigma( \bW^{(2)} \mathbb{E}[ \bk | \bx ]  ).\label{eq:xxxxx}
  \end{align}
Factoring in the network structure to expand $\mathbb{E}[ \bk | \bx ]$, and further assuming $\|\bW^{(2)}\|_F~\ll~\|\bW^{(1)}\|_F$, which is justified because the second layer has fewer neurons than the first, we can rewrite~\eqref{eq:xxxxx} as
\begin{equation}
\sigma( \bW^{(2)} \sum_{\{ \by \}} ( \sigma( \bW^{(1)} \bx + {\bW^{(2)}}^T \by )P(\by|\bx))) \approx  \sigma( \bW^{(2)}  \sigma( \bW^{(1)} \bx   )).
\end{equation}
We collect all terms and substitute into~\eqref{eq:grad_log_likelihood}. Since the argument of the expectation does not depend on $\by$, we replace the expectation $\mathbb{E}_{\hat{\bx}}$ with $\mathbb{E}_{\hat{\bx} ,\hat{\by}}$ and rewrite~\eqref{eq:grad_log_likelihood} as
\begin{equation}
\small
\label{eq:gradient_EBM}
\partial_{{W^{(1)}}_{j,i}}   \mathbb{E}_{\hat{\bx},\hat{\by}} {\left[\log { P({\by}|{\bx})}    \right]} \approx 
       \mathbb{E}_{\hat{\bx},\hat{\by}}\cdot \\ \left[ x_i \left( 
            \partial\sigma( \bW^{(1)} {\bx} )
             {\bW^{(2)}}^T \left( {\by}  - \sigma( \bW^{(2)} \sigma( \bW^{(1)} {\bx}))
            \right)
       \right)_j      
  \right].
\end{equation}
Similarly, we can rewrite gradient~\eqref{eq:grad_log_likelihood_W2} using the same stpdf; see Appendix \ref{sec:appendix_equivalence} for details.

The output of an MLP with one hidden layer is 
\begin{equation}
\bz(\bx) = \sigma( \bW^{(2)} \sigma( \bW^{(1)} {\bx})).
\end{equation}
The standard sigmoid cross-entropy loss between ground-truth label $\by$ and output $\bz$ is
\begin{equation}
 \mathcal{L}({\by}, \bz) = -{\by}\log{(\bz)} - (1-{\by})\log{(1-\bz)}
\end{equation}
The gradient of the loss for $W^{(1)}_{j,i}$ is
\begin{equation}
\begin{split}
\label{eq:gradient_MLP}
\partial_{{W^{(1)}}_{j,i}} \mathcal{L}{(\by, \bz)} &= 
 \partial_{{W^{(1)}}_{j,i}} \mathcal{L}({\by}, \sigma( \bW^{(2)} \sigma( \bW^{(1)} {\bx})) ) \\ 
 &= x_i 
       \left( 
            \partial\sigma( \bW^{(1)} {\bx} )
             {\bW^{(2)}}^T \left( {\by}  - \sigma( \bW^{(2)} \sigma( \bW^{(1)} {\bx}))
            \right)
       \right)_j.
\end{split}
\end{equation}
Taking the average of~\eqref{eq:gradient_MLP} over the dataset $\cD$ yields~\eqref{eq:gradient_EBM}. In a similar way, we can equate the gradient of $\mathcal{L}$ for $W^{(2)}_{j,i}$ with gradient of the expectation of the conditional log-likelihood w.r.t.~$W^{(2)}_{j,i}$; see Appendix \ref{sec:appendix_equivalence} for details.

Typical image datasets have continuous pixel values. The relations derived in the current section are also valid for $\bx$ that take continuous values by generalizing the discrete probability distributions to a continuous domain for the variables $\bx$.   
In the next section, we show experiments where the input $\bx$ are images.

\begin{figure*}[ht]\centering
\hspace{-10pt}
     \begin{subfigure}{8cm}
         \centering
         \includegraphics[width=8cm]{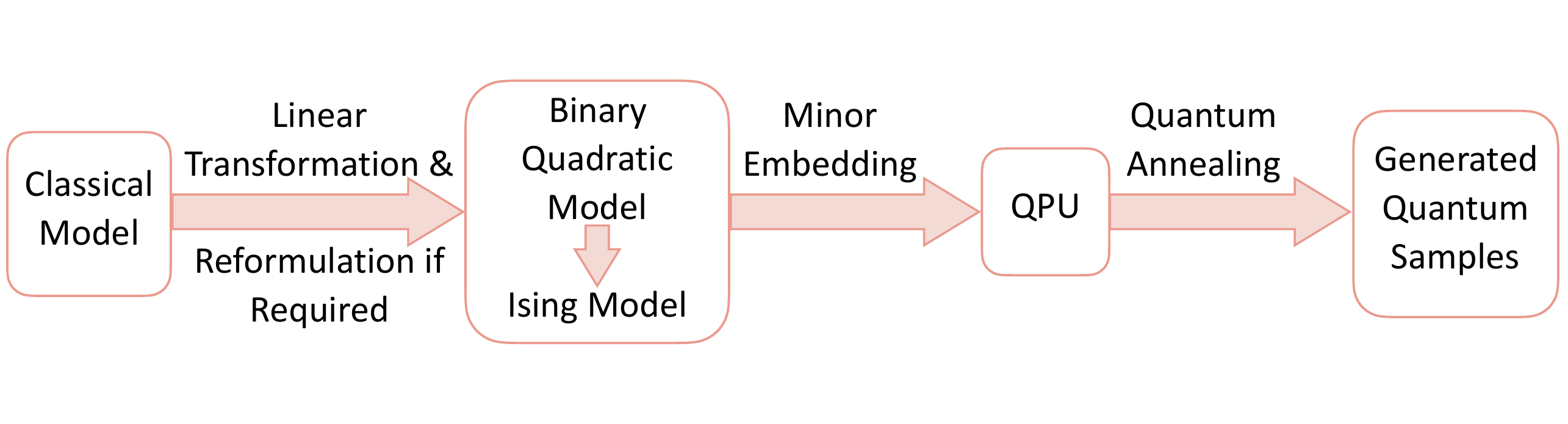}
     \end{subfigure}
     \hspace{5pt}
     \begin{subfigure}{5cm}
         \centering
         \includegraphics[width=5cm]{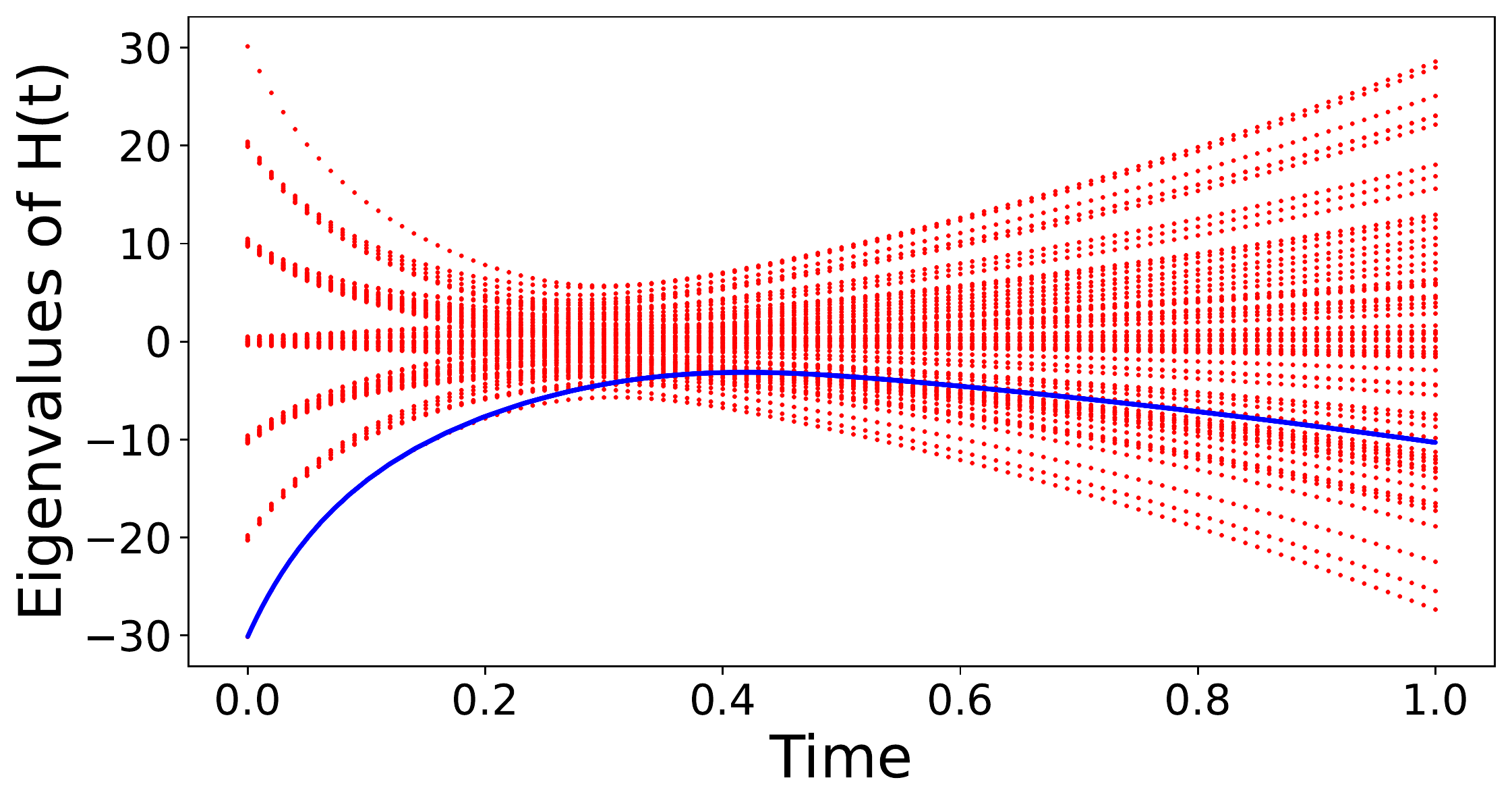}
     \end{subfigure}  
     \vspace{-10pt}
\caption{\textbf{Left}: The quantum sampling process starts from reformulating the classical model to the Ising model through BQM. Minor embedding brings the process onto the hardware. Sampling is then conducted by measuring the output state of the process. \textbf{Right}: Quantum annealing schedule. Time is normalised from the annealing time, which is 20$\mu s$ by default in the D-Wave system.}\vspace{-10pt}
\label{flow}
\end{figure*}

\section{Training EBM with quantum sampling}
\label{sampler}
In this section, we sketch how quantum annealers can be employed to generate samples from EBMs, which in turn can be used to compute the gradient of the expectation of log-likelihood. The equivalence of the gradients of our EBM and MLP established in Sec.~\ref{sec:duality} thus implies that quantum sampling can used to train the MLP.

\subsection{Quantum sampling preliminaries}\label{sec:prelim}
Previous works show that QA can generate samples from Boltzmann-like distributions for training BMs~\cite{adachi2015application, dixit2021training}. Fig.~\ref{flow} (left) shows the workflow of sampling on a quantum annealer; more details as follows.

The evolution of QA in an $n$ qubit system is summarised by the time-dependent Hamiltonian, which in the D-Wave system is defined as
\begin{equation}\label{eq:overallH}
    \bH(t) = -\frac{A(t)}{2}\sum_{i=1}^n \hat{\bsigma}_x^{(i)}  + \frac{B(t)}{2}\left(\sum_{i=1}^n h_i \hat{\bsigma}_z^{(i)}\right. + \left.\sum_{i>j}J_{i,j} \hat{\bsigma}_z^{(i)}\hat{\bsigma}_z^{(j)}\right),
\end{equation}
where $\hat{\bsigma}_x^{(i)}$ and $\hat{\bsigma}_z^{(i)}$ are Pauli matrices operating on qubits with index $i$ or $j$ with all other positions being the identity matrix and outer products between them. The dimensions of $\bH(t)$ is $2^n \times 2^n$; see \cite{albash2018adiabatic} for details.

Note that $\bH(t)$ is the sum of two terms---the initial Hamiltonian and the final Hamiltonian. As the annealing functions $A(t)$ and $B(t)$ change with time, the energy of the system moves towards the final Hamiltonian.

The final Hamiltonian derives from the model of interest, which is encoded as a binary quadratic model (BQM)
\begin{equation} \label{bqm}
    E(\bq) = \sum_i Q_{i,i}q_i + \sum_{i<j} Q_{i,j}q_i q_j = \mathbf{\bq}^T \bQ \mathbf{\bq},
\end{equation}
where $\bq \in \{0,1\}^n$ and $q_i$ is the $i$-th element of $\bq$, and $\bQ$ is an $n 
\times n$ upper triangular matrix with $Q_{i, i}$ being linear coefficients and the non-zero off-diagonal terms, and $Q_{i,j}$ being quadratic coefficients; Sec.~\ref{sec:reformulation} will define $\bQ$ for the model of interest. By substituting
\begin{equation}
q_i = (s_i+1)/2
\end{equation}
into \eqref{bqm}, the BQM is converted to a model of spin variables, which is the Ising model with parameters $\mathbf{J}$ and $\mathbf{h}$
\begin{equation}\label{ising}
E'(\bs) = -\sum_{i} h_i s_i-\sum_{i}\sum_{j<i}J_{ij} s_i s_j,
\end{equation}
where $\bs \in \{-1,1\}^n$, $h_i$ encodes the qubit bias of $s_i$ and has a range of $[-2.0,\; 2.0]$, and
$J_{ij}$ is the coupling strength between $s_i$ and $s_j$ with a range $[-1.0, \;1.0]$. The precision of the control parameters depends on the QPU. After minor embedding,~\eqref{ising} becomes the final Hamiltonian.

The QA process is depicted on the right of Fig.~\ref{flow}, where the red dots are eigenvalues of the Hamiltonian, whose eigenstates compose a certain quantum state. The blue dots are values of the Hamiltonian calculated for the quantum state at each time step, which is obtained from integrating the Schr\"{o}dinger's equation~\cite{albash2018adiabatic}. The blue dots can be interpreted as expectations of energy corresponding to a certain quantum state and Hamiltonian, \ie, sums of red dots at the same time instance times their corresponding probabilities. For sampling, we are interested in generating the target probability distribution that makes the blue dot at the final time instance by feeding in optimal control parameters including the qubit biases $\bh$, coupling strength $\bJ$, and effective temperature $\beta_{eff}$. The output of the quantum annealer are the samples from the probability distribution that makes the final blue dot. 

Quantum annealers can sample models with complex correlations, which is computationally prohibitive on classical computers. In D-Wave Advantage, one qubit can have a maximum of 15 connections with other qubits, which is a richer connectivity than previous annealers.
However, QA can be affected by noise from the external environment such as temperature and magnetic fields~\cite{gardas2018defects}. The hardware noise complicates the process of inferring the instance-dependent effective temperature, which is vital for effective training of the probabilistic model.

\subsection{BQM for sampling conditional probability}\label{sec:reformulation}

To sample from $P(\by|\bx)$ of the EBM using QA, we need to construct the corresponding BQM. Let the biases in the hidden and output layers be $\mathbf{b}$ and $\mathbf{c}$. 
The linear coefficients of the BQM are thus
\begin{equation}\label{quantum_bias}
    \mathbf{b}_q = \mathbf{\bW^{(1)}} \mathbf{x} + \mathbf{b} \in \mathbb{R}^{K} \;\;\;\; \textrm{and} \;\;\;\; \mathbf{c} \in \mathbb{R}^{M}, 
\end{equation}
where $\bx$ is constant (clamped to a specific input). Defining
\begin{equation}
    \mathbf{B} = diag(\mathbf{b}_q) \;\;\; \textrm{and} \;\;\;\; \mathbf{C} = diag(\mathbf{c}),
\end{equation}
the BQM coefficient matrix $\bQ$ is thus
\begin{equation}\label{eq:bigQ}
    \mathbf{Q} = \frac{1}{\beta_{eff}}
    \begin{bmatrix}
    \mathbf{B} & \mathbf{W}^{(2)}\\
    \mathbf{0} & \mathbf{C}
    \end{bmatrix},
\end{equation}
which is of size $(K+M)\times(K+M)$. $\beta_{eff}$ at this point remains undetermined. Note that the size of $\mathbf{Q}$ does not depend on the dimensionality of the input $\bx$. This is in stark contrast to Sasdelli and Chin~\cite{sasdelli21quantum}, where their QUBO to train BNNs scales with the feature dimensions \emph{and} number of connections/weights. Embedding weights as continuous coupling strength $J_{ij}$ allows bigger models to fit on quantum hardware. Additionally, we can do batch training as we do not require embedding all the data in the QUBO.

\subsection{Training algorithm}\label{sec:gradest}
Given a batch of training data $\mathcal{T} \subset \cD$, we wish to compute the gradients \eqref{eq:grad_log_likelihood} and \eqref{eq:grad_log_likelihood_W2} to update the weights. This is achieved via quantum sampling as follows:
\begin{enumerate}[noitemsep,nolistsep]
  \item Compute~\eqref{eq:first_term_expectation} by taking the average over $\mathcal{T}$.
  \item For each $(\bx^\ell,\by^\ell) \in \mathcal{T}$
  \begin{enumerate}
  \vspace{-5pt}
      \item Fix $\bx = \bx^\ell$.
      \item Calculate $\mathbf{Q}$~\eqref{eq:bigQ} with an appropriate $\beta_{eff}$.
      \item Convert BQM~\eqref{bqm} to Ising model~\eqref{ising}.
      \item $\mathcal{S} \gets \mathcal{S} \cup \{ \textrm{$\Gamma$ samples from QA~\eqref{ising}} \}$.
  \end{enumerate}
  \item $\mathcal{S}$ contains samples of both $\bk$ and $\by$; let $\{ \tilde{\by}^r \}^{R}_{r=1}$ be the samples of $\by$. The expectation in \eqref{eq:second_term_expectation} is estimated to $x^\ell_i \sum_{r = 1}^{R}
       \left(
       \sigma( \bW^{(1)} \bx^\ell + {\bW^{(2)}}^T \tilde{\by}^r )\right)_j/R$.
  \item Calculate the gradient as in~\eqref{eq:grad_log_likelihood} and \eqref{eq:grad_log_likelihood_W2}.
\end{enumerate}
Throughout our experiments, we set $\Gamma =$ 1k. The gradient is then utilised in the ADAM optimiser for weight updating.

\section{Experiments}\label{experiments}

\subsection{Equivalence between MLP and EBM}\label{sec:equivexp}

We first verify the equivalence established in Sec.~\ref{sec:duality}.
\vspace{-12pt}
\paragraph{Datasets}

We utilised the MNIST dataset consisting of handwritten digits of 28$\times$28 pixels. Each test case involved random selection of two digits, which were used to create a binary classification task with all training and test data from the chosen classes in MNIST, \ie, approximately 12k training images and 2k test images both with balanced classes.

\vspace{-12pt}
\paragraph{Network size and training}
In both Sec.~\ref{sec:equivexp} and \ref{sec:quantum}, we created an MLP/EBM with size 784$\times$548$\times$1, \ie, $N = 784$ (image size), $K=548$ and $M = 1$. Small network sizes were used to stay within the qubit count of the QPU.

The MLP was trained with backpropagation (specifically ADAM optimiser) with cross-entropy loss, while the EBM was trained with CD~\cite{carreira2005contrastive} to maximise log conditional likelihood~\eqref{eq:p_difference}. The weights of the EBM were initialised from a Gaussian distribution with a zero-mean and standard deviation of 0.01 \cite{hinton2012practical}. Otherwise, the weights of MLP were initialised with the standard initialisation from PyTorch. For both training algorithms, the batch size was 1k, while the learning rate was 0.001.
\vspace{-12pt}
\paragraph{Results}
For each network, we monitored the training loss and test accuracy during training. For each network, we also evaluated the metrics above with the weights inherited from the other network. Specifically, we evaluated
\begin{itemize}[leftmargin=1em,itemsep=0pt,parsep=0pt,topsep=2pt]
    \item The cross-entropy loss and test accuracy for the MLP with weights inherited from the EBM.
    \item The log conditional likelihood and test accuracy for the EBM with weights inherited from the MLP.
\end{itemize}

Additionally, the symmetrised KL divergence measured in nats \cite{kullback1951information} is calculated between the sigmoidal output of the MLP with weights trained by backpropagation and that of the MLP that has inherited weights from the EBM.

\begin{figure}[ht]
     \centering
     \begin{subfigure}{6.5cm}
         \centering
         \includegraphics[width=6.5cm]{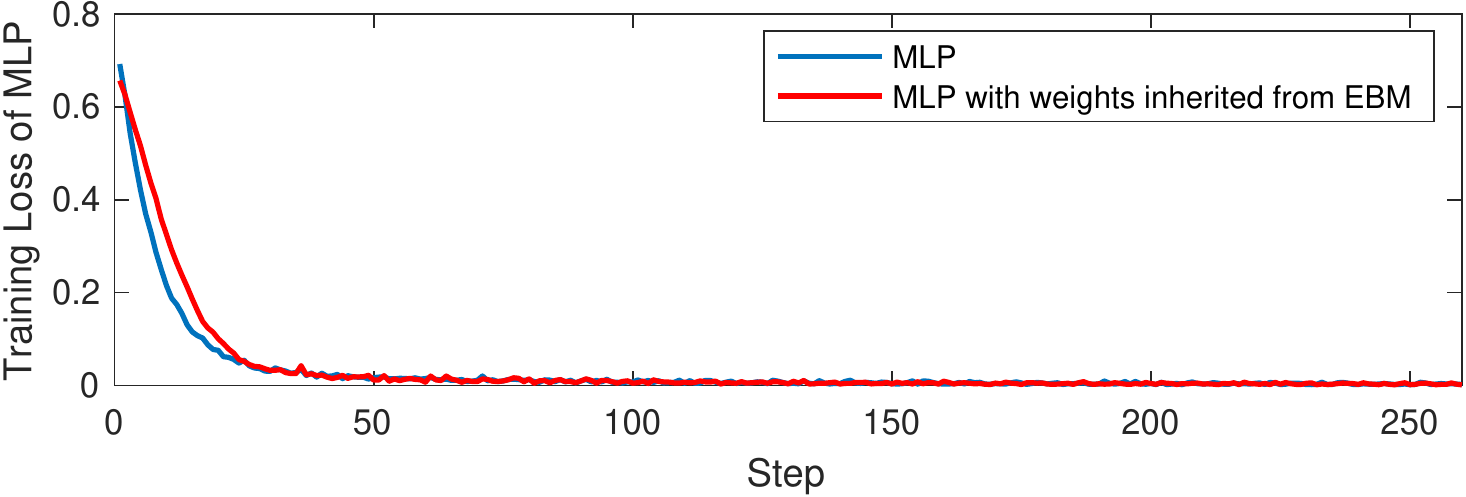}
         \label{loss_ff}
     \end{subfigure}
     \begin{subfigure}{6.5cm}
         \centering
         \includegraphics[width=6.5cm]{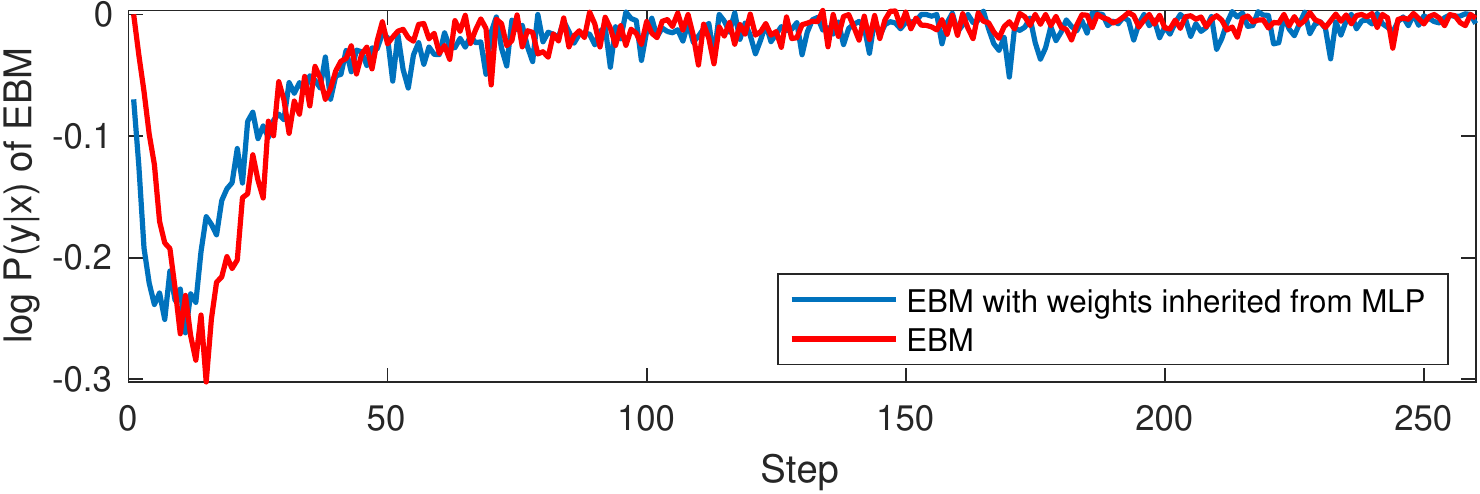}
         \label{loss_ebm}
     \end{subfigure}\\
     \begin{subfigure}{6.5cm}
         \centering
         \includegraphics[width=6.5cm]{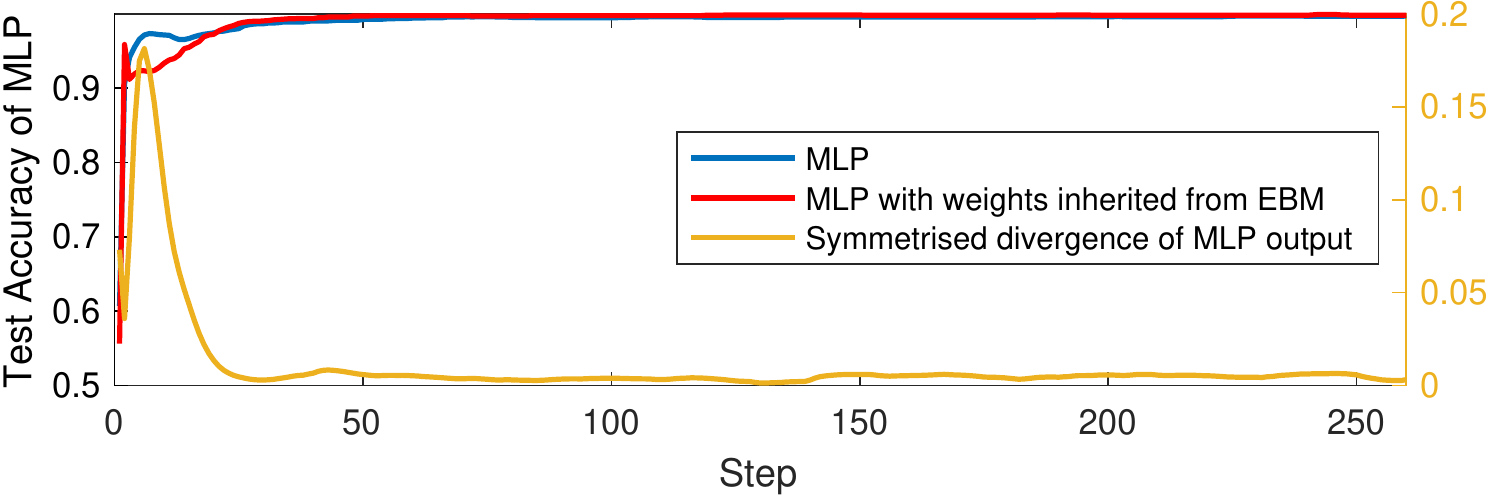}
         \label{test_acc_ff}
     \end{subfigure}
      \begin{subfigure}{6.5cm}
         \centering
         \includegraphics[width=6.5cm]{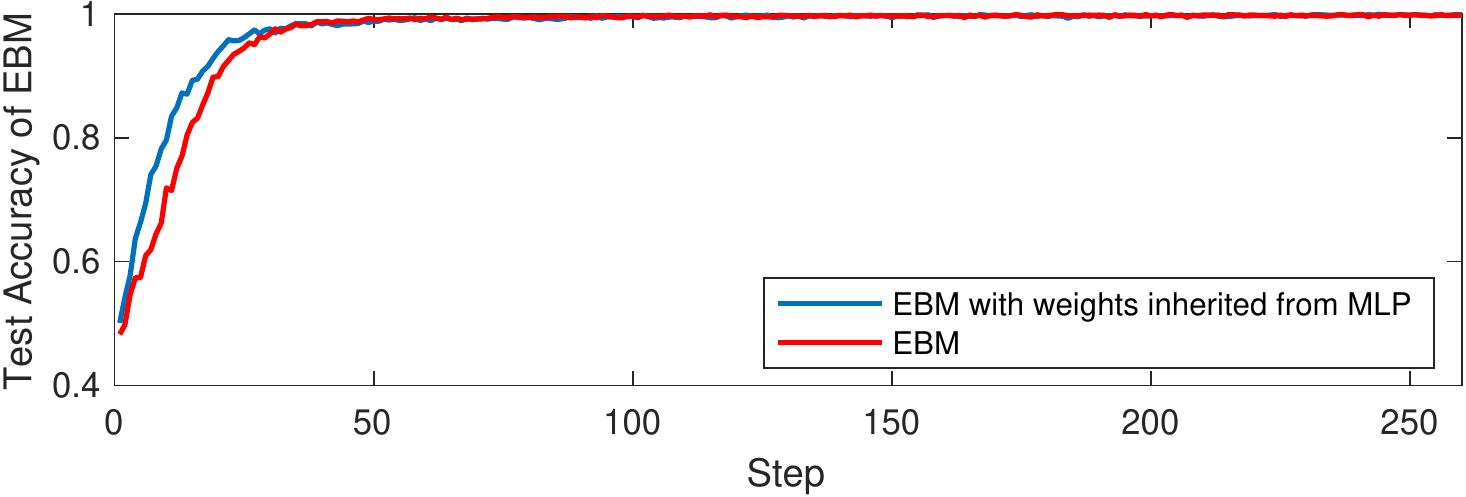}\vspace{-1pt}
         \label{test_acc_ebm}
     \end{subfigure}
     \vspace{-10pt}
        \caption{Equivalence of MLP and EBM on MNIST binary classification. The top panels show the cross-entropy loss and log conditional likelihood (trained and evaluated). The bottom panels show the test accuracy. Symmetrised KL divergence is shown in the bottom left panel.}\vspace{-10pt}
        \label{verify_dual}
\end{figure}

Figs.~\ref{verify_dual} shows the case of classification between digits 0 and 1. The metrics are highly similar across the training iterations, and the output divergence measured with MLP approaches zero, which supports the equivalence of the two training problems established in Sec.~\ref{sec:duality}. Refer to Appendix \ref{app:equimodels} for evaluation on more digit pairs.

\subsection{Training MLP with quantum sampling}\label{sec:quantum}
We now present results for the algorithm in Sec.~\ref{sec:gradest} with sampling conducted by a D-Wave Advantage quantum annealer.
Note that besides limited QPU capacity, our access time was also restricted. \emph{The results presented here have fully exhausted our available quantum compute quota.}

\vspace{-12pt}
\paragraph{Datasets}
We experimented with randomly generated binary classification problems on both MNIST and Fashion-MNIST datasets. We trained on random subsets of the training data, specifically 20 training images for each case, and test data was generated in the same way as Sec.~\ref{sec:equivexp}. The classes were balanced in both sets.

The following methods and settings were executed:
\begin{itemize}[leftmargin=1em,itemsep=0pt,parsep=0pt,topsep=2pt]
    \item \textbf{Classical-1:} MLP with backpropagation (ADAM optimiser) to minimise cross-entropy loss. For both datasets, a batch size of 5 and a learning rate of 0.1 were used.
    \item \textbf{Classical-2:} EBM trained with the algorithm in Sec.~\ref{sec:gradest} but with Gibbs sampling (classical) to generate the samples. Specifically, an RBM \cite{hinton2012practical} was created from $\bk$ and $\by$ in the EBM. For both datasets, a batch size of 5 and a learning rate of 0.1 were used. 
    \item \textbf{Quantum-1:} EBM trained with the algorithm in Sec.~\ref{sec:gradest} with quantum sampling. The other settings were the same as in Classical-2. The qubit biases $\bh$ are capped according to the hardware constraints mentioned in Sec. \ref{sec:prelim}.
\end{itemize}

In Classical-2 and Quantum-1, since our focus is on training MLP, the optimised weights for the EBM were transferred to an MLP for test accuracy evaluation.

\begin{figure}
     \raggedright 
     \begin{subfigure}{14cm}
     \centering
         \includegraphics[width=.9\columnwidth]{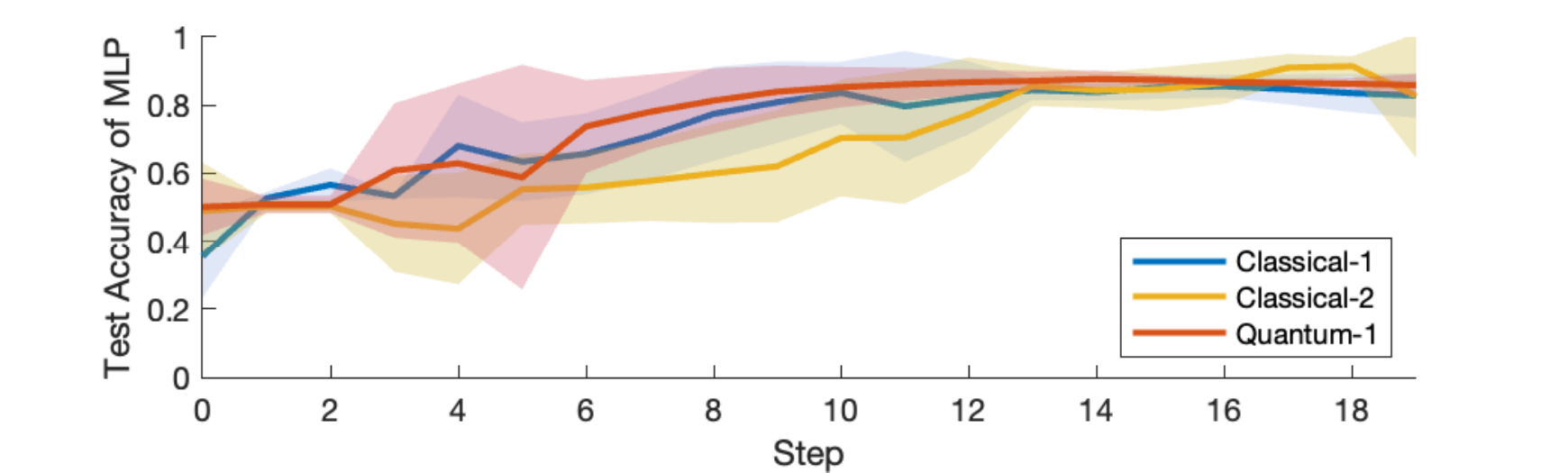}
     \end{subfigure}

      \begin{subfigure}{14cm}
      \centering
         \includegraphics[width=.9\columnwidth]{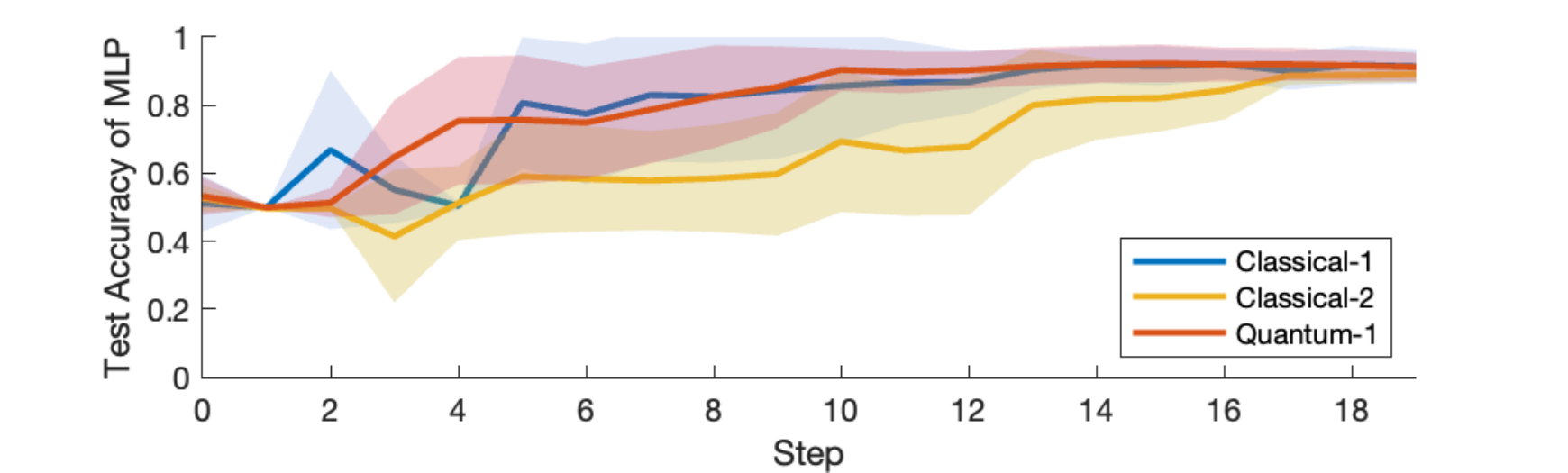}
     \end{subfigure}
     \captionsetup{width=11cm}
       \caption{Avg test accuracy of MLP for binary classification over at least 3 successful runs. The unbiased sample std.~dev.~is shown as the shaded area. The model parameters were trained by MLP (Classical-1), EBM with Gibbs sampling (Classical-2), and EBM with quantum sampling (Quantum-1). \textbf{Top:} Digits 2 \& 3 of MNIST; \textbf{Bottom:} ``Pullover'' \& ``Dress'' of Fashion-MNIST.}\vspace{-5pt}
       \label{quantum_compare}
\end{figure}

\vspace{-12pt}
\paragraph{Results}
For Quantum-1, a grid search was conducted and the effective temperature $\beta$ of 16 was found to be viable for MNIST and Fashion-MNIST. Fig.~\ref{quantum_compare} illustrates the average test accuracy over at least 3 successful trials (with identical setup) of all the methods above, with unbiased sample standard deviation highlighted as the shaded areas. The randomness is due to batch generation, weight initialisation, and the sampling processes involved. Refer to Appendix \ref{app:quantum} for plots on more test cases. Tables~\ref{tab:acc}, \ref{tab:step}, and \ref{tab:success} present results over same sets of experimental trials for 11 digit pairs from MNIST and 5 class pairs from Fashion-MNIST, typically $3$--$16$ trials for each case. The numbers of test cases and trials were restricted by our available quantum compute quota. Table \ref{tab:acc} shows the mean test accuracy. Table \ref{tab:step} reports the number of training stpdf at which the test accuracy surpassed  70$\%$ for the first time. Table \ref{tab:success} displays the success rates of the methods, with successful trials referring to those that exhibited a significant increase in test accuracy during training, while unsuccessful trials display a constant/volatile test accuracy curve. 

In both datasets, Quantum-1 performed successfully; the mean test accuracy of the successful runs is comparable to the classical counterparts across the training iterations. 

\begin{table}
\begin{center}
\scriptsize
\renewcommand{\arraystretch}{.8}
\begin{tabular}{l|cccc}\toprule
Dataset&Class &Classical-1 &Classical-2 &Quantum-1 \\\midrule
\multirow{11}{*}{MNIST} &2-3 & 0.8277 & 0.7742 & {\bf0.8586}\\
&3-9 & 0.9024 & 0.8619 & {\bf0.9050}\\
&5-7 & {\bf0.8921} & 0.8911 & 0.8447\\
&4-8 & 0.7536 & 0.7720 & {\bf0.8247}\\
&5-6 & 0.7656 & 0.7660 & {\bf0.7752}\\
&5-9 & 0.7628 & {\bf0.6937} & 0.6740\\
&0-6 & 0.8572 & {\bf0.8594} & 0.8436\\
&6-8 & 0.9237 & 0.9254 & {\bf0.9346}\\
&0-1 & 0.9965 & 0.9936 & {\bf0.9986}\\
&1-3 & {\bf0.9652} & 0.9288 & 0.9288\\
&8-9 & {\bf0.7376} & 0.6622 & 0.7202\\
\midrule
\multirow{5}{.1\linewidth}{Fashion-MNIST} &Pullover-Dress & {\bf0.9136} & 0.8909 & 0.9116\\
&Dress-Ankle Boot & 0.9917 & 0.9937 & {\bf0.9970}\\
&Sandal-Sneaker & 0.7047 & {\bf0.7142} & 0.7046\\
&Sandal-Shirt &{\bf0.9750} &0.9635 &0.9346\\
&Coat-Bag &{\bf0.9158} &0.8000 &0.8997\\
\bottomrule
\end{tabular}\vspace{-7pt}
\end{center}
\caption{Test accuracy on random cases averaged over at least 3 successful runs.}\vspace{-7pt}
\label{tab:acc}
\end{table}

\begin{table}
\begin{center}
\scriptsize
\renewcommand{\arraystretch}{.8}
\begin{tabular}{l|cccc}\toprule
Dataset&Class &Classical-1 &Classical-2 &Quantum-1 \\\midrule
\multirow{11}{*}{MNIST} &2-3 & 8 & 12 & {\bf7}\\
&3-9 & 7 & 13 & {\bf5}\\
&5-7 & 12 & 18 & {\bf10}\\
&4-8 & {\bf7} & 9 & 9\\
&5-6 & 15 & {\bf9} & {\bf9}\\
&5-9 & {\bf11} & $>$20 & $>$20\\
&0-6 & 12 & 9 & {\bf5}\\
&6-8 & {\bf4} & 5 & 5\\
&0-1 & 8 & {\bf4} & {\bf4}\\
&1-3 & 7 & 10 & {\bf6}\\
&8-9 & 12 & 14 & {\bf8}\\
\midrule
\multirow{5}{.1\linewidth}{Fashion-MNIST} 
&Pullover-Dress & 6 & 14 & {\bf5}\\
&Dress-Ankle Boot & {\bf4} & {\bf4} & 6\\
&Sandal-Sneaker & 11 & 10 & {\bf7}\\
&Sandal-Shirt &{\bf7} &6 &5\\
&Coat-Bag &8 &15 &{\bf6}\\
\bottomrule
\end{tabular}\vspace{-7pt}
\end{center}
\caption{Number of stpdf at which the test accuracy surpasses 70$\%$ for the first time.}\vspace{-15pt}
\label{tab:step}
\end{table}

\begin{table}
\begin{center}
\scriptsize
\renewcommand{\arraystretch}{.8}
\begin{tabular}{l|cccc}\toprule
Dataset&Class &Classical-1 &Classical-2 &Quantum-1 \\\midrule
\multirow{11}{*}{MNIST} &2-3 & {\bf50} & 40 & 28 \\
&3-9 & {\bf80} & 60 & 50\\
&5-7 & {\bf75} & {\bf75} & 40\\
&4-8 & {\bf100} & 50 & 50\\
&5-6 & 50 & {\bf100} & 75\\
&5-9 & 60 & {\bf75} & 50 \\
&0-6 & 60 & {\bf100} & {\bf100}\\
&6-8 & {\bf100} & {\bf100} & 75\\
&0-1 & 100 & 100 & 100\\
&1-3 & {\bf100} & 75 & {\bf100}\\
&8-9 & {\bf44} & 40 & 30\\
\midrule
\multirow{5}{.1\linewidth}{Fashion-MNIST} 
&Pullover-Dress & 20 & 20 & {\bf29}\\
&Dress-Ankle Boot & {\bf67} & 50 & 63\\
&Sandal-Sneaker & 33 & {\bf75} & 50\\
&Sandal-Shirt &50 &{\bf63} &50 \\
&Coat-Bag &{\bf60} &21 &38 \\
\bottomrule
\end{tabular}\vspace{-5pt}
\end{center}
\caption{Success rate in percentage. Typically $3$--$16$ trials were run for each problem.}
\label{tab:success}
\end{table}

\subsection{Runtime analysis}

\begin{figure}
  \centering
  \includegraphics[width=10cm]{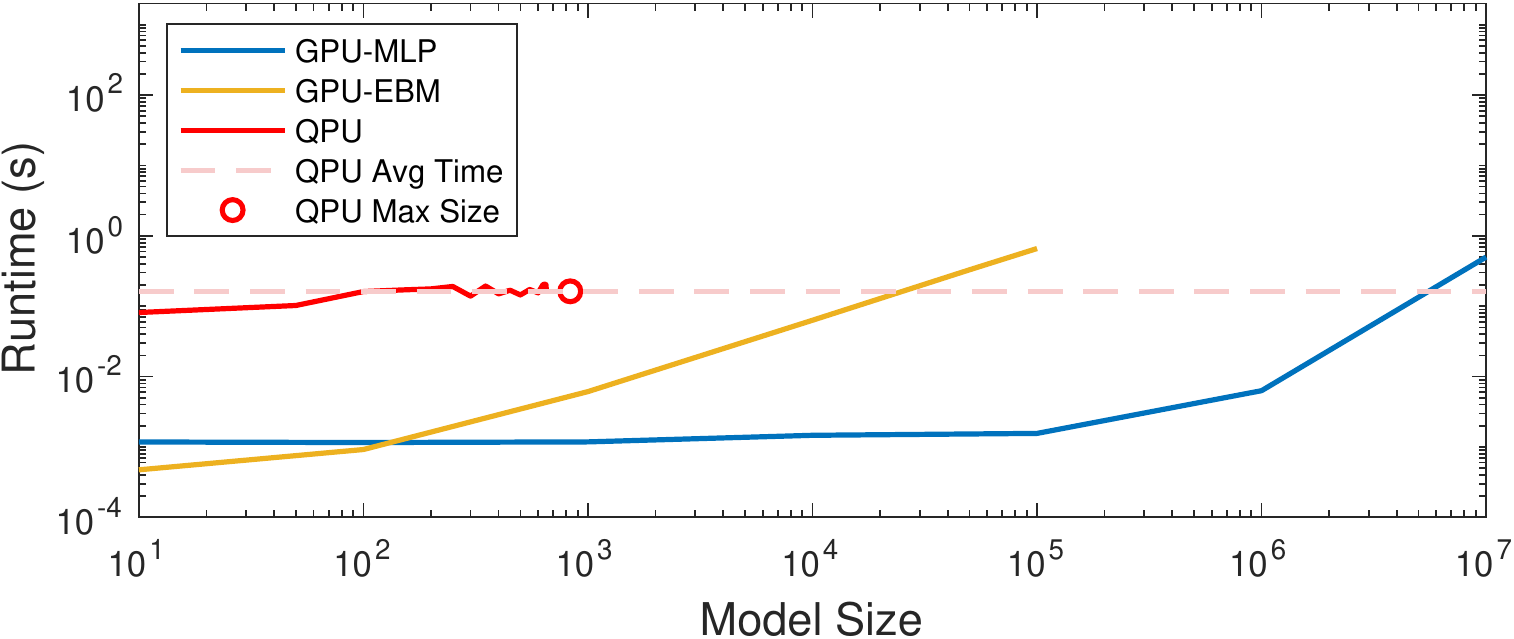}\vspace{-3pt}
  \captionsetup{width=10cm}
  \caption{Comparison of runtime for the three schemes. The model size specifies the number of input/visible nodes, while the number of hidden nodes is always 1.}\vspace{-15pt}
  \label{runtime}
\end{figure}

In this experiment, we are interested to compare the processing time of quantum sampling to the classical training methods. For fairness, a comparison is made between the computational time for matrix multiplication between the hidden and output layer in MLP and the sampling time for the hidden and visible layer in EBM trained with both classical and quantum sampling. The sampling setups are consistent with those in Sec. \ref{experiments}. The classical models were running on the GPU (GeForce RTX 3060) while the quantum sampling was done on the QPU (D-Wave Advantage 4.1 System). To enable a direct comparison, the number of hidden nodes was kept 1 while the number of input/visible nodes was specified as the model size on the x-axis. The results are shown in Fig. \ref{runtime} in a log scale. As the model gets larger, linear growth is observed in the sampling time of both MLP and classical EBM, while that of the quantum annealer fluctuates roughly at the same level. The average processing time for quantum sampling is plotted as the dashed line. Provided the constant annealing time used in our work, \ie, 20$\mu s$, the runtime of the two classical models will eventually exceed that of the quantum-assisted model. The capacity of the Zephyr architecture, which is the latest QPU architecture under development by D-Wave is highlighted as the red dot in Fig. \ref{runtime}. This illustrates great potential for the proposed quantum-assisted training method.

\section{Conclusions}\label{conclusion}

We have presented a method for utilising QA to train MLPs. It is noted that only under certain conditions can a quantum annealer sample from a Boltzmann-like distribution. We found proper hyper-parameters for the quantum method, which achieved comparable performance to backpropagation. Though model sizes were limited on the current QPUs, extensive results on feasible instances illustrate excellent runtime scaling. An important future work is to explore more stable training performance on more complex models through better tuning of the QA control parameters.

\newpage
{\small
\bibliographystyle{ieee_fullname}
\bibliography{iclr2023_conference}

\begin{thebibliography}{10}\itemsep=-1pt

\bibitem{adachi2015application}
Steven~H Adachi and Maxwell~P Henderson.
\newblock Application of quantum annealing to training of deep neural networks.
\newblock {\em arXiv preprint arXiv:1510.06356}, 2015.

\bibitem{albash2018adiabatic}
Tameem Albash and Daniel~A Lidar.
\newblock Adiabatic quantum computation.
\newblock {\em Reviews of Modern Physics}, 90(1):015002, 2018.

\bibitem{albawi2017understanding}
Saad Albawi, Tareq~Abed Mohammed, and Saad Al-Zawi.
\newblock Understanding of a convolutional neural network.
\newblock In {\em 2017 international conference on engineering and technology
  (ICET)}, pages 1--6. Ieee, 2017.

\bibitem{amin2018quantum}
Mohammad~H Amin, Evgeny Andriyash, Jason Rolfe, Bohdan Kulchytskyy, and Roger
  Melko.
\newblock Quantum boltzmann machine.
\newblock {\em Physical Review X}, 8(2):021050, 2018.

\bibitem{beer2020training}
Kerstin Beer, Dmytro Bondarenko, Terry Farrelly, Tobias~J Osborne, Robert
  Salzmann, Daniel Scheiermann, and Ramona Wolf.
\newblock Training deep quantum neural networks.
\newblock {\em Nature communications}, 11(1):1--6, 2020.

\bibitem{benedetti2016estimation}
Marcello Benedetti, John Realpe-G{\'o}mez, Rupak Biswas, and Alejandro
  Perdomo-Ortiz.
\newblock Estimation of effective temperatures in quantum annealers for
  sampling applications: A case study with possible applications in deep
  learning.
\newblock {\em Physical Review A}, 94(2):022308, 2016.

\bibitem{benedetti2017quantum}
Marcello Benedetti, John Realpe-G{\'o}mez, Rupak Biswas, and Alejandro
  Perdomo-Ortiz.
\newblock Quantum-assisted learning of hardware-embedded probabilistic
  graphical models.
\newblock {\em Physical Review X}, 7(4):041052, 2017.

\bibitem{benkner2020adiabatic}
Marcel~Seelbach Benkner, Vladislav Golyanik, Christian Theobalt, and Michael
  Moeller.
\newblock Adiabatic quantum graph matching with permutation matrix constraints.
\newblock In {\em 2020 International Conference on 3D Vision (3DV)}, pages
  583--592. IEEE, 2020.

\bibitem{benkner2021q}
Marcel~Seelbach Benkner, Zorah L{\"a}hner, Vladislav Golyanik, Christof
  Wunderlich, Christian Theobalt, and Michael Moeller.
\newblock Q-match: Iterative shape matching via quantum annealing.
\newblock In {\em Proceedings of the IEEE/CVF International Conference on
  Computer Vision}, pages 7586--7596, 2021.

\bibitem{biamonte2017quantum}
Jacob Biamonte, Peter Wittek, Nicola Pancotti, Patrick Rebentrost, Nathan
  Wiebe, and Seth Lloyd.
\newblock Quantum machine learning.
\newblock {\em Nature}, 549(7671):195--202, 2017.

\bibitem{birdal2021quantum}
Tolga Birdal, Vladislav Golyanik, Christian Theobalt, and Leonidas~J Guibas.
\newblock Quantum permutation synchronization.
\newblock In {\em Proceedings of the IEEE/CVF Conference on Computer Vision and
  Pattern Recognition}, pages 13122--13133, 2021.

\bibitem{carreira2005contrastive}
Miguel~A Carreira-Perpinan and Geoffrey Hinton.
\newblock On contrastive divergence learning.
\newblock In {\em International workshop on artificial intelligence and
  statistics}, pages 33--40. PMLR, 2005.

\bibitem{cormen2022introduction}
Thomas~H Cormen, Charles~E Leiserson, Ronald~L Rivest, and Clifford Stein.
\newblock {\em Introduction to algorithms}.
\newblock MIT press, 2022.

\bibitem{dalal2005histograms}
Navneet Dalal and Bill Triggs.
\newblock Histograms of oriented gradients for human detection.
\newblock In {\em 2005 IEEE computer society conference on computer vision and
  pattern recognition (CVPR'05)}, volume~1, pages 886--893. Ieee, 2005.

\bibitem{dixit2021training}
Vivek Dixit, Raja Selvarajan, Muhammad~A Alam, Travis~S Humble, and Sabre Kais.
\newblock Training restricted boltzmann machines with a {D}-{W}ave quantum
  annealer.
\newblock {\em Front. Phys.}, 2021.

\bibitem{do2016learning}
Thanh-Toan Do, Anh-Dzung Doan, and Ngai-Man Cheung.
\newblock Learning to hash with binary deep neural network.
\newblock In {\em European Conference on Computer Vision}, pages 219--234.
  Springer, 2016.

\bibitem{doan2022hybrid}
Anh-Dzung Doan, Michele Sasdelli, David Suter, and Tat-Jun Chin.
\newblock A hybrid quantum-classical algorithm for robust fitting.
\newblock In {\em Proceedings of the IEEE/CVF Conference on Computer Vision and
  Pattern Recognition}, pages 417--427, 2022.

\bibitem{dorband2015boltzmann}
John~E Dorband.
\newblock A boltzmann machine implementation for the {D}-{W}ave.
\newblock In {\em 2015 12th International Conference on Information
  Technology-New Generations}, pages 703--707. IEEE, 2015.

\bibitem{farhi2000quantum}
Edward Farhi, Jeffrey Goldstone, Sam Gutmann, and Michael Sipser.
\newblock Quantum computation by adiabatic evolution.
\newblock {\em arXiv preprint quant-ph/0001106}, 2000.

\bibitem{gardas2018defects}
Bart{\l}omiej Gardas, Jacek Dziarmaga, Wojciech~H Zurek, and Michael Zwolak.
\newblock Defects in quantum computers.
\newblock {\em Scientific reports}, 8(1):1--10, 2018.

\bibitem{girshick2014rich}
Ross Girshick, Jeff Donahue, Trevor Darrell, and Jitendra Malik.
\newblock Rich feature hierarchies for accurate object detection and semantic
  segmentation.
\newblock In {\em Proceedings of the IEEE conference on computer vision and
  pattern recognition}, pages 580--587, 2014.

\bibitem{golyanik2016gravitational}
Vladislav Golyanik, Sk~Aziz Ali, and Didier Stricker.
\newblock Gravitational approach for point set registration.
\newblock In {\em Proceedings of the IEEE conference on computer vision and
  pattern recognition}, pages 5802--5810, 2016.

\bibitem{hinton2012practical}
Geoffrey~E Hinton.
\newblock A practical guide to training restricted boltzmann machines.
\newblock In {\em Neural networks: Tricks of the trade}, pages 599--619.
  Springer, 2012.

\bibitem{kae2013augmenting}
Andrew Kae, Kihyuk Sohn, Honglak Lee, and Erik Learned-Miller.
\newblock Augmenting crfs with boltzmann machine shape priors for image
  labeling.
\newblock In {\em Proceedings of the IEEE conference on computer vision and
  pattern recognition}, pages 2019--2026, 2013.

\bibitem{kullback1951information}
Solomon Kullback and Richard~A Leibler.
\newblock On information and sufficiency.
\newblock {\em The annals of mathematical statistics}, 22(1):79--86, 1951.

\bibitem{larasati2022trends}
Harashta~Tatimma Larasati, Howon Kim, et~al.
\newblock Trends of quantum computing applications to computer vision.
\newblock In {\em 2022 International Conference on Platform Technology and
  Service (PlatCon)}, pages 7--12. IEEE, 2022.

\bibitem{lecun2006tutorial}
Yann LeCun, Sumit Chopra, Raia Hadsell, M Ranzato, and Fujie Huang.
\newblock A tutorial on energy-based learning.
\newblock {\em Predicting structured data}, 1(0), 2006.

\bibitem{dwave2022advantage}
Catherine McGeoch and Pau Farré.
\newblock Advantage processor overview.
\newblock Technical report, D-Wave, 01 2022.

\bibitem{nie2015generative}
Siqi Nie, Ziheng Wang, and Qiang Ji.
\newblock A generative restricted boltzmann machine based method for
  high-dimensional motion data modeling.
\newblock {\em Computer Vision and Image Understanding}, 136:14--22, 2015.

\bibitem{raymond2016global}
Jack Raymond, Sheir Yarkoni, and Evgeny Andriyash.
\newblock Global warming: Temperature estimation in annealers.
\newblock {\em Frontiers in ICT}, 3:23, 2016.

\bibitem{rumelhart1985learning}
David~E Rumelhart, Geoffrey~E Hinton, and Ronald~J Williams.
\newblock Learning internal representations by error propagation.
\newblock Technical report, California Univ San Diego La Jolla Inst for
  Cognitive Science, 1985.

\bibitem{rumelhart1986learning}
David~E Rumelhart, Geoffrey~E Hinton, and Ronald~J Williams.
\newblock Learning representations by back-propagating errors.
\newblock {\em nature}, 323(6088):533--536, 1986.

\bibitem{sasdelli21quantum}
Michele Sasdelli and Tat-Jun Chin.
\newblock Quantum annealing formulation for binary neural networks.
\newblock In {\em Digital Image Computing: Techniques and Applications
  (DICTA)}, 2021.

\bibitem{schuld2018supervised}
Maria Schuld and Francesco Petruccione.
\newblock {\em Supervised learning with quantum computers}, volume~17.
\newblock Springer, 2018.

\bibitem{schuld2015introduction}
Maria Schuld, Ilya Sinayskiy, and Francesco Petruccione.
\newblock An introduction to quantum machine learning.
\newblock {\em Contemporary Physics}, 56(2):172--185, 2015.

\bibitem{sherstinsky2020fundamentals}
Alex Sherstinsky.
\newblock Fundamentals of recurrent neural network (rnn) and long short-term
  memory (lstm) network.
\newblock {\em Physica D: Nonlinear Phenomena}, 404:132306, 2020.

\bibitem{smolensky1986information}
Paul Smolensky.
\newblock Information processing in dynamical systems: Foundations of harmony
  theory.
\newblock Technical report, Colorado Univ at Boulder Dept of Computer Science,
  1986.

\bibitem{svozil1997introduction}
Daniel Svozil, Vladimir Kvasnicka, and Jiri Pospichal.
\newblock Introduction to multi-layer feed-forward neural networks.
\newblock {\em Chemometrics and intelligent laboratory systems}, 39(1):43--62,
  1997.

\bibitem{taud2018multilayer}
Hind Taud and JF Mas.
\newblock Multilayer perceptron (mlp).
\newblock In {\em Geomatic approaches for modeling land change scenarios},
  pages 451--455. Springer, 2018.

\bibitem{viola2001rapid}
Paul Viola and Michael Jones.
\newblock Rapid object detection using a boosted cascade of simple features.
\newblock In {\em Proceedings of the 2001 IEEE computer society conference on
  computer vision and pattern recognition. CVPR 2001}, volume~1, pages I--I.
  Ieee, 2001.

\bibitem{winci2020path}
Walter Winci, Lorenzo Buffoni, Hossein Sadeghi, Amir Khoshaman, Evgeny
  Andriyash, and Mohammad~H Amin.
\newblock A path towards quantum advantage in training deep generative models
  with quantum annealers.
\newblock {\em Machine Learning: Science and Technology}, 1(4):045028, 2020.

\bibitem{wittek2014quantum}
Peter Wittek.
\newblock {\em Quantum machine learning: what quantum computing means to data
  mining}.
\newblock Academic Press, 2014.

\bibitem{wu2013facial}
Yue Wu, Zuoguan Wang, and Qiang Ji.
\newblock Facial feature tracking under varying facial expressions and face
  poses based on restricted boltzmann machines.
\newblock In {\em Proceedings of the IEEE Conference on Computer Vision and
  Pattern Recognition}, pages 3452--3459, 2013.

\bibitem{zaech2022adiabatic}
Jan-Nico Zaech, Alexander Liniger, Martin Danelljan, Dengxin Dai, and Luc
  Van~Gool.
\newblock Adiabatic quantum computing for multi object tracking.
\newblock In {\em Proceedings of the IEEE/CVF Conference on Computer Vision and
  Pattern Recognition}, pages 8811--8822, 2022.

\end{thebibliography}
}

\newpage
\appendix
\begin{appendices}
\section{Proof of Equivalence}\label{proof}\label{sec:appendix_equivalence}

\paragraph{Biases}

To introduce biases we need to define the energy function as:
\begin{equation}\label{eq:energy_with_bias}
\hat{E}(\bx,\bk,\by) = \bk^T{\bW^{(1)}}\bx +\by^T{\bW^{(2)}}\bk + {\mathbf{b}}^T\bk + {\mathbf{c}}^T\by,
\end{equation}
we define new weights and variables appending one extra dimension to $\bx$ and $\bk$. We constraint these new variables to $1$. 
We can now identify with the bias $\mathbf{b}$ the additional column of $\bW^{(1)}$ and with bias $\mathbf{c}$ the additional row of $\bW^{(2)}$:
\begin{equation}
\hat{\bx} = 
\begin{bmatrix}
x_0 \\
x_1 \\
... \\
x_{N-1} \\
1
\end{bmatrix},
\hat{\bk} = 
\begin{bmatrix}
k_0 \\
k_1 \\
... \\
k_{K-1} \\
1
\end{bmatrix},
\hat{\bW}^{(1)} = 
\begin{bmatrix}
W^{(1)}_{00} & W^{(1)}_{01} & ... & b_0 \\
W^{(1)}_{10} & W^{(1)}_{11} & ... & b_1 \\
... & ... & ... & ... \\
0 & 0 & ... & 0 \\
\end{bmatrix},
\hat{\bW}^{(2)} = 
\begin{bmatrix}
W^{(2)}_{00} & W^{(2)}_{01} & ... \\
W^{(2)}_{10} & W^{(2)}_{11} & ...  \\
... & ... & ...  \\
c_{0} & c_{1} & ...\\
\end{bmatrix}
\end{equation}
we can reduce the case of biases to the simple case without biases:
\begin{equation}\label{eq:energy_without_bias}
\hat{E}(\hat{\bx},\hat{\bk},\hat{\by}) = \hat{\bk}^T{\hat{\bW}^{(1)}}\bx +\hat{\by}^T{\hat{\bW}^{(2)}}\hat{\bk}.
\end{equation}
Notice that we did not introduce biases on the inputs $\bx$, as these are always conditioned to the data distribution.

\paragraph{Derivative of $\bW^{(2)}$}

The gradient for $\bW^{(2)}$  in the EBM is obtained as
\begin{equation} 
\partial_{{W^{(2)}}_{j,i}}   \mathbb{E}_{\hat{\bx},\hat{\by}} {\left[\log { P({\by}|{\bx})}    \right]} =  \mathbb{E}_{\hat{\bx},\hat{\by}} {\left[ { {y}_j{k_i}}    \right]} -  \mathbb{E}_{\hat{\bx}} {\left[ { {y_j}{k_i}}    \right]}
\end{equation}
we then rewrite:
\begin{align}
\mathbb{E}_{\hat{\bx},\hat{\by}} {\left[ { {y_j}{k}_i}    \right]} &
=\mathbb{E}_{\hat{\bx},\hat{\by}}  \left[ \mathbb{E} {\left[ { {y_j k_i}}    \right | {\bx},{\by}]} 
 \right]
 =\mathbb{E}_{\hat{\bx},\hat{\by}}  \left[ y_j \mathbb{E} {\left[ { { k_i}}    \right | {\bx},{\by}]} 
 \right],\\
\nonumber 
\mathbb{E}_{\hat{\bx}} {\left[ { {y_j}{k}_i}    \right]} 
&= \mathbb{E}_{\hat{\bx}}  \left[ \mathbb{E} {\left[ { {y_j k_i}}    \right | {\bx}]}
 \right]\\
&= \mathbb{E}_{\hat{\bx}}
\left[ 
       \sum_{\{\by\}}
       \left(  \mathbb{E} \left[ 
                                       { {y_j k_i}} | {\bx}, \by 
                                    \right]
                                    {P}{\left( \by | {\bx}  \right)}
       \right)      
 \right] \\
&= \mathbb{E}_{\hat{\bx}}
\left[ 
       \sum_{\{\by\}}
       \left(  y_j \mathbb{E} \left[ 
                                       { { k_i}} | {\bx}, \by 
                                    \right]
                                    {P}{\left( \by | {\bx}  \right)}
       \right)      
 \right].
\end{align}
Then, the EBM structure allows us to rewrite:
\begin{align} 
\mathbb{E}_{\hat{\bx},\hat{\by}} {\left[ { {y_j}{k}_i}    \right]} &= \mathbb{E}_{\hat{\bx},\hat{\by}}  \left[ y_j  \sigma(\bW^{(1)} { \bx} + {\bW^{(2)}}^T {\by})_i  \right], \\
\label{eq:second_term_expectation_yk}
\mathbb{E}_{\hat{\bx}} {\left[ { {y_j}{k}_i}    \right]} &=
\mathbb{E}_{\hat{\bx}}
\left[ 
 \sum_{\{\by\}}
       \left( y_j
       \sigma( \bW^{(1)} \bx + {\bW^{(2)}}^T \by )
           \mathop{{P}}{\left( \by | {\bx}  \right)}
       \right)_i      
  \right].
\end{align}
We can estimate the first equation directly from the dataset and the second requires sampling from  $P(\by|\bx)$. 
Like before, we approximate~\eqref{eq:second_term_expectation_yk}  as
\begin{equation} 
\mathbb{E}_{\hat{\bx}} {\left[ { {y_j}{k}_i}    \right]} 
\approx
 \mathbb{E}_{\hat{\bx}} \left[ 
     \mathbb{E}[y_j|{\bx} ]   \sigma( \bW^{(1)} {\bx} + {\bW^{(2)}}^T \mathbb{E}[\by|{\bx} ])_i
      \right)
   \biggr].
\end{equation} 
We now perform Taylor expansions of the sigmoid function around $\bW^{(1)} {\bx}$:
\begin{equation} 
\begin{split}
\mathbb{E}_{\hat{\bx}, \hat{\by}} {\left[ { {y_j}{k}_i}    \right]} 
&\approx \mathbb{E}_{\hat{\bx},\hat{\by}} \left[ y_j \left(
  \sigma( \bW^{(1)} {\bx})+
      {\bW^{(2)}}^T \by   \partial\sigma( \bW^{(1)} {\bx} )
       \right)_i      
  \right] \\
& \approx  \mathbb{E}_{\hat{\bx},\hat{\by}} \left[ 
    y_j
       \sigma( \bW^{(1)} {\bx})_i  \right] +
     \mathbb{E}_{\hat{\bx},\hat{\by}} \left[ y_j \left( {\bW^{(2)}}^T \by   \partial\sigma( \bW^{(1)} {\bx} )
       \right)_i      
 \right]
\end{split}
\end{equation} 
and
\begin{equation}
\begin{split}
\mathbb{E}_{\hat{\bx}} {\left[ { {y_j}{k}_i}    \right]}
&\approx 
 \mathbb{E}_{\hat{\bx}} \left[\mathbb{E}[y_j|{\bx}] \left(
       \sigma( \bW^{(1)} {\bx})+
      {\bW^{(2)}}^T \mathbb{E}[\by|{\bx}]   \partial\sigma( \bW^{(1)} {\bx} )
       \right)_i      
 \right] \\
& \approx \mathbb{E}_{\hat{\bx}} \left[ \mathbb{E}[y_j|{\bx}] 
       \sigma( \bW^{(1)} {\bx})_i  \right] +
     \mathbb{E}_{\hat{\bx}} \left[\mathbb{E}[y_j|{\bx}] \left( {\bW^{(2)}}^T \mathbb{E}[\by|{\bx}]   \partial\sigma( \bW^{(1)} {\bx} )
       \right)_i      
 \right].
\end{split}
\end{equation}

As before, we approximate $\mathbb{E}[\by|\bx] 
\approx  \sigma( \bW^{(2)}  \sigma( \bW^{(1)} \bx   ))
$ and combine everything together.
Differently from the gradient of the log-likelihood with respect to  ${\bW^{(1)}}$, the leading orders do not cancel out. Consequently, we obtain:
\begin{equation}
\partial_{{W^{(2)}}_{j,i}}   \mathbb{E}_{\hat{\bx},\hat{\by}} {\left[\log { P({\by}|{\bx})}    \right]} \approx 
       \mathbb{E}_{\hat{\bx},\hat{\by}}
       \left[ 
        \sigma( \bW^{(1)} {\bx})_i
        \left( 
             {\by}  - \sigma( \bW^{(2)} \sigma( \bW^{(1)} {\bx}))
       \right)_j
  \right],
\end{equation}
The function inside the expectation is equal to the gradient of $\mathcal{L}$ for $W^{(2)}_{j,i}$ in the MLP.

\section{Minor Embedding} \label{app:minor}
\begin{figure}[t!]
\vspace{-20pt}
     \centering
         \includegraphics[width=.7\linewidth]{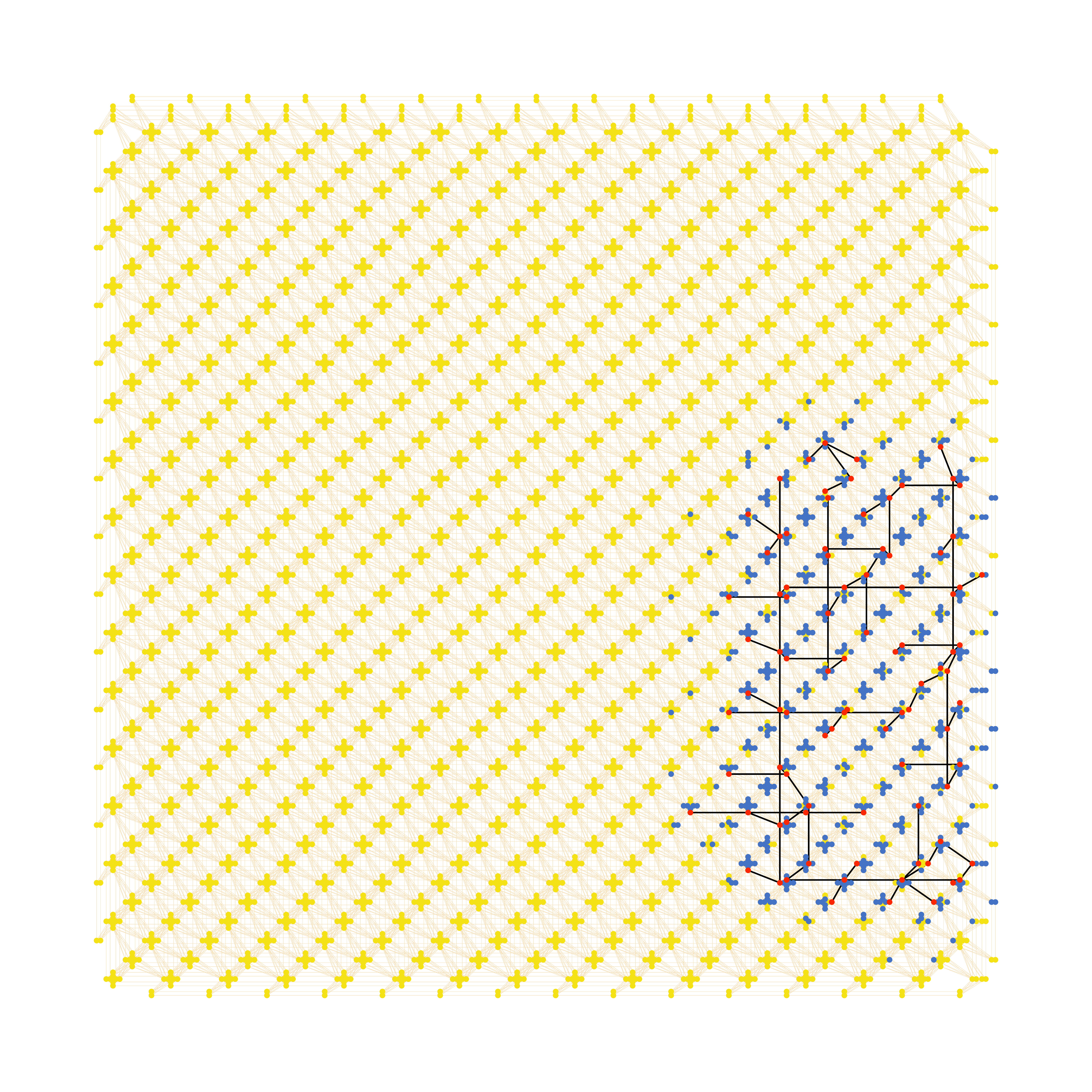}
     \vspace{-20pt}
     \captionsetup{width=8cm}
     \caption{An RBM of size 654$\times$1 embedded in the D-Wave Advantage System with a Pegasus P16 graph. The Visible nodes are highlihgted in blue, the hidden nodes are highlighted in red, and the QPU graph is highlighted in orange (edge) and yellow (vertices). The black lines in the graph shows how the qubit chains are formed for the hidden node (red).}\vspace{20pt}
     \label{embed_pegasus}
\end{figure}

\begin{figure}[t!]
     \centering
     \vspace{-30pt}
         \includegraphics[width=.7\linewidth]{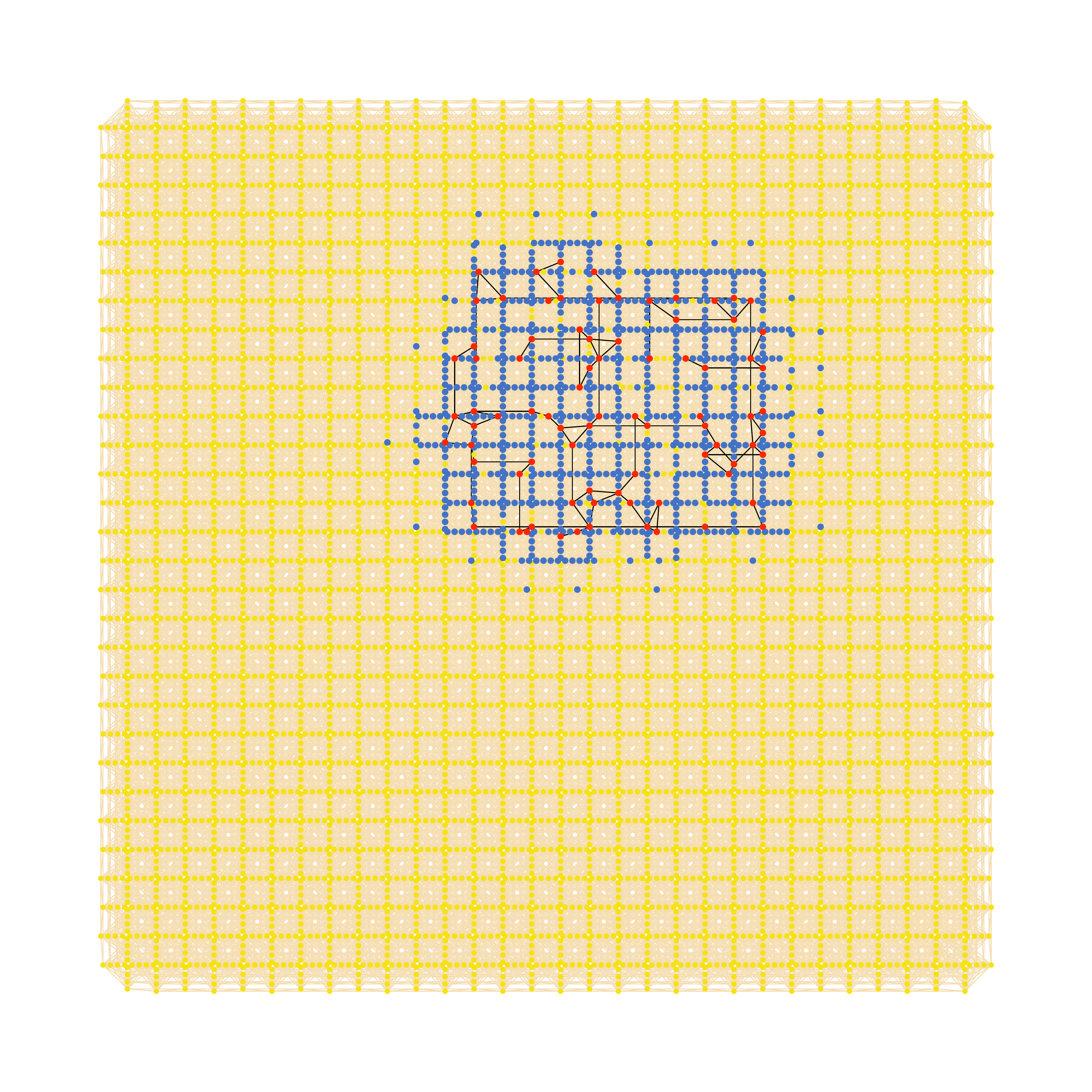}
     \vspace{-20pt}
     \captionsetup{width=8cm}
     \caption{An RBM of size 834$\times$1 embedded in the D-Wave Advantage2 System (under development) with a Zephyr Z15 graph.The Visible nodes are highlihgted in blue, the hidden nodes are highlighted in red, and the QPU graph is highlighted in orange (edge) and yellow (vertices). The black lines in the graph shows how the qubit chains are formed for the hidden node (red).} \vspace{10pt}
     \label{embed_zephyr}
\end{figure}

Fig. \ref{embed_pegasus} and Fig. \ref{embed_zephyr} each plots the embedding of an RBM of maximum number of visible nodes given one single hidden node in the respective D-Wave quantum processing unit (QPU) architecture, \ie, Pegasus P16 graph in the Advantage QPU and Zephyr Z15 in the Advantage2 Prototype. Each vertex in a graph represents a quantum bit, or qubit and each edge represents a coupler between two qubits that enables the interaction. Specifically, the Pegasus graph contains 5760 qubits with 15 couplers per qubit while Zephyr scales up to 7440 qubits with 20 couplers per qubit. However, the number of couplers per qubit is still significantly smaller than the number of connections required by the target models. The physical qubits in the same chain connected by the black lines represent one single logical qubit in the target problem, \ie, the hidden node in this case. The chains help to expand the number of connections allowed for the qubits, which is less restricted to the limited connectivity in the QPU. It is observed that there is still a fair amount of spare qubits in both plots. This is due to full connectivity being exploited for some physical qubit(s). 

\section{Additional Experimental Results}
\subsection{Equivalence between MLP and EBM}\label{app:equimodels}
\begin{figure*}[!t]
\centering
\begin{minipage}[t]{6.5cm}
\vspace{10pt}
    \centering
     \begin{subfigure}[b]{6.5cm}
         \centering
         \includegraphics[width=\linewidth]{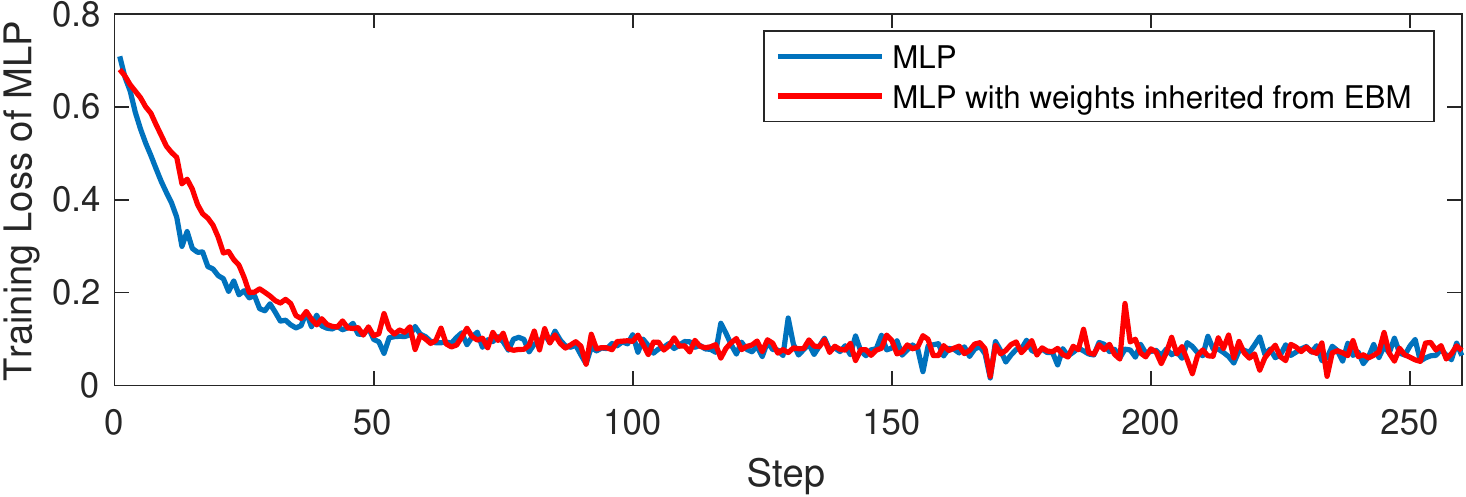}
     \end{subfigure}\\
    \vspace{-5pt}
     \begin{subfigure}[b]{6.5cm}
         \centering
             \includegraphics[width=\textwidth]{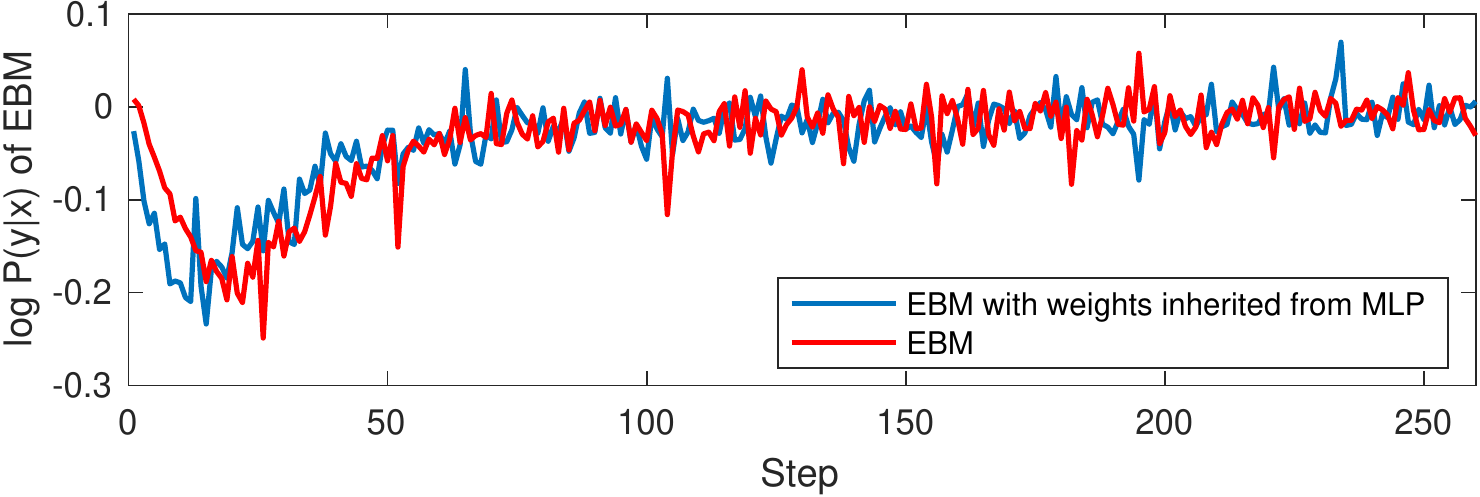}
     \end{subfigure}
      \begin{subfigure}[b]{6.5cm}
         \centering
         \includegraphics[width=\linewidth]{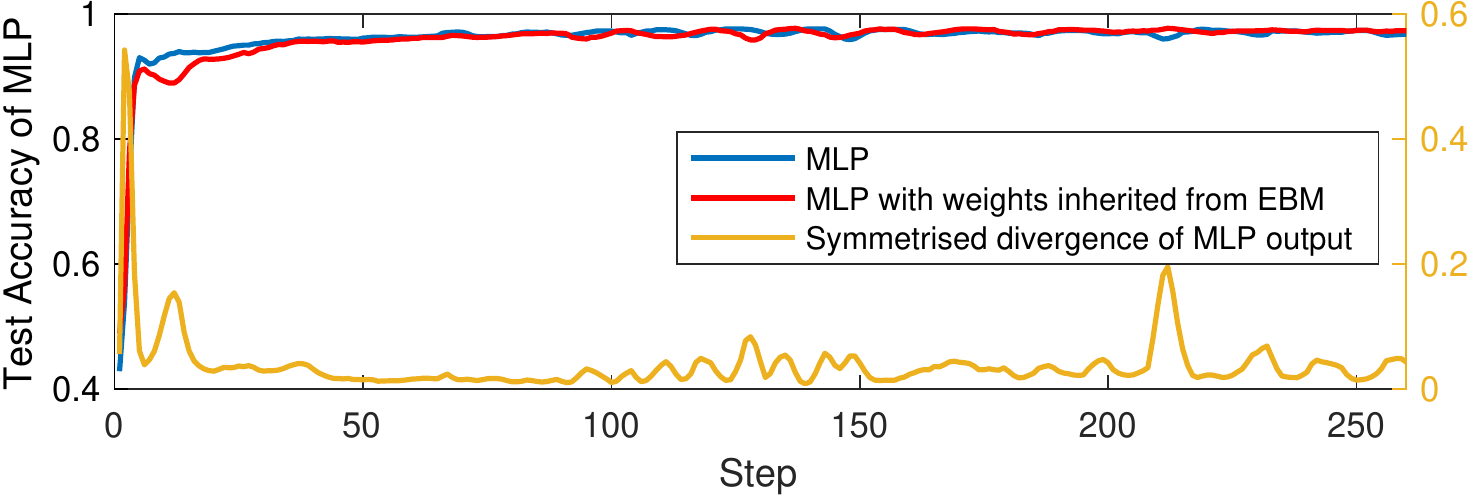}
     \end{subfigure}\\
    \vspace{-5pt}
     \begin{subfigure}[b]{6.5cm}
         \centering
             \includegraphics[width=\textwidth]{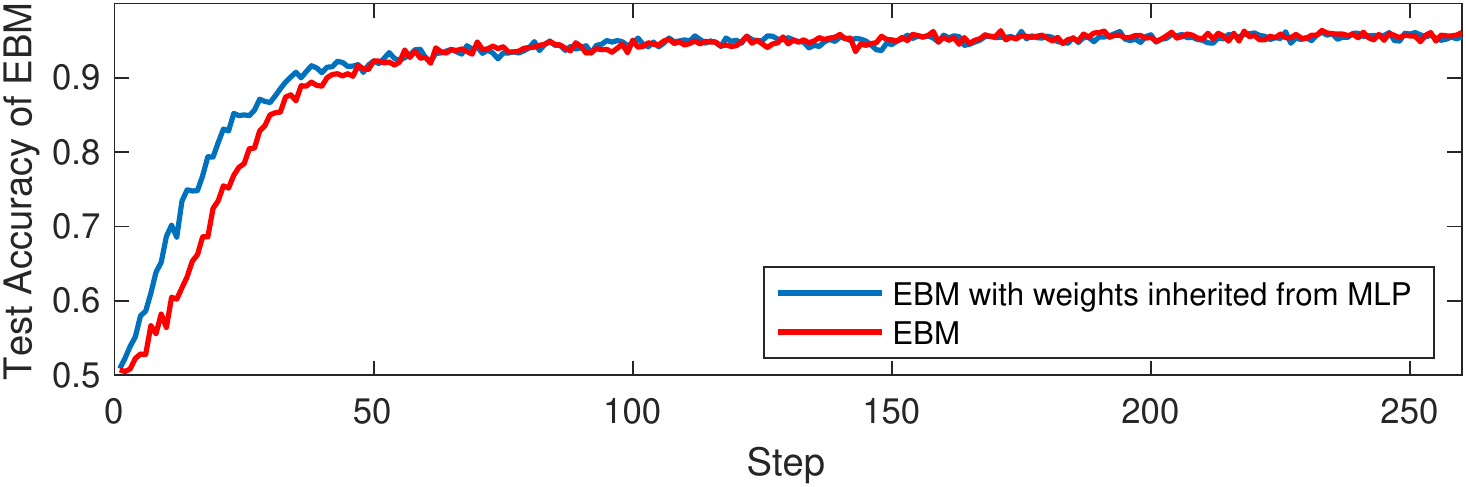}
     \end{subfigure}
    \captionsetup{width=6.5cm}
        \caption{Equivalence of MLP and EBM on MNIST binary classification between digits 2 \& 3. Panels 1 \& 2 show cross-entropy loss and log conditional likelihood (trained and evaluated). Panels 3 \& 4 show the test accuracy. Symmetrised KL divergence is shown in Panel 3.}
    \label{fig:eq_23}
\end{minipage}
\quad
\begin{minipage}[t]{6.5cm}
\vspace{10pt}
    \centering
     \begin{subfigure}[b]{6.5cm}
         \centering
         \includegraphics[width=\linewidth]{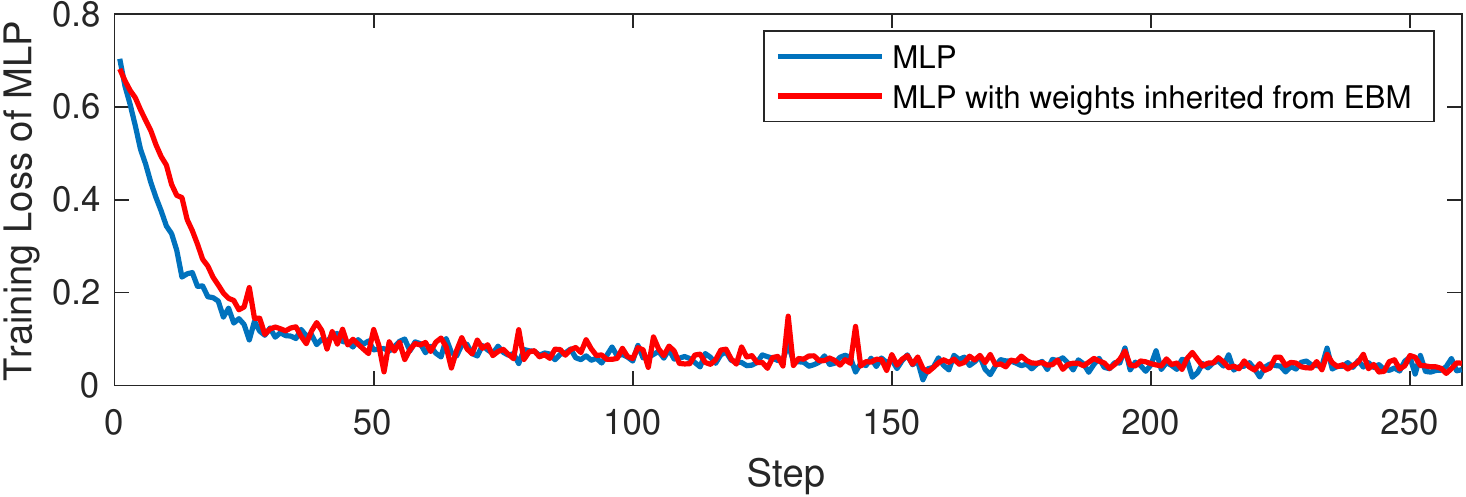}
     \end{subfigure}\\
    \vspace{-5pt}
     \begin{subfigure}[b]{6.5cm}
         \centering
             \includegraphics[width=\textwidth]{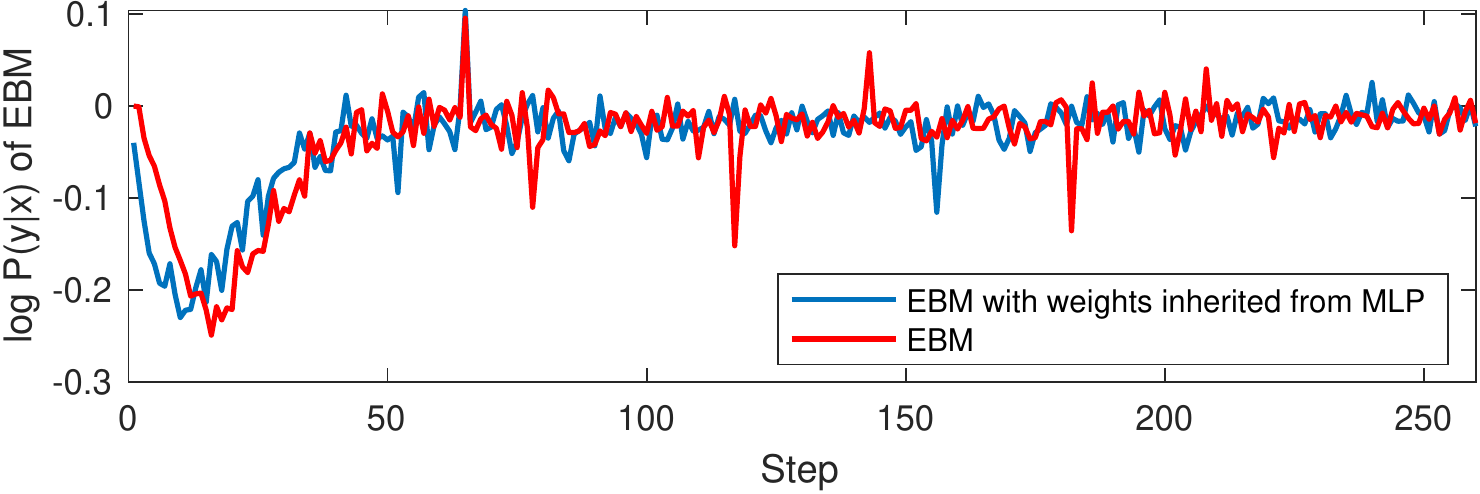}
     \end{subfigure}
      \begin{subfigure}[b]{6.5cm}
         \centering
         \includegraphics[width=\linewidth]{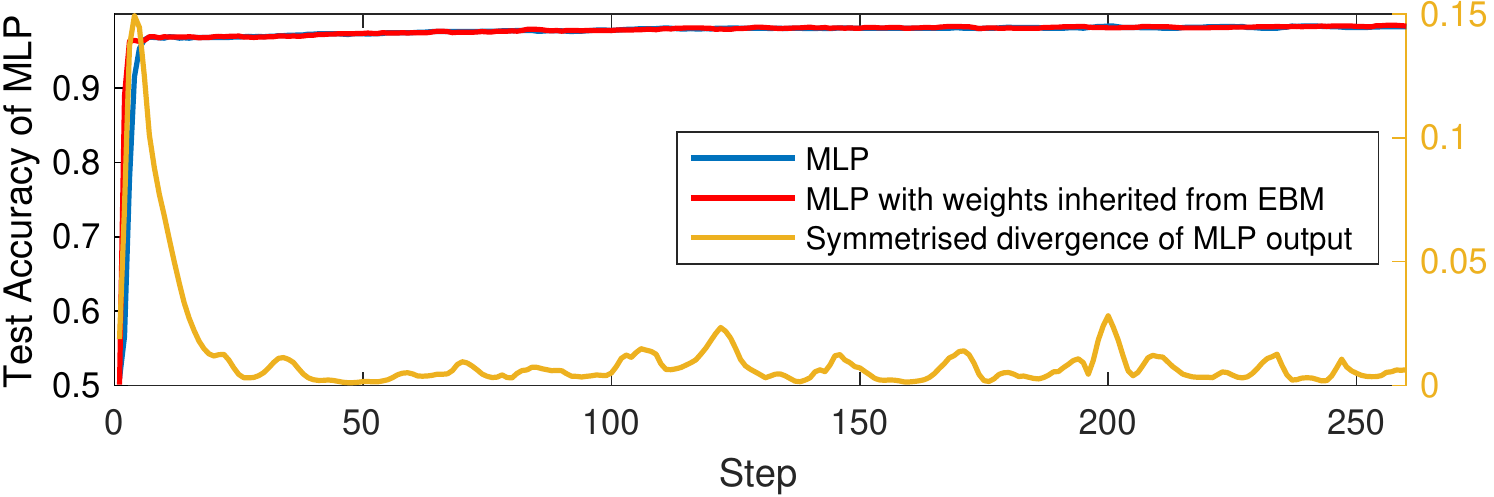}
     \end{subfigure}\\
    \vspace{-5pt}
     \begin{subfigure}[b]{6.5cm}
         \centering
             \includegraphics[width=\textwidth]{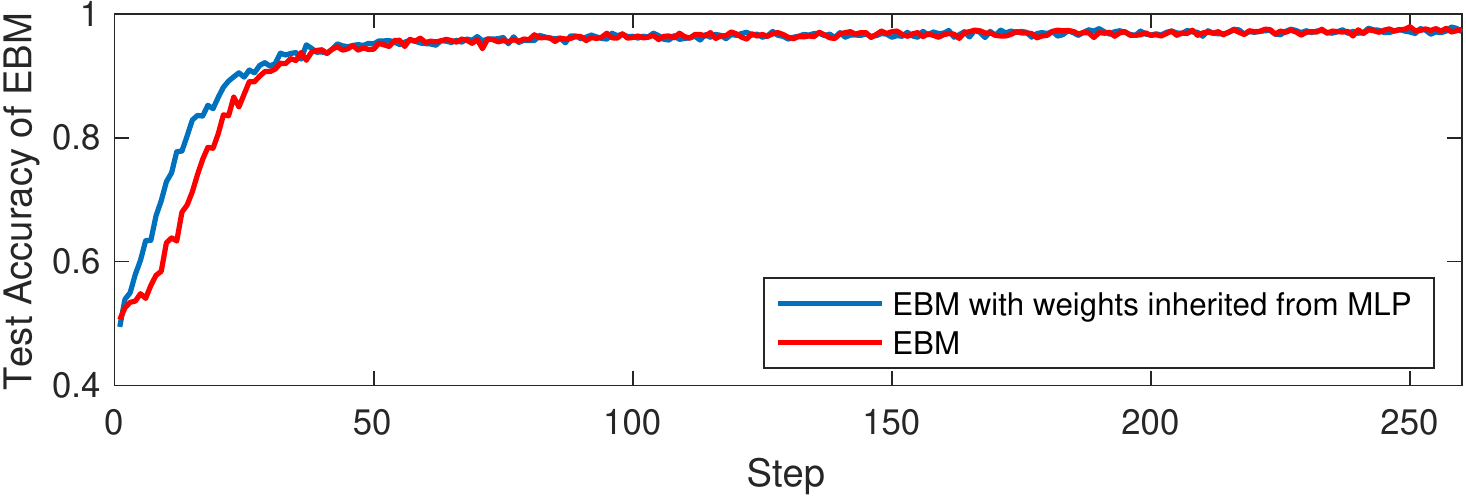}
     \end{subfigure}
     \captionsetup{width=6.5cm}
        \caption{Equivalence of MLP and EBM on MNIST binary classification between digits 3 \& 9. Panels 1 \& 2 show cross-entropy loss and log conditional likelihood (trained and evaluated). Panels 3 \& 4 show the test accuracy. Symmetrised KL divergence is shown in Panel 3.}
    \label{fig:eq_39}
\end{minipage}
\hfill
\vspace{-10pt}
\end{figure*}

\begin{figure*}[!t]
\centering
\begin{minipage}[t]{6.5cm}
\vspace{10pt}
    \centering
     \begin{subfigure}[b]{6.5cm}
         \centering
         \includegraphics[width=\linewidth]{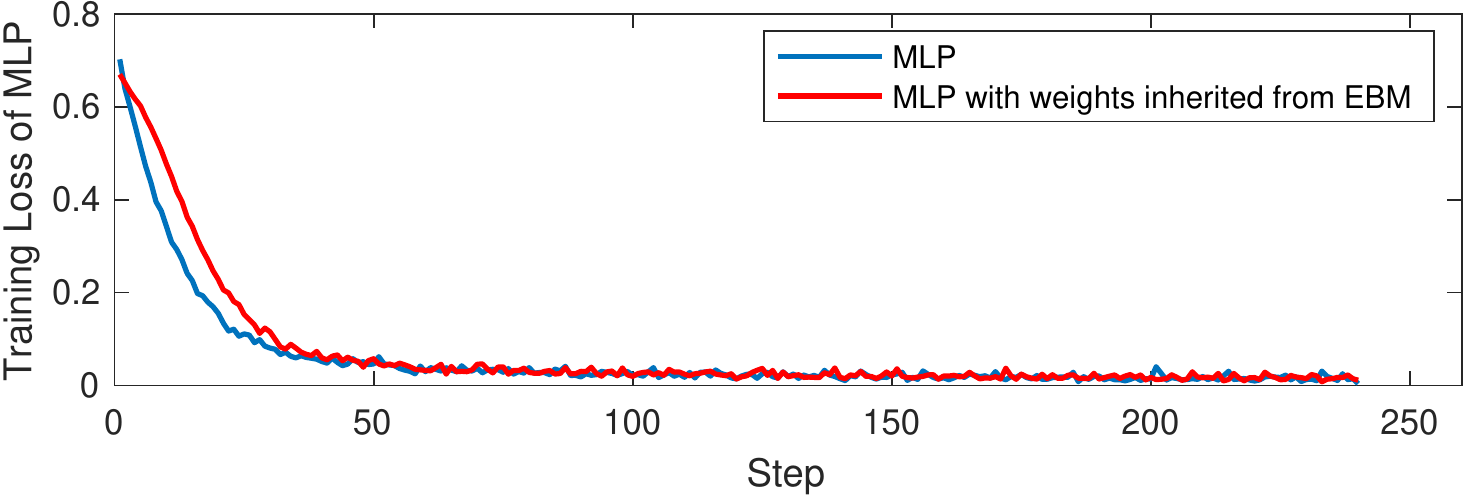}
     \end{subfigure}\\
    \vspace{-5pt}
     \begin{subfigure}[b]{6.5cm}
         \centering
             \includegraphics[width=\textwidth]{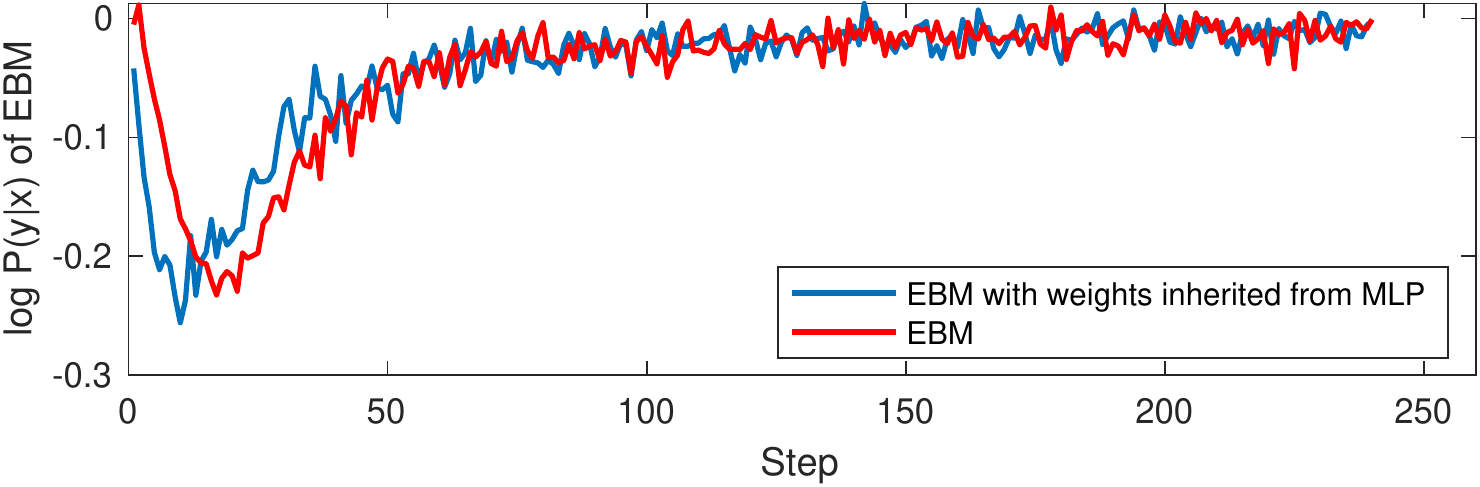}
     \end{subfigure}
      \begin{subfigure}[b]{6.5cm}
         \centering
         \includegraphics[width=\linewidth]{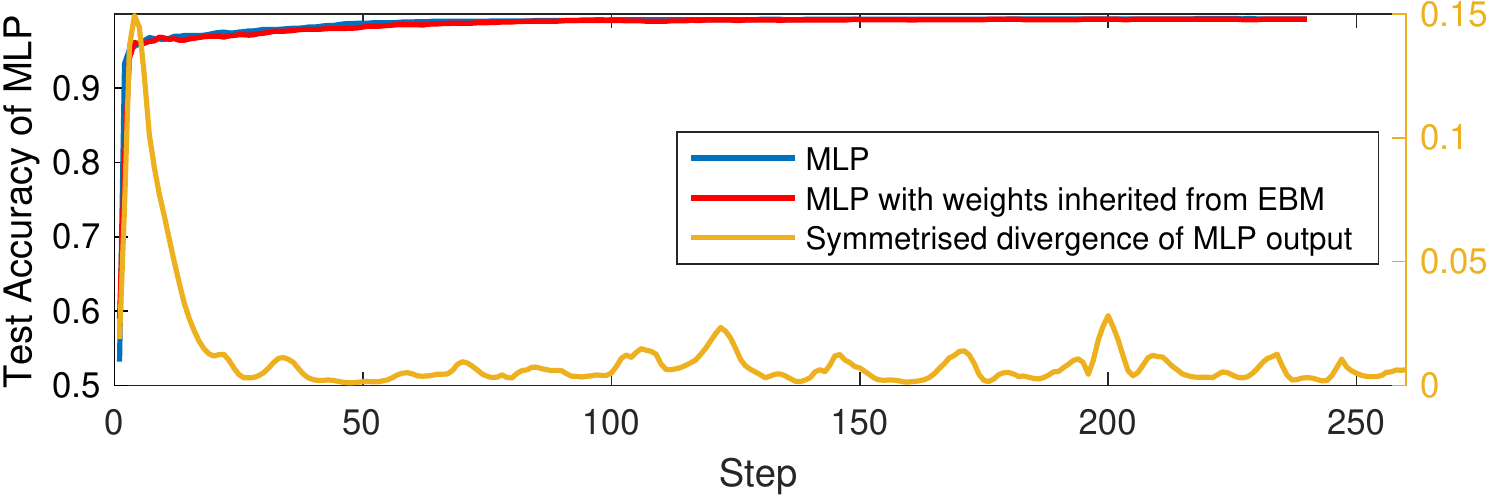}
     \end{subfigure}\\
    \vspace{-5pt}
     \begin{subfigure}[b]{6.5cm}
         \centering
             \includegraphics[width=\textwidth]{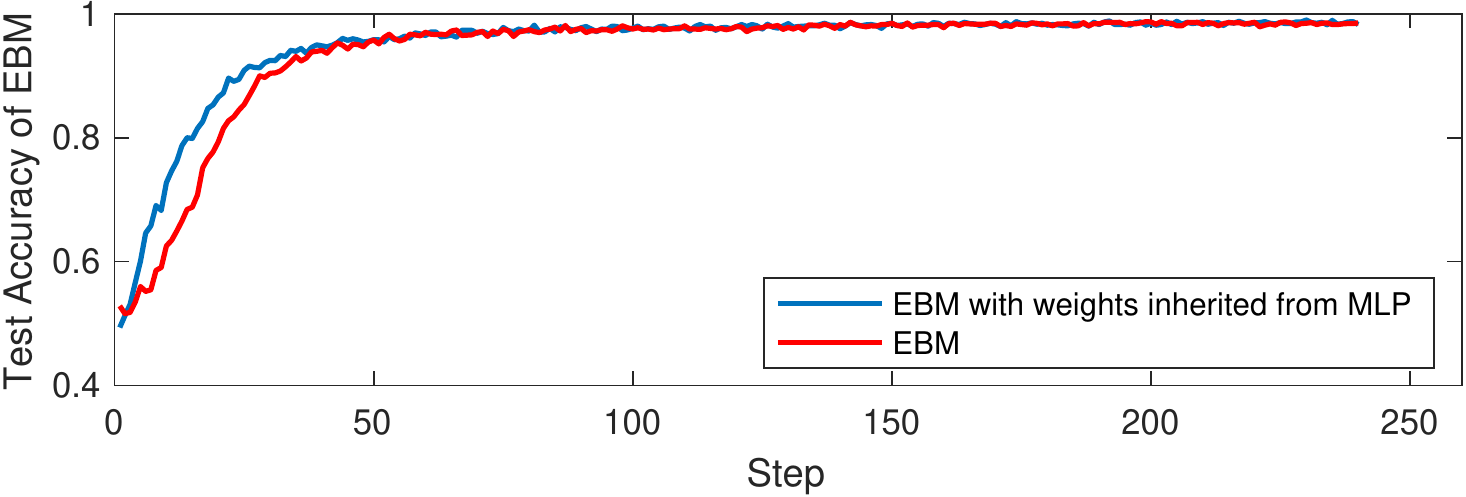}
     \end{subfigure}
    \captionsetup{width=6.5cm}
        \caption{Equivalence of MLP and EBM on MNIST binary classification between digits 5 \& 7. Panels 1 \& 2 show cross-entropy loss and log conditional likelihood (trained and evaluated). Panels 3 \& 4 show the test accuracy. Symmetrised KL divergence is shown in Panel 3.}
    \label{fig:eq_57}
\end{minipage}
\quad
\begin{minipage}[t]{6.5cm}
\vspace{10pt}
    \centering
     \begin{subfigure}[b]{6.5cm}
         \centering
         \includegraphics[width=\linewidth]{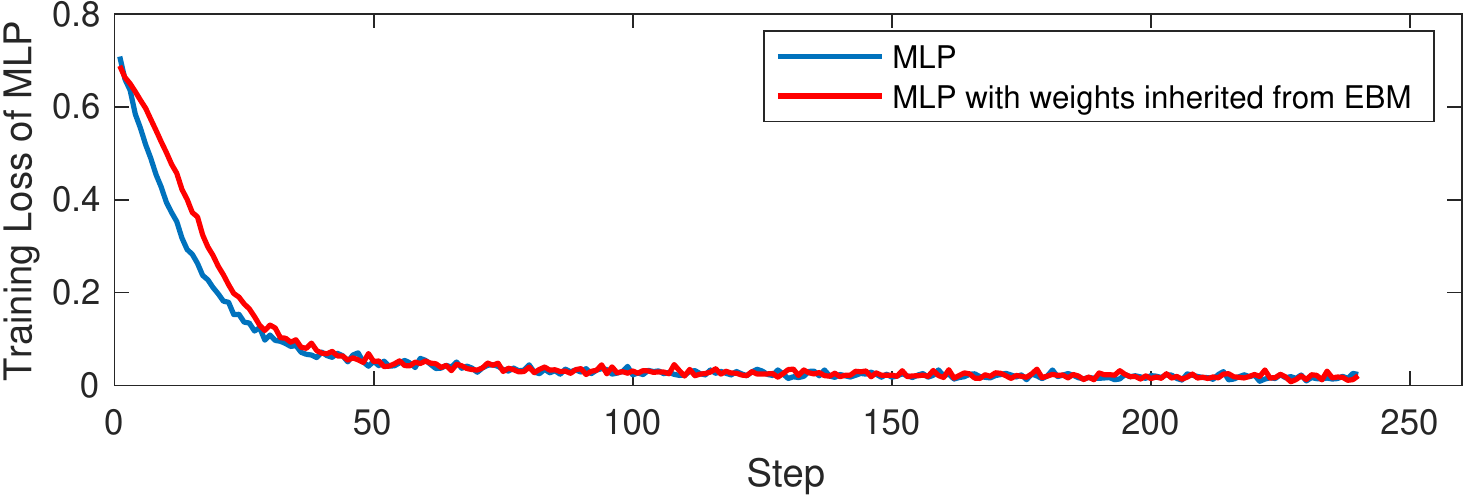}
     \end{subfigure}\\
    \vspace{-5pt}
     \begin{subfigure}[b]{6.5cm}
         \centering
             \includegraphics[width=\textwidth]{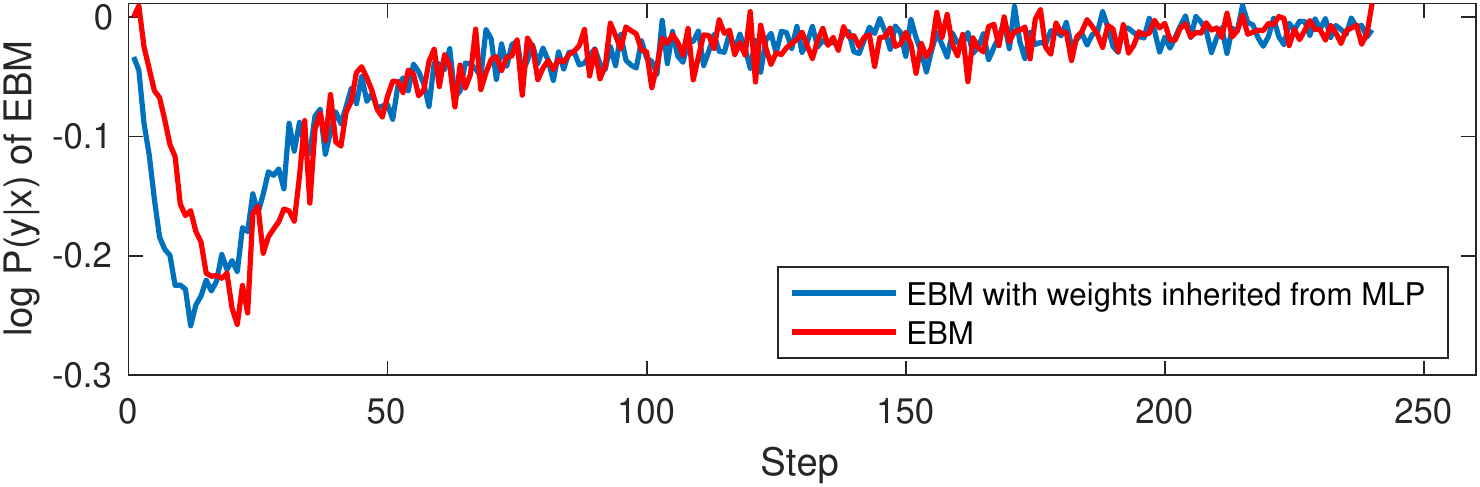}
     \end{subfigure}
      \begin{subfigure}[b]{6.5cm}
         \centering
         \includegraphics[width=\linewidth]{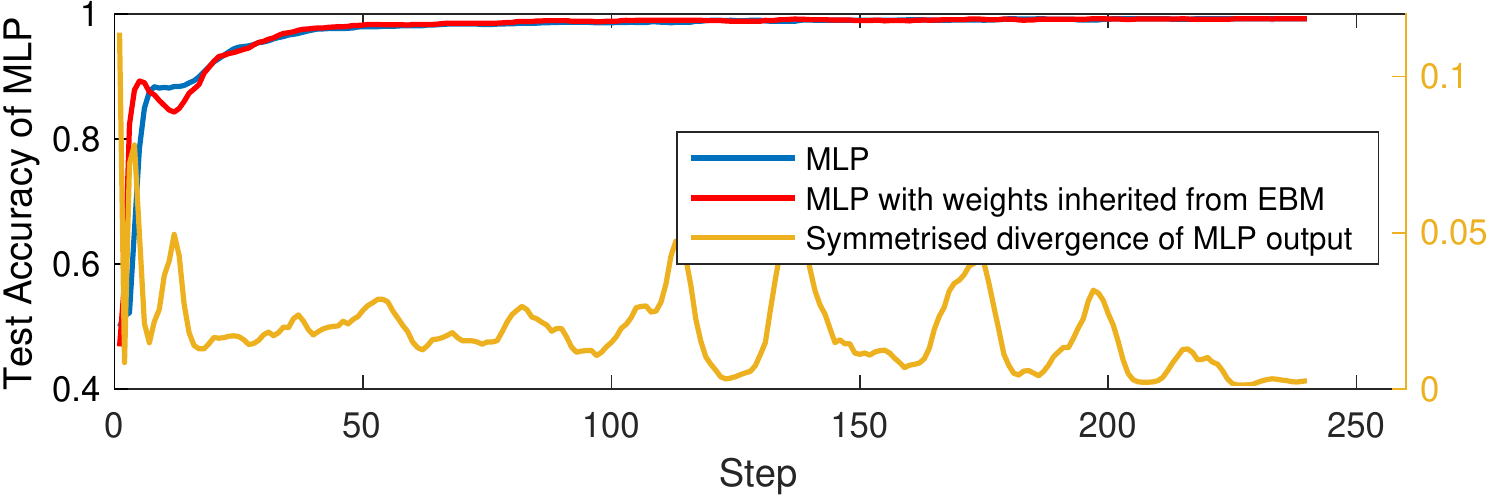}
     \end{subfigure}\\
    \vspace{-5pt}
     \begin{subfigure}[b]{6.5cm}
         \centering
             \includegraphics[width=\textwidth]{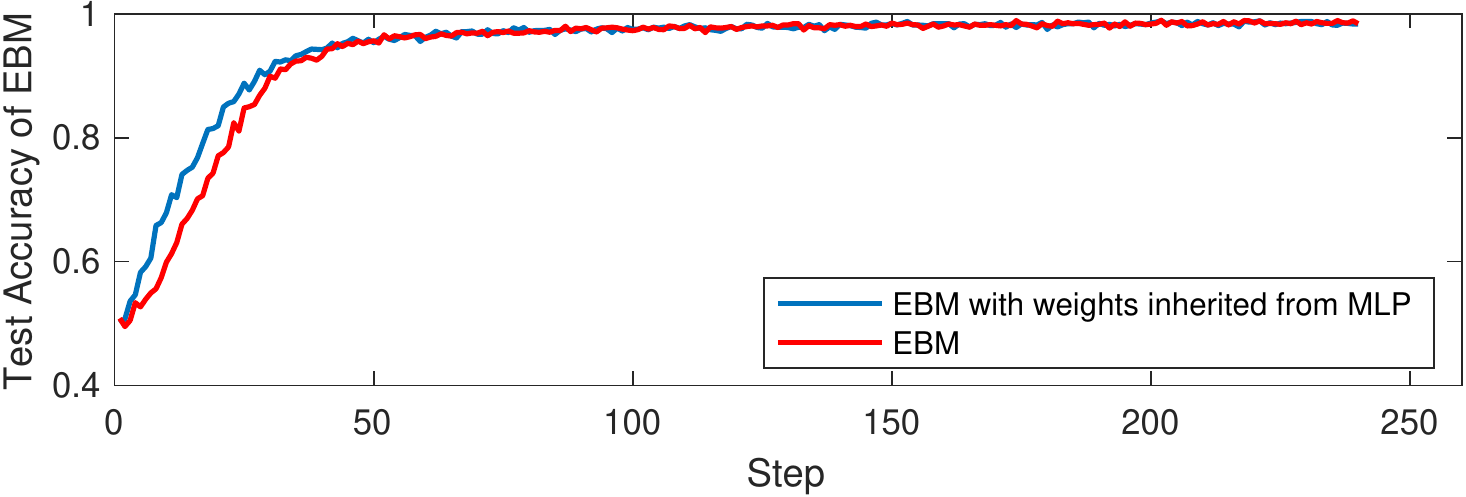}
     \end{subfigure}
     \captionsetup{width=6.5cm}
        \caption{Equivalence of MLP and EBM on MNIST binary classification between digits 4 \& 8. Panels 1 \& 2 show cross-entropy loss and log conditional likelihood (trained and evaluated). Panels 3 \& 4 show the test accuracy. Symmetrised KL divergence is shown in Panel 3.}
    \label{fig:eq_48}
\end{minipage}
\hfill
\vspace{-10pt}
\end{figure*}

\begin{figure*}[!t]
\centering
\begin{minipage}[t]{6.5cm}
\vspace{10pt}
    \centering
     \begin{subfigure}[b]{6.5cm}
         \centering
         \includegraphics[width=\linewidth]{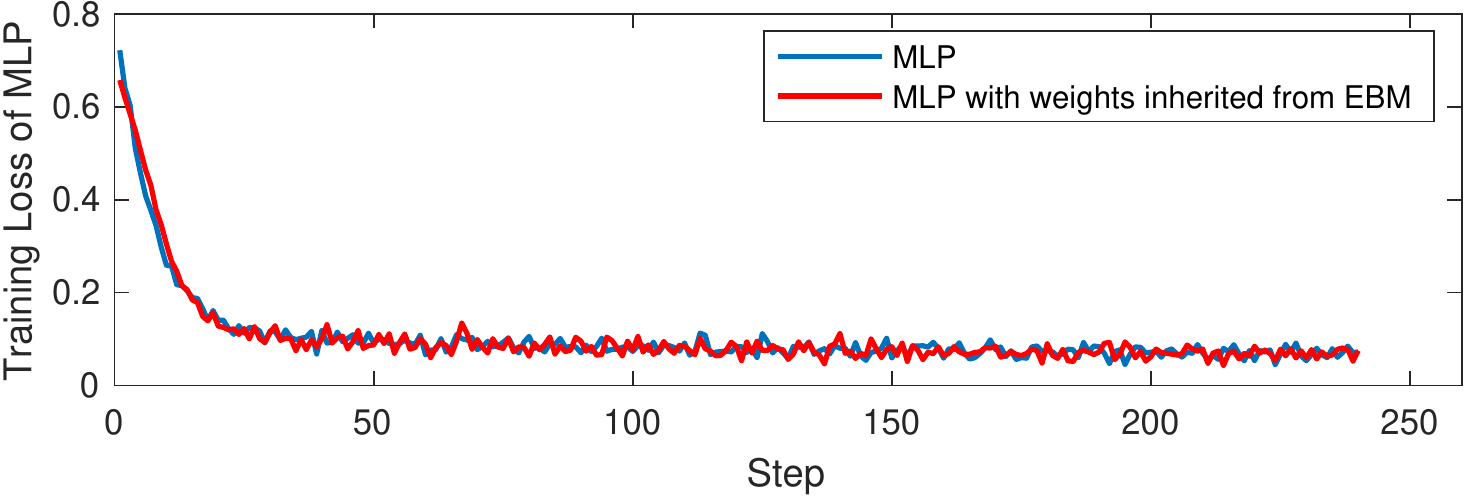}
     \end{subfigure}\\
    \vspace{-5pt}
     \begin{subfigure}[b]{6.5cm}
         \centering
             \includegraphics[width=\textwidth]{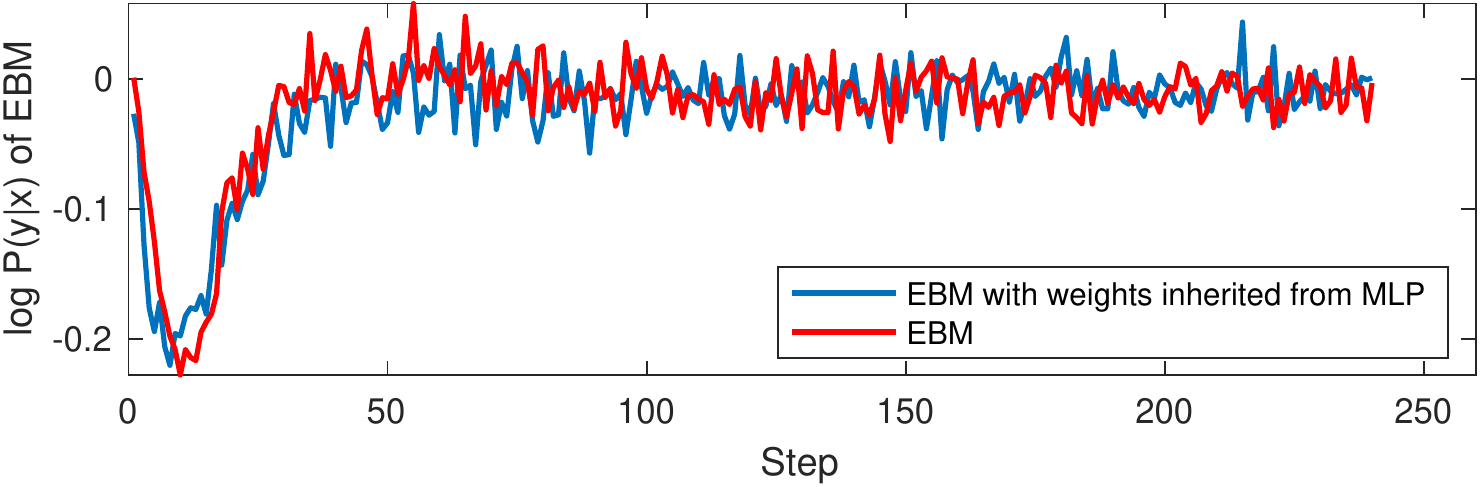}
     \end{subfigure}
      \begin{subfigure}[b]{6.5cm}
         \centering
         \includegraphics[width=\linewidth]{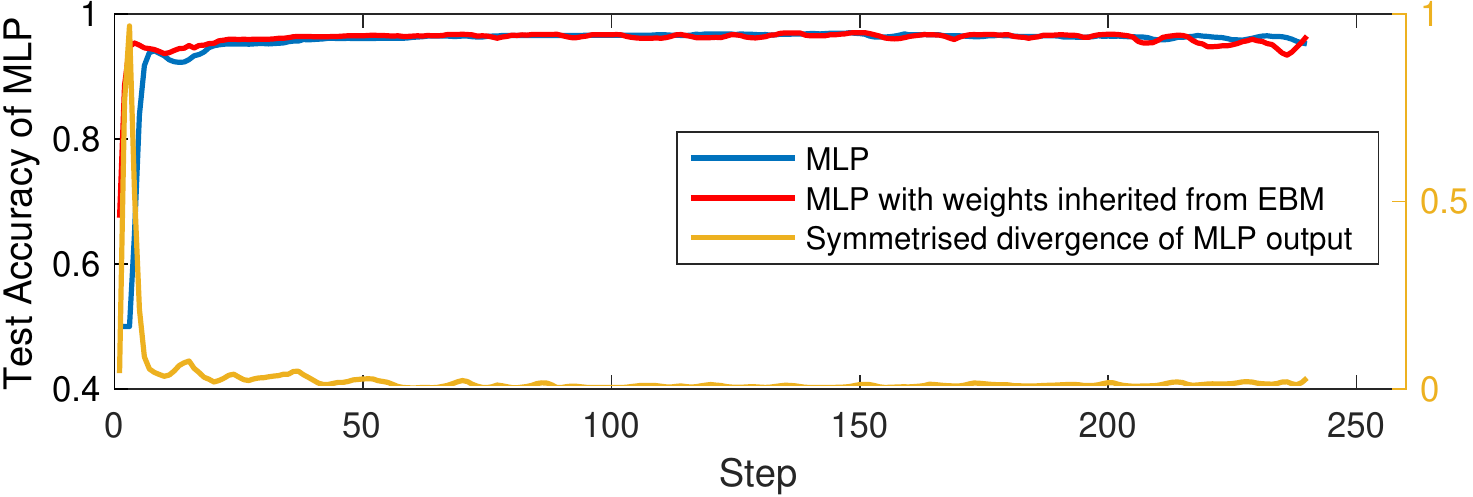}
     \end{subfigure}\\
    \vspace{-5pt}
     \begin{subfigure}[b]{6.5cm}
         \centering
             \includegraphics[width=\textwidth]{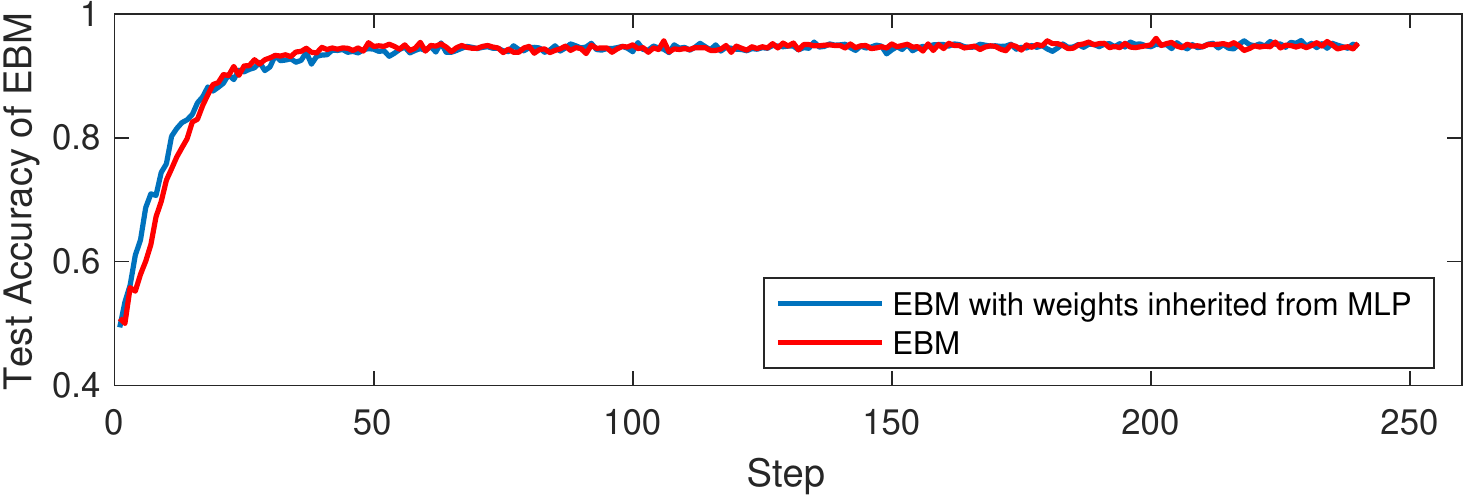}
     \end{subfigure}
    \captionsetup{width=6.5cm}
        \caption{Equivalence of MLP and EBM on Fashion-MNIST binary classification between class Pullover \& Dress. Panels 1 \& 2 show cross-entropy loss and log conditional likelihood (trained and evaluated). Panels 3 \& 4 show the test accuracy. Symmetrised KL divergence is shown in Panel 3.}
    \label{fig:eq_f23}
\end{minipage}
\quad
\begin{minipage}[t]{6.5cm}
\vspace{10pt}
    \centering
     \begin{subfigure}[b]{6.5cm}
         \centering
         \includegraphics[width=\linewidth]{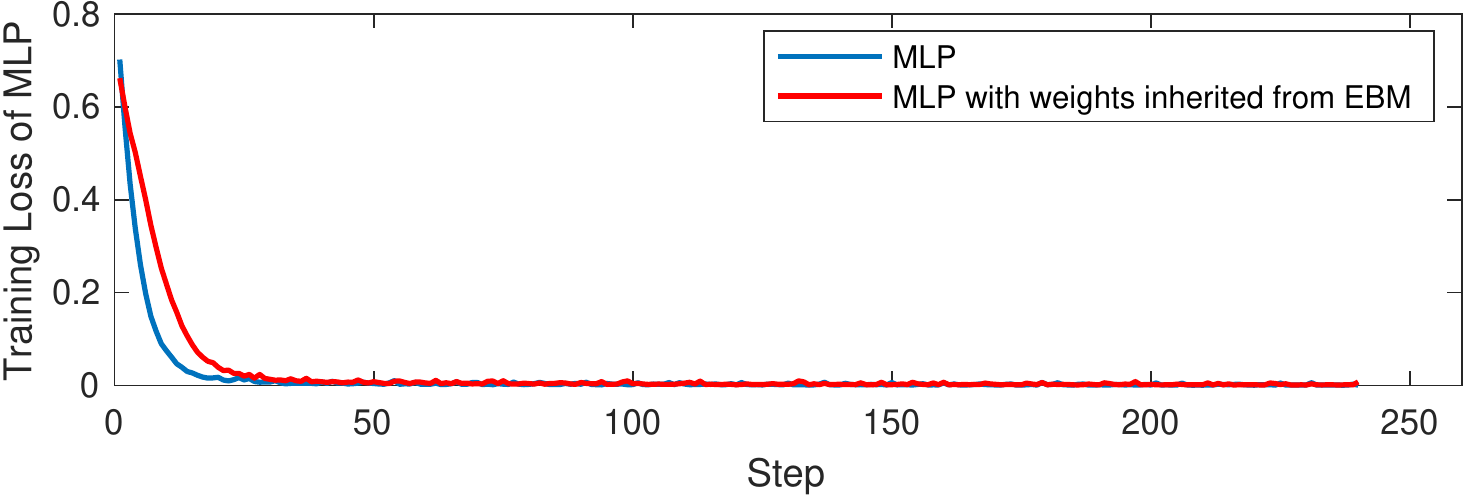}
     \end{subfigure}\\
    \vspace{-5pt}
     \begin{subfigure}[b]{6.5cm}
         \centering
             \includegraphics[width=\textwidth]{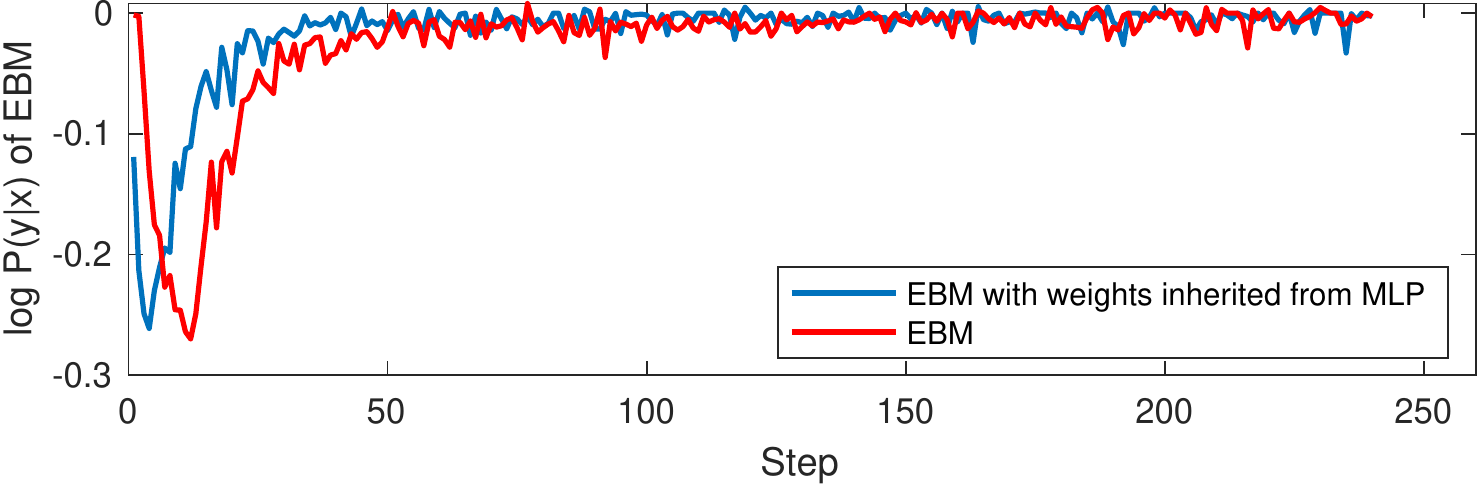}
     \end{subfigure}
      \begin{subfigure}[b]{6.5cm}
         \centering
         \includegraphics[width=\linewidth]{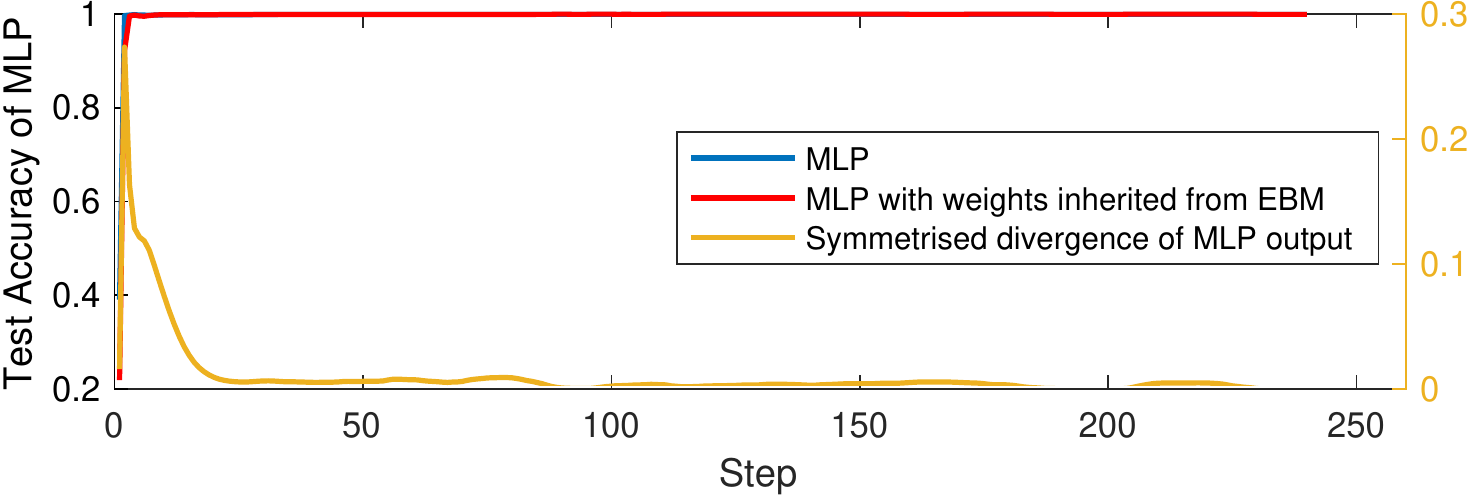}
     \end{subfigure}\\
    \vspace{-5pt}
     \begin{subfigure}[b]{6.5cm}
         \centering
             \includegraphics[width=\textwidth]{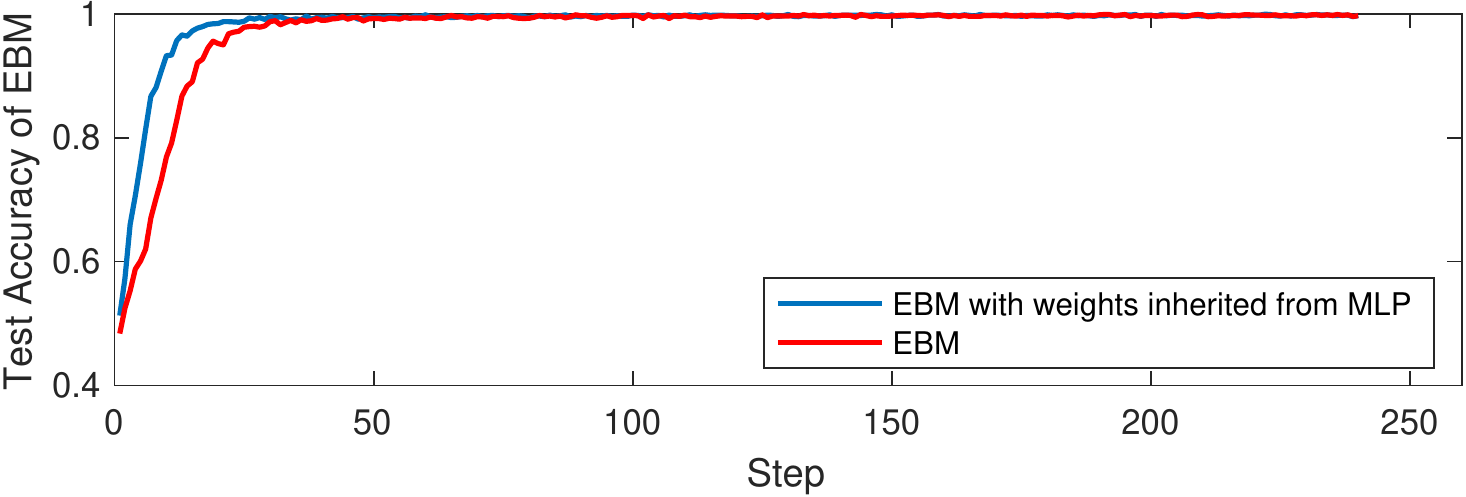}
     \end{subfigure}
     \captionsetup{width=6.5cm}
        \caption{Equivalence of MLP and EBM on Fashion-MNIST binary classification between class Dress \& Ankle boot. Panels 1 \& 2 show cross-entropy loss and log conditional likelihood (trained and evaluated). Panels 3 \& 4 show the test accuracy. Symmetrised KL divergence is shown in Panel 3.}
    \label{fig:eq_f39}
\end{minipage}
\hfill
\vspace{-10pt}
\end{figure*}

Please refer to Fig. \ref{fig:eq_23}--\ref{fig:eq_f39} for more test cases regarding the equivalence analysis described in Sec. \ref{sec:equivexp}. The results were gained on both MNIST and Fashion-MNIST. Note that the digit/class pairs were generated randomly. The same performance metrics as mentioned in the paper have been applied:
\begin{itemize}[leftmargin=1em,itemsep=0pt,parsep=0pt,topsep=2pt]
    \item The cross-entropy loss and test accuracy for the MLP with weights inherited from the EBM.
    \item The log conditional likelihood and test accuracy for the EBM with weights inherited from the MLP.
    \item The symmetrised KL divergence calculated between the sigmoidal output of the MLP with weights trained by backpropagation and that of the MLP that has inherited weights from the EBM.
\end{itemize}

\subsection{Training MLP with quantum sampling}\label{app:quantum}
\begin{figure}
\centering
\begin{subfigure}{0.49\textwidth}
    \includegraphics[width=\textwidth]{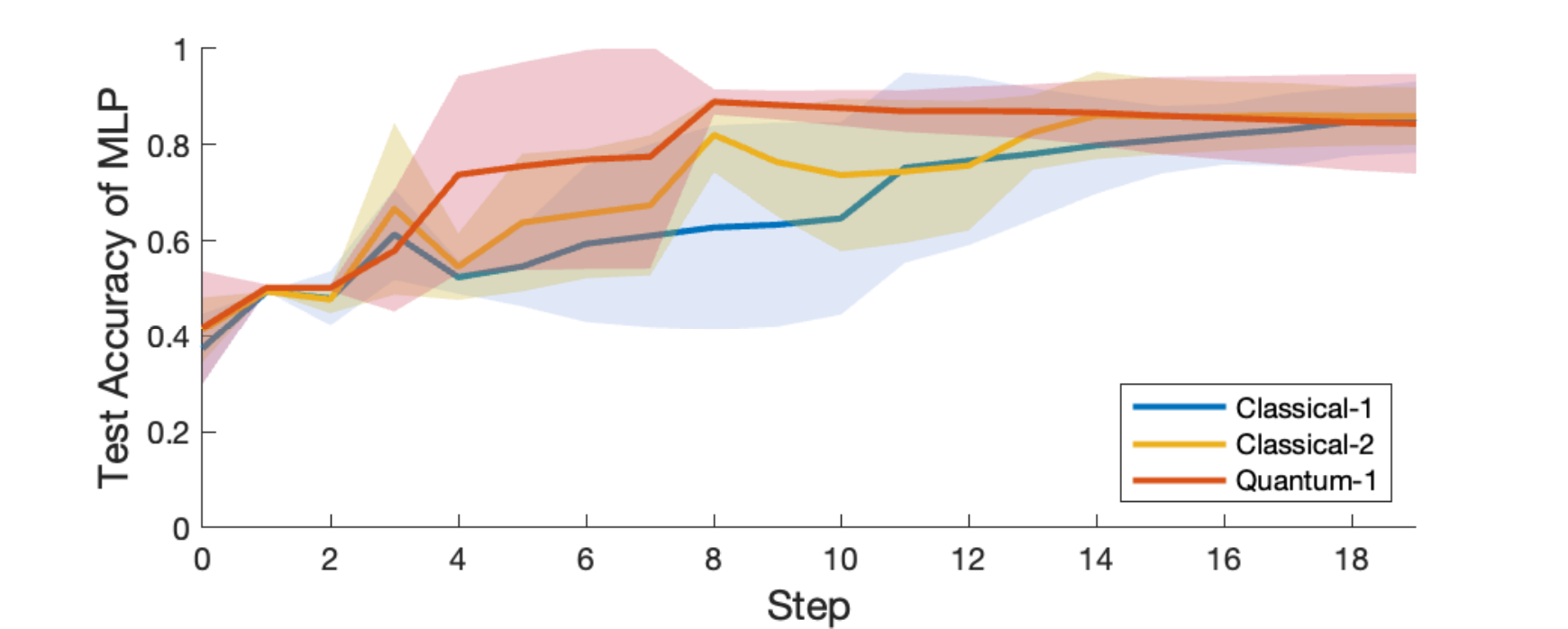}
    \caption{0-6}
\end{subfigure}
\hfill
\begin{subfigure}{0.49\textwidth}
    \includegraphics[width=\textwidth]{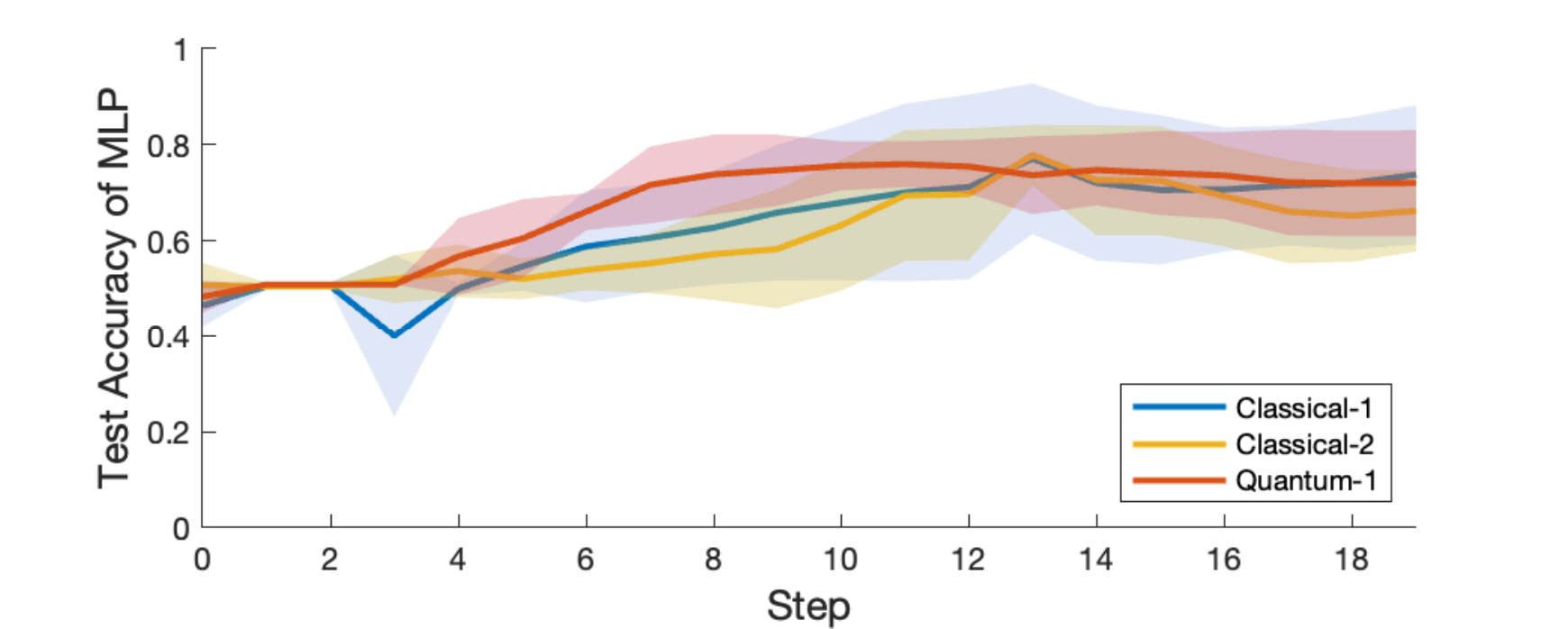}
    \caption{8-9}
\end{subfigure}
\hfill
\begin{subfigure}{0.49\textwidth}
    \includegraphics[width=\textwidth]{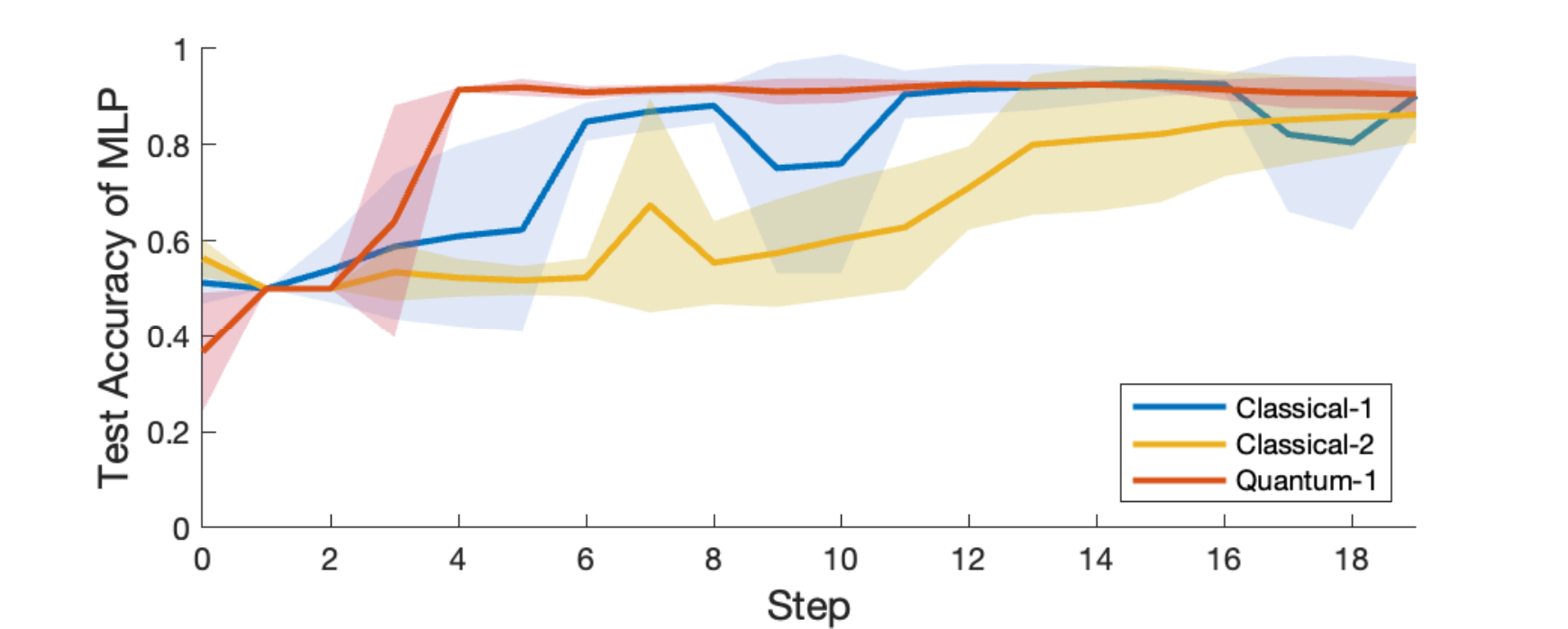}
    \caption{3-9}
\end{subfigure}
\hfill
\begin{subfigure}{0.49\textwidth}
    \includegraphics[width=\textwidth]{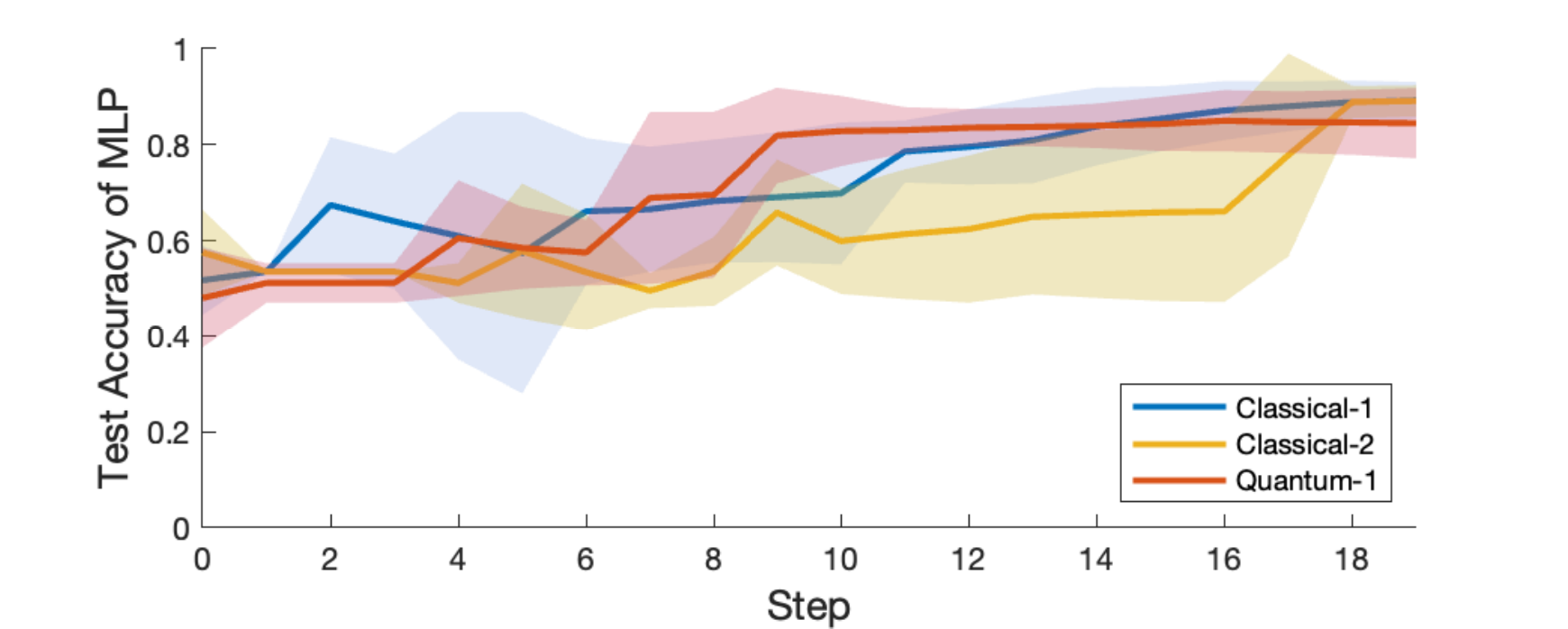}
    \caption{5-7}
\end{subfigure}
\hfill
\begin{subfigure}{0.49\textwidth}
    \includegraphics[width=\textwidth]{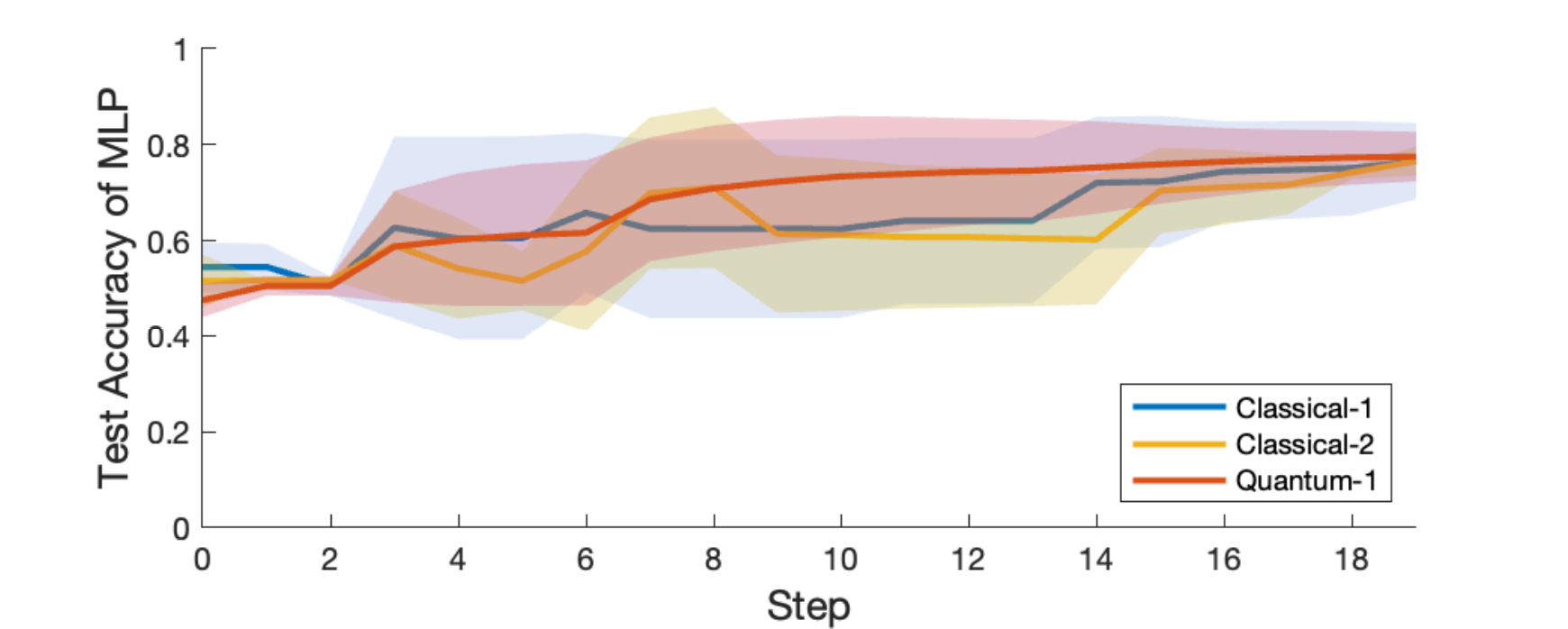}
    \caption{5-6}
\end{subfigure}
\hfill
\begin{subfigure}{0.49\textwidth}
    \includegraphics[width=\textwidth]{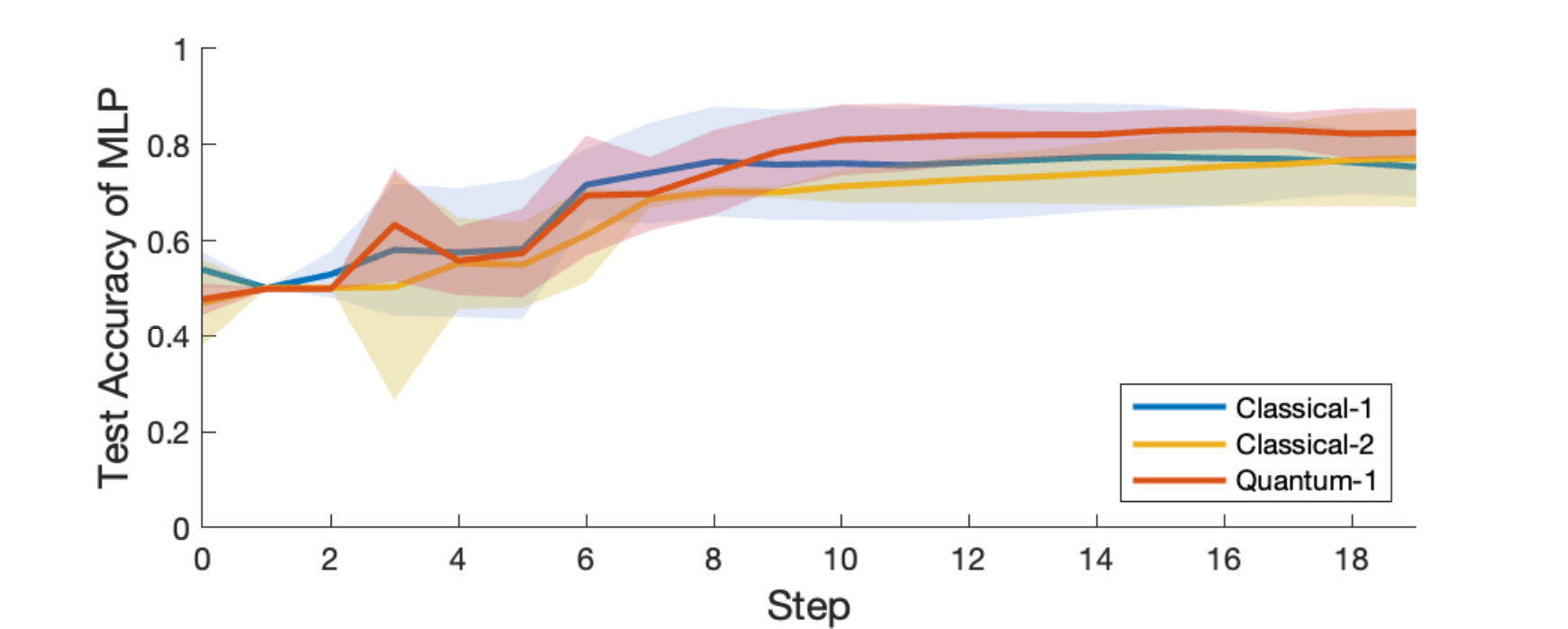}
    \caption{4-8}
\end{subfigure}
\hfill
\begin{subfigure}{0.49\textwidth}
    \includegraphics[width=\textwidth]{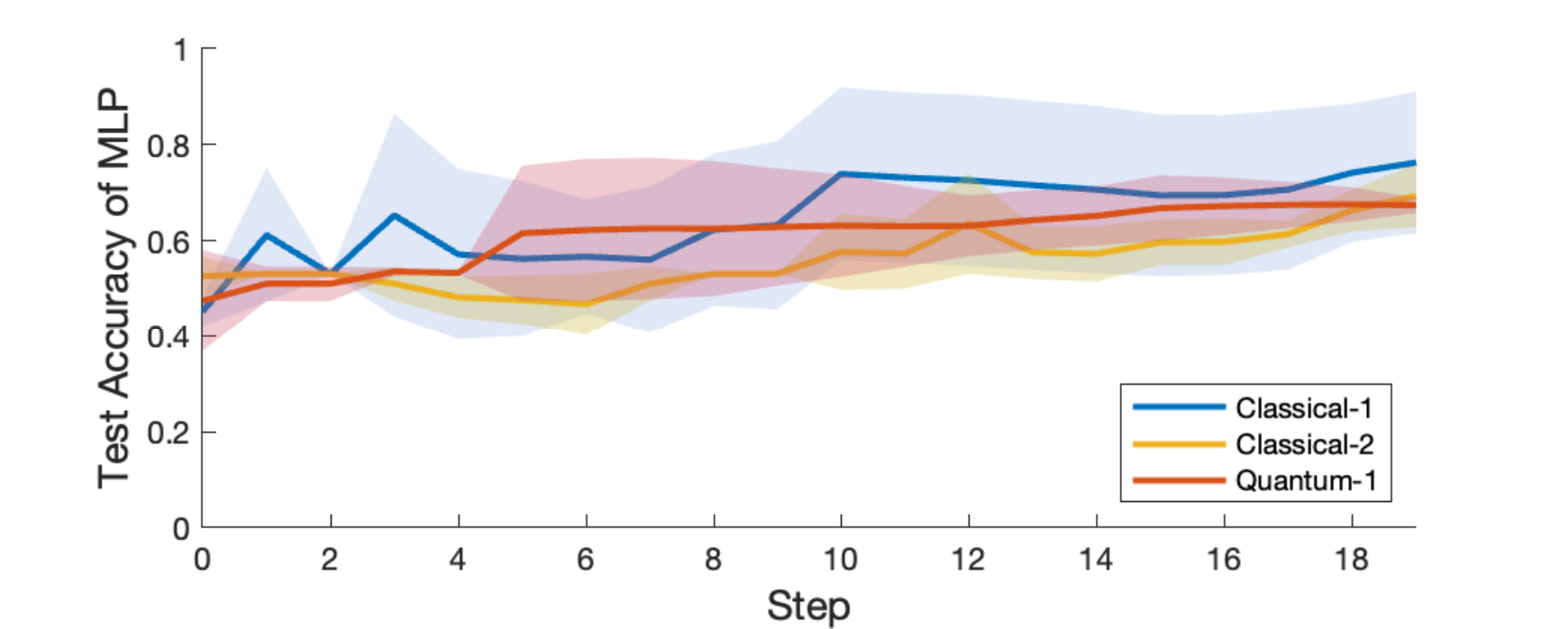}
    \caption{5-9}
\end{subfigure}
\hfill
\begin{subfigure}{0.49\textwidth}
    \includegraphics[width=\textwidth]{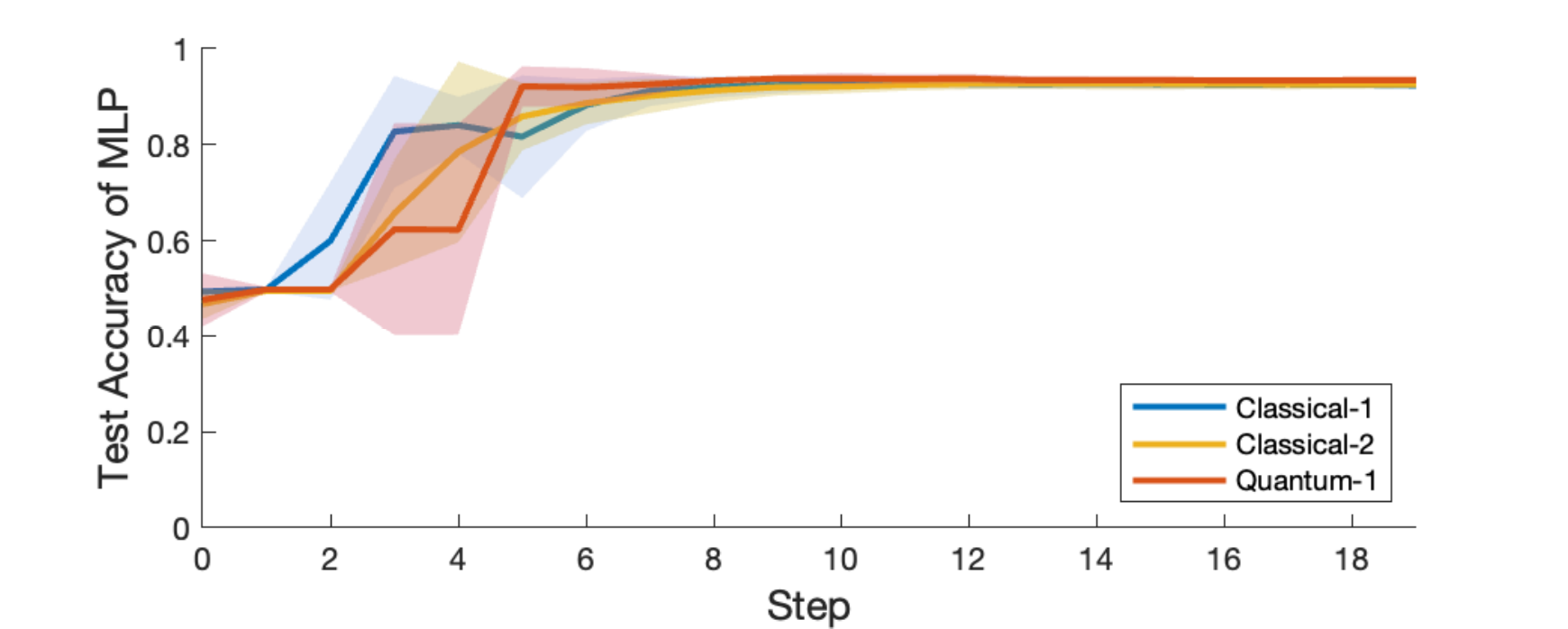}
    \caption{6-8}
\end{subfigure}
\hfill
\begin{subfigure}{0.49\textwidth}
    \includegraphics[width=\textwidth]{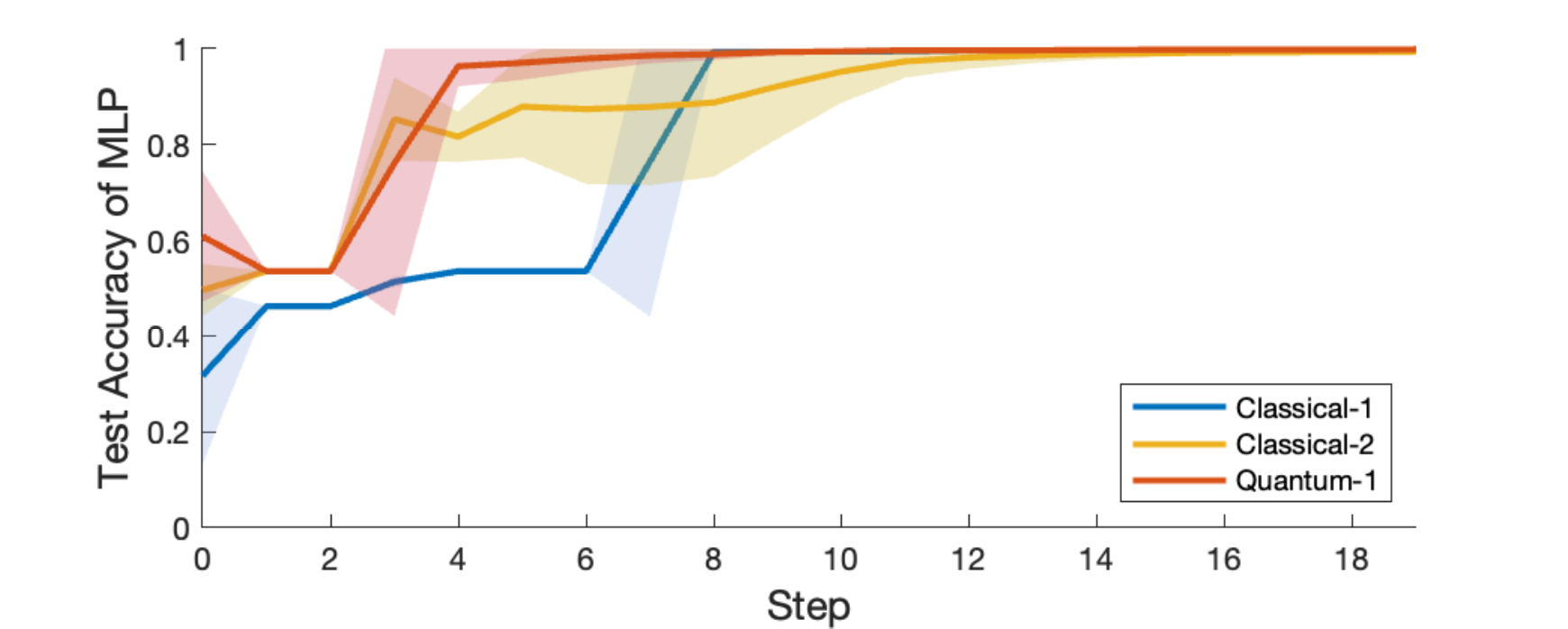}
    \caption{0-1}
\end{subfigure}
\hfill
\begin{subfigure}{0.49\textwidth}
    \includegraphics[width=\textwidth]{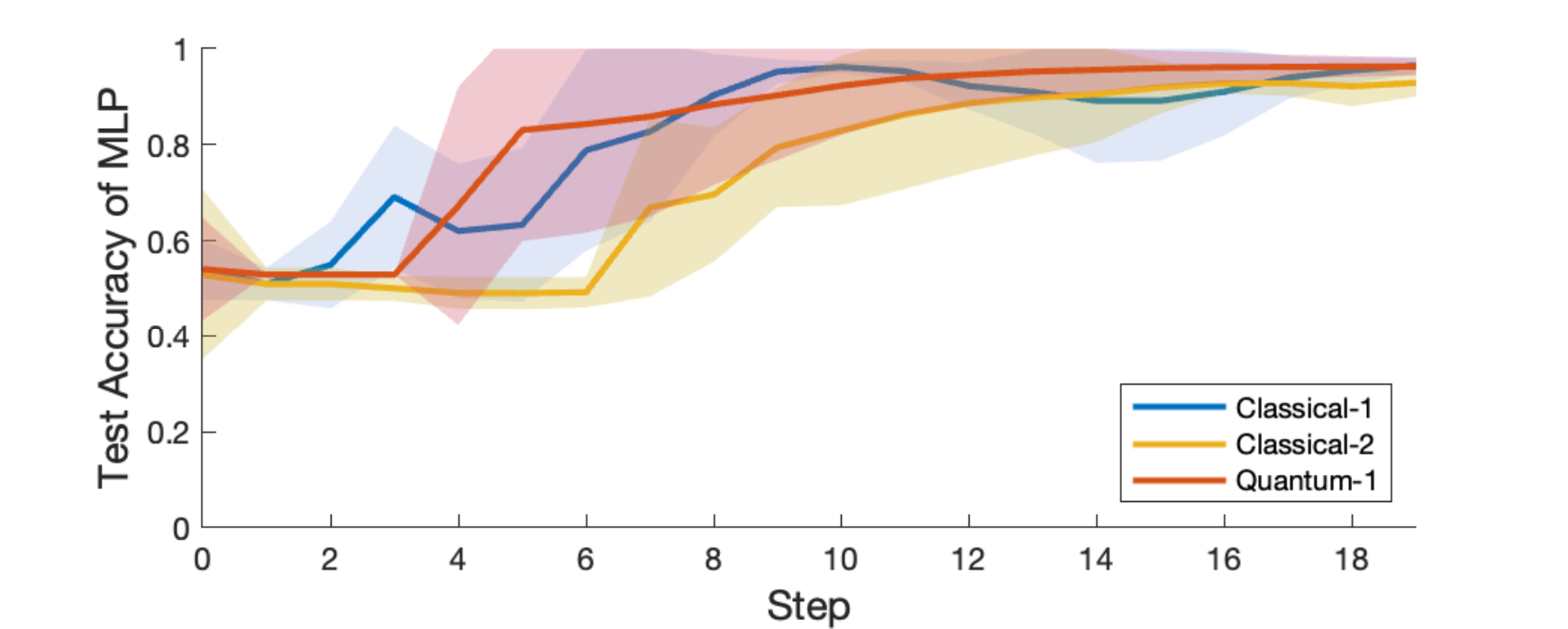}
    \caption{1-3}
\end{subfigure}
\captionsetup{width=13cm}        
\caption{Test accuracy of MLP for binary classification of digit pairs randomly selected from MNIST averaged over at least 3 successful runs. The unbiased sample std.~dev.~is shown as the shaded area. The model parameters were trained by MLP (Classical-1), EBM with Gibbs sampling (Classical-2), and EBM with quantum sampling (Quantum-1).}\vspace{-10pt}
\label{exp_mnist}
\end{figure}

\begin{figure}
\centering
\begin{subfigure}{0.49\textwidth}
    \includegraphics[width=\textwidth]{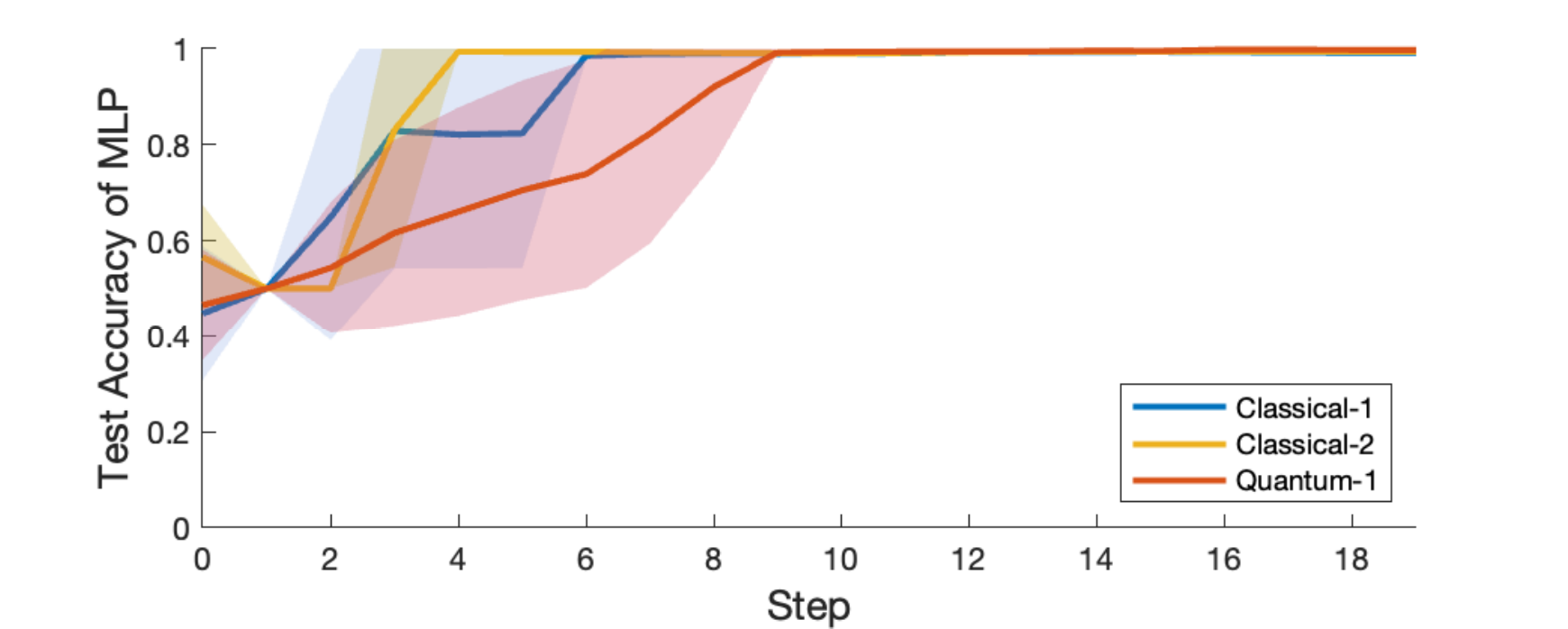}
    \caption{Dress-Ankle boot}
\end{subfigure}
\hfill
\begin{subfigure}{0.49\textwidth}
    \includegraphics[width=\textwidth]{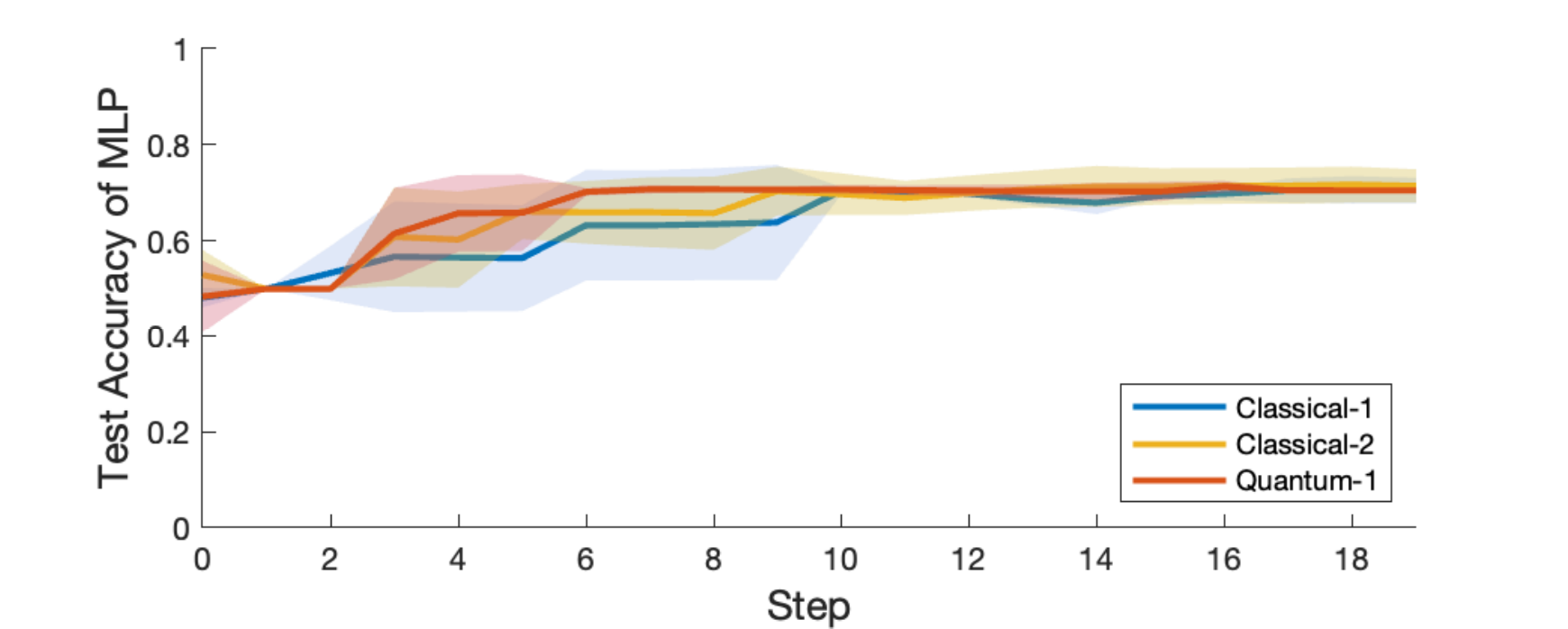}
    \caption{Sandal-Sneaker}
\end{subfigure}
\hfill
\begin{subfigure}{0.49\textwidth}
    \includegraphics[width=\textwidth]{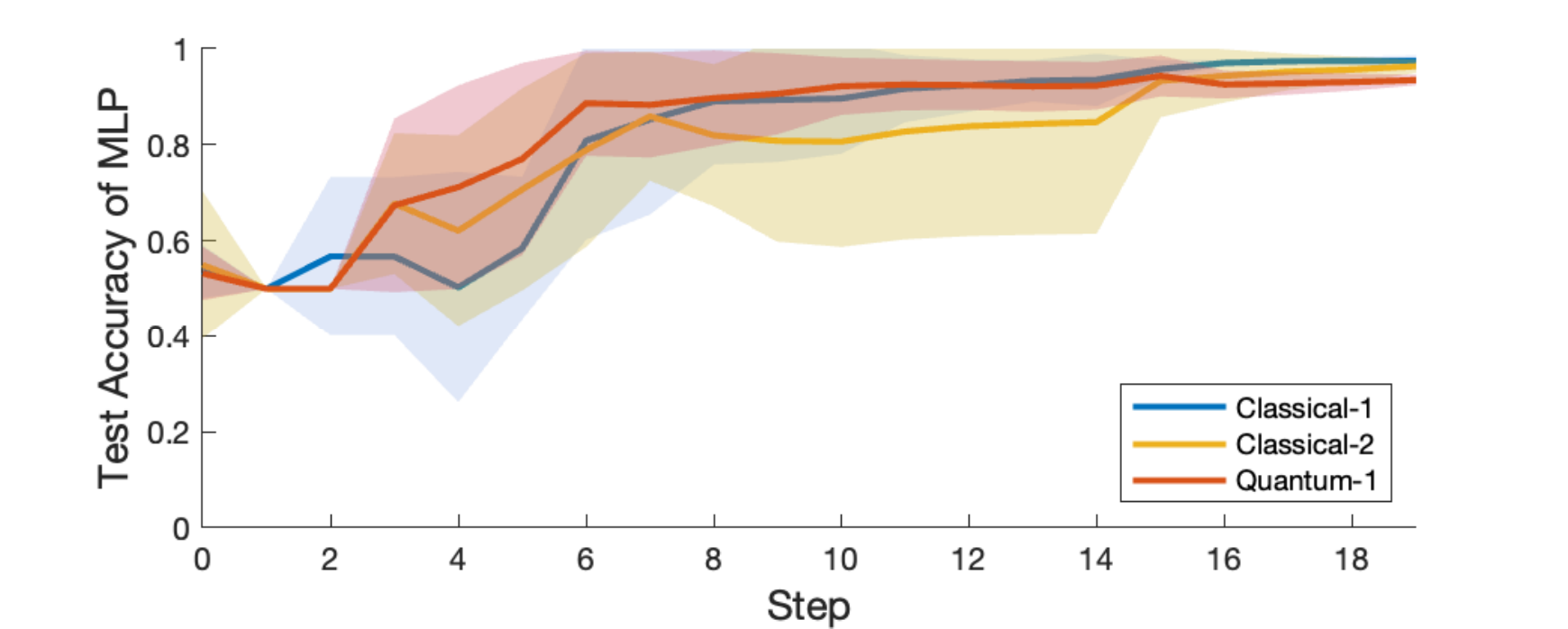}
    \caption{Sandal-Shirt}
\end{subfigure}
\hfill
\begin{subfigure}{0.49\textwidth}
    \includegraphics[width=\textwidth]{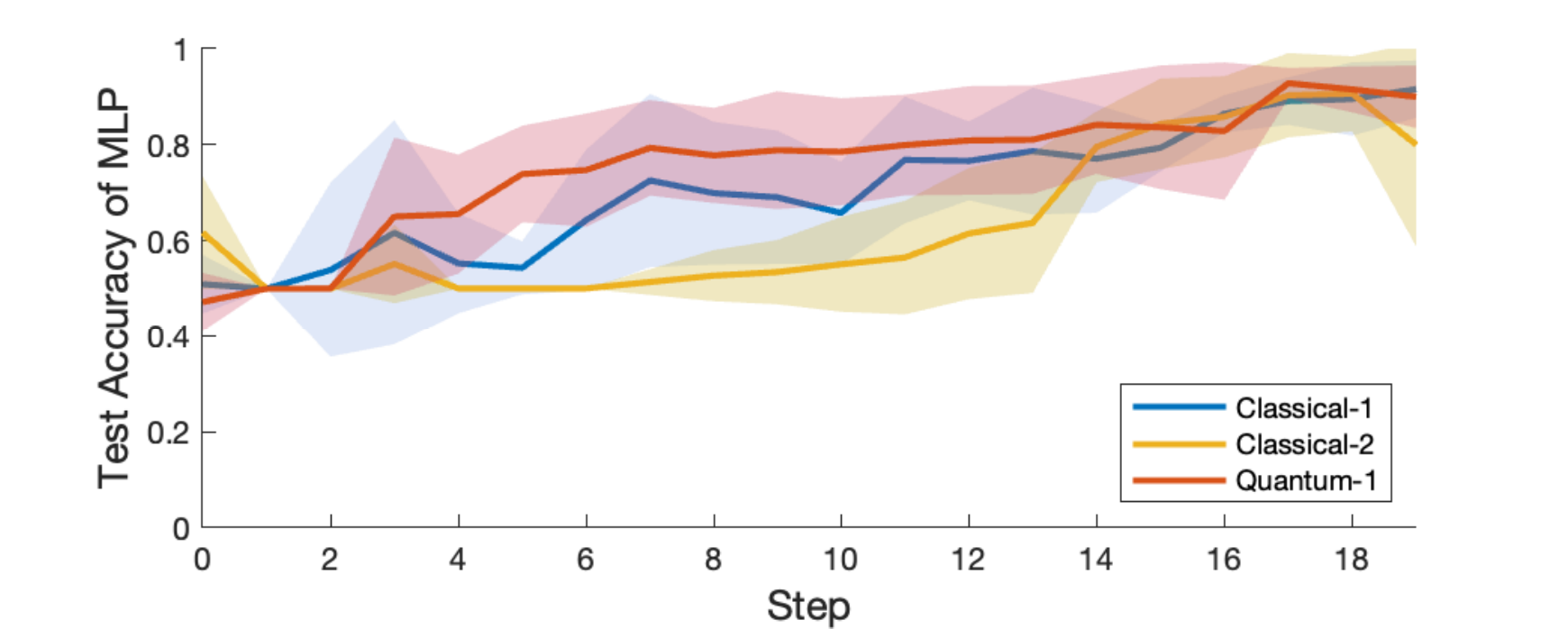}
    \caption{Coat-Bag}
\end{subfigure}
\captionsetup{width=13cm}        
\caption{Test accuracy of MLP for binary classification of classes randomly selected from Fashion-MNIST averaged over at least 3 successful runs. The unbiased sample std.~dev.~is shown as the shaded area. The model parameters were trained by MLP (Classical-1), EBM with Gibbs sampling (Classical-2), and EBM with quantum sampling (Quantum-1).}\vspace{-10pt}
\label{exp_fashionmnist}
\end{figure}

We present additional evaluation results of the three experimental setups introduced in Sec. \ref{sec:quantum}. Fig. \ref{exp_mnist} and \ref{exp_fashionmnist} contain evaluation results on more test cases plotted in the same way as Fig. \ref{quantum_compare}. The plots cover the test accuracy of the three setups on all the test cases summarised in Table \ref{tab:acc}--\ref{tab:success}.

\end{appendices}

\end{document}